\documentclass[11pt]{article}

\usepackage[utf8]{inputenc}
\usepackage[T1]{fontenc}
\usepackage{lmodern}
\usepackage{hyperref}
\usepackage{url}
\usepackage{booktabs}
\usepackage{amsfonts}
\usepackage{nicefrac}
\usepackage{microtype}
\usepackage{subcaption}
\usepackage[dvipsnames]{xcolor}

\usepackage{parskip}

\usepackage{amsmath}
\usepackage{amssymb}
\usepackage{mathtools}
\usepackage{amsthm}
\usepackage{graphicx}
\usepackage{float}
\usepackage{natbib}         

\floatstyle{ruled}
\newfloat{algorithm}{tbp}{loa}
\floatname{algorithm}{Algorithm}
\newenvironment{algorithmic}[1][]{\begin{enumerate}\setlength{\itemsep}{2pt}\setlength{\parskip}{1pt}}{\end{enumerate}}
\newcommand{\State}{\item}
\newcommand{\Require}{\item[] \hspace*{-\leftmargin}\textbf{Input: }}
\newcommand{\For}[1]{\item[]\textbf{for }#1\textbf{ do}\begin{enumerate}}
\newcommand{\EndFor}{\end{enumerate}\item[]\textbf{end for}}
\newcommand{\If}[1]{\item[]\textbf{if }#1\textbf{ then}\begin{enumerate}}
\newcommand{\EndIf}{\end{enumerate}\item[]\textbf{end if}}
\newcommand{\Return}{\textbf{return} }

\usepackage[margin=1.2in]{geometry}

\newtheorem{theorem}{Theorem}
\newtheorem{lemma}[theorem]{Lemma}

\newtheorem{proposition}[theorem]{Proposition}

\newtheorem{assumption}[theorem]{Assumption}

\title{How Efficient Is Your Agent?\\ Online Predictive Bounds under Budget Constraints}
\newcommand{\ourmethod}{\texttt{PROBE}}
\newcommand{\fullmethodname}{\textbf{PR}edictive Bounds for \textbf{O}nline \textbf{B}udget \textbf{E}fficiency}

\author{
    {Roy Maor Lotan}\thanks{Equal contribution.}$^{\:\,1}$ \quad
    {Shai Feldman}$^{*2}$ \quad
    {Yaniv Romano}$^{1,2}$ \\[0.2cm]
    $^1$Department of Electrical and Computer Engineering, Technion, Israel \\
    $^2$Department of Computer Science, Technion, Israel
}

\date{}

\providecommand{\aistatstitle}[1]{\begin{center}
    \LARGE\bfseries #1\par
  \end{center}}

\begin{document}

\maketitle

\begin{abstract}
We evaluate AI agents in an online setting where tasks arrive sequentially and, before running each task, we construct an upper predictive bound on its \emph{time-to-success}---the number of interactions until the task is completed successfully. Under a limited budget, some runs are stopped early, so their time-to-success is unobserved (right-censored), and the correctness of the bound may be unknown. We introduce \emph{\fullmethodname} (\ourmethod{}), a method for constructing calibrated upper predictive bounds while adaptively allocating a limited budget across agent tasks to reduce the anticipated coverage deficit. Our method calibrates the bounds online using inverse-probability-weighted feedback together with a correction designed to minimize coverage deficit while remaining unbiased. We prove that \ourmethod{} provides a coverage guarantee for any finite time horizon while satisfying the budget constraint. Experiments on software engineering, retrieval-augmented generation, and instruction following show that \ourmethod{} attains a coverage rate closer to the target level with tighter predictive bounds than a calibrated baseline.
\end{abstract}

\begin{figure}[t!]
    \centering
    \includegraphics[width=0.99\linewidth]{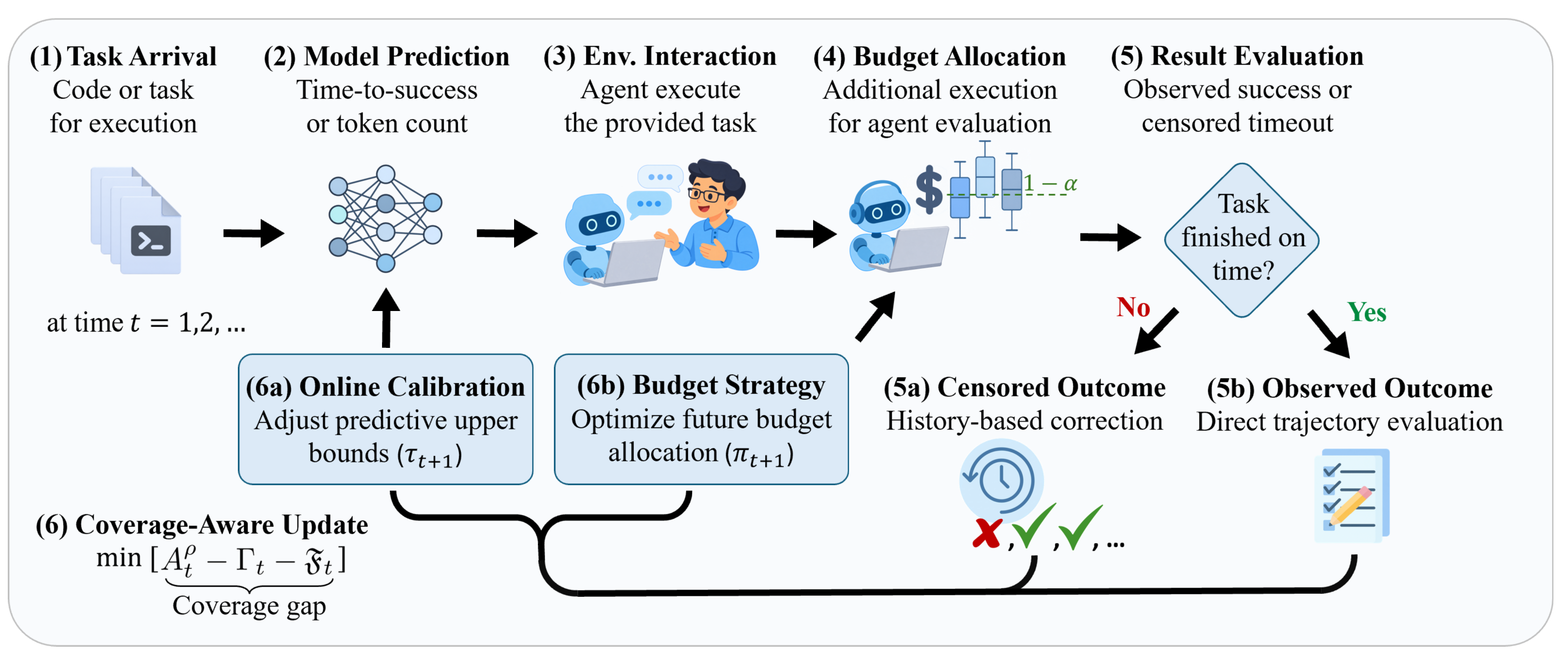}
    \caption{\textbf{Illustration of \ourmethod{} framework.} For each arriving task (1), the system predicts the required time-to-success or token count (2). After environmental interaction where the agent executes the provided task (3), a budget is allocated for additional execution for agent evaluation (4). The result evaluation step (5) determines whether the task finished on time, leading to either a censored outcome with history-based correction (5a) or an observed outcome with direct trajectory evaluation (5b). Both outcomes are integrated via a coverage-aware update (6) (Theorem~\ref{thm:probe-coverage}) consisting of (6a) online calibration to adjust predictive upper bounds ($\tau_{t+1}$) and (6b) a budget strategy to optimize future budget allocation ($\pi_{t+1}$).}
        \label{fig:method_illustration}
\end{figure}

\section{Introduction}
\label{sec:intro}

AI agents rarely complete complex tasks in a single model call~\citep{park2023generative, wang2024survey}. A coding agent, for example, may inspect a repository, modify files, run commands, diagnose errors, and repeatedly revise its approach before producing a correct solution~\citep{yang2024swe,jimenez2024swe}. To effectively evaluate agents, we should therefore account not only for the final outcome, but also for how many interactions were required to reach the event of interest.

We consider an online setting where tasks arrive sequentially as a stream. At time $t$, the agent receives a task $X_t$ and interacts with its environment until either the task is stopped or successfully completed. Success is determined by an external verifier, such as a unit test, a task-specific correctness checker, or an independent LLM-based evaluator~\citep{zheng2023judging}. Only then does the next task $X_{t+1}$ arrive. An interaction may correspond to a model call, a tool invocation, an environment action, or an edit-and-test cycle, depending on the application~\citep{ma2024agentboard, xie2024osworld}.

Following~\citet{davidov2026calibrated, feldman2026}, we define $T_t$ as the \emph{time-to-success}---the number of interactions required for the agent to complete task $X_t$. For a software-engineering agent, $T_t$ may count the number of edit-and-test iterations required to fix a bug; for a browser agent, it may count the actions required to complete a transaction; for a research or data-analysis agent, it may count the tool calls required to produce a verifiably correct response~\citep{zhou2024webarena,yang2024swe}. 
Unlike a binary success indicator, $T_t$ captures the computational effort required to obtain that success and hence provides a more informative measure of agentic efficiency.

Our goal is to construct a calibrated \emph{upper predictive bound} (UPB) $\hat{U}_t(X_t)$ for the unknown time-to-success $T_t$ of an arriving task $X_t$ before executing it. We write $U_t=\hat U_t(X_t)$ for ease of notation. 
Formally, for a given horizon $T$,
 at a user-specified target level $1-\alpha$, we seek to satisfy the following finite-horizon average conditional coverage requirement:
\vspace{-0.05cm}
\begin{equation}\label{eq:coverage_req}
    \frac{1}{T}\sum_{t=1}^{T}\mathbb P\!\left(
        T_t\le U_t
        \,\middle|\,
        X_t,\mathcal D_{t-1}
    \right)
    \ge 1-\alpha.
\end{equation}
Here, $\mathcal D_{t-1}$ denotes the completed observed history before task
$t$, as defined in Section~\ref{subsec:problem_setup}.  

In practice, however, we do not always observe the time-to-success for all tasks as agent interactions are computationally expensive. We therefore assume an average budget of $C_{\mathrm{task}}$ agentic iterations per task. To satisfy this constraint, we assign a censoring time $C_t$ for each task $t$, which limits the maximum number of iterations the agent may spend on that task. The censored event time is therefore formulated as: $\tilde{T}_t=\min\{T_t,C_t\}$. Importantly, the bound $U_t$ and the censoring time $C_t$ are deliberately distinct: when $C_t<U_t$, censoring may prevent us from determining whether the UPB $U_t$ successfully covered the true time-to-success $T_t$.

For example, consider an organization whose employees use AI agents for tasks such as software engineering. Before running the agent, an employee may want to know \emph{how many interactions it will take to complete the task successfully}. To address this, the organization can provide calibrated UPBs on the time-to-success, updated continually from the employees' own interactions as projects, users, and deployed agents change over time. However, if an employee stops before the task is completed, the time-to-success and correctness of the UPB may remain unknown. To recover this information, the organization can continue the stopped trajectory in a controlled environment, such as a sandbox, without affecting the employee's workflow or production systems. Since continuing every trajectory to completion is expensive, the organization can continue only some of them. Our framework runs on top of existing usage of the user and calibrates the UPBs from the resulting censored observations, providing UPBs with coverage guarantees under a limited budget.

Our problem is related to conformal survival analysis works~\citep{candes2023conformalized,gui2024conformalized,sesia2025conformal,sesia2024doubly, davidovconformalized}, which study how to construct calibrated predictive bounds from censored observations. However, these methods are \emph{offline}: they collect a fixed dataset of censored samples prior to calibrating the predictive bounds. In contrast, in our setting, the censoring times and predictive bounds are updated \emph{online}. For each arriving task, we must determine for how many iterations the agent should execute based on information accumulated from previous tasks and the current task $X_t$. Crucially, this decision affects not only the computational cost of the current task, but also the quality of future predictions: the resulting censored trajectory becomes part of the data used to construct subsequent upper predictive bounds. Thus, prediction and data collection are inherently coupled.

\begin{figure}[t!]
    \centering
    \begin{minipage}[t]{0.58\linewidth}
        \vspace{0pt}
        \includegraphics[width=\linewidth]{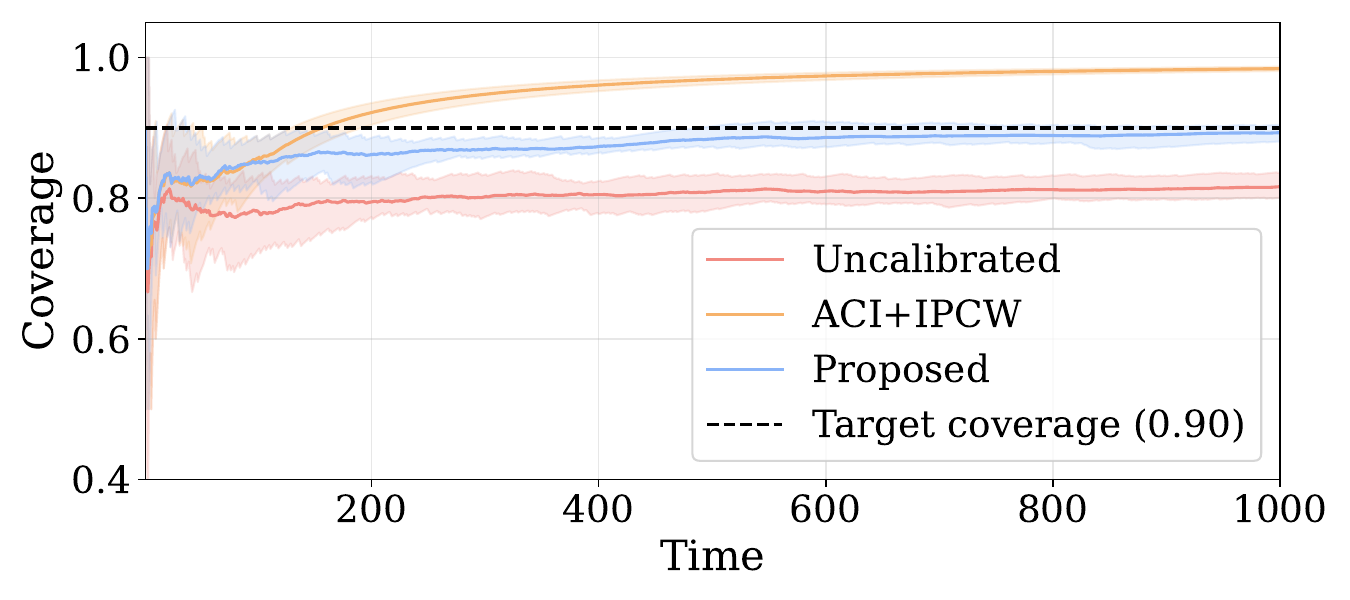}
    \end{minipage}\hfill
    \begin{minipage}[t]{0.41\linewidth}
        \vspace{0pt}
        \includegraphics[width=\linewidth]{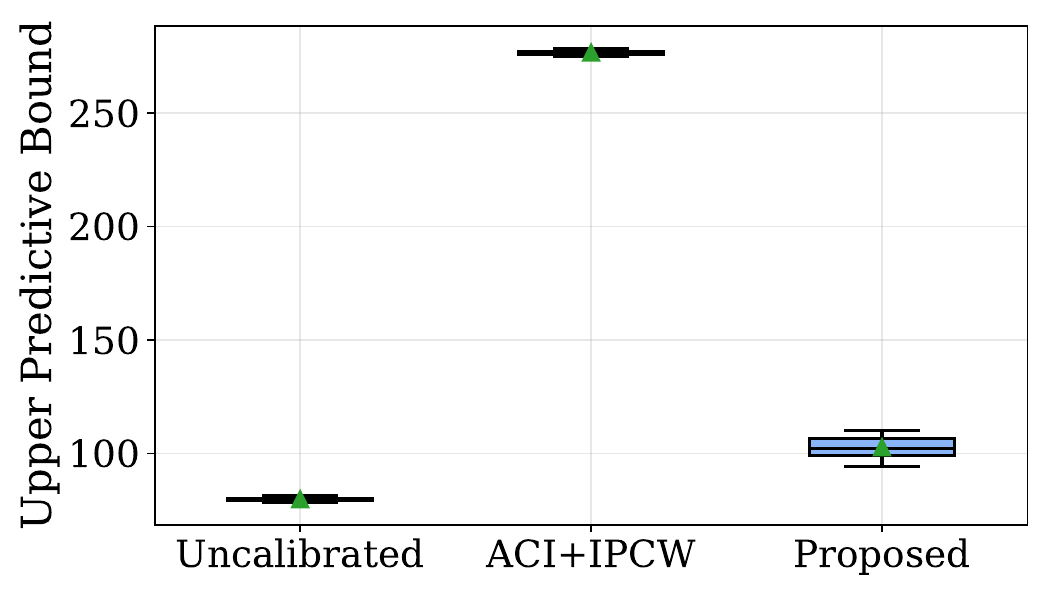}
    \end{minipage}
    \caption{Empirical evaluation of Upper Predictive Bounds on the SWE data. \textbf{Left:} Cumulative coverage rate over time comparing the \texttt{Uncalibrated} baseline, \texttt{ACI+IPCW} (Section~\ref{sec:background}), and the proposed method against a $90\%$ target coverage rate. \textbf{Right:} Boxplots of the mean UPB across repetitions for the corresponding methods (lower is better). Full experimental details are provided in Section~\ref{sec:experiments}.}
    \label{fig:coverage_over_time_nbswe}
\end{figure}

To illustrate the challenge in this setup, we conduct an experiment on the software engineering tasks ~\citep{nebius2024sweagenttrajectories}.
Figure~\ref{fig:coverage_over_time_nbswe} shows that ACI of~\citet{gibbs2021adaptive} that is adapted to account for censoring yields overly conservative upper bounds under a limited budget. Censoring can hide prediction errors, making it more challenging to maintain informative bounds from the available feedback. This motivates 
the novel adaptive strategy we propose, 
which achieves the target coverage level by coupling budget allocation with an online calibration update that is explicitly designed to maintain tight and informative bounds under censored feedback.

This raises two questions: \emph{First, how can we construct \textbf{valid} upper predictive bounds from censored online data? Second, how should we allocate the budget across tasks to derive \textbf{efficient} and \textbf{informative} bounds?}

\subsection{Our Contribution}
\label{subsec:contributions}

We introduce \emph{\fullmethodname} (\ourmethod{}), an online framework for constructing calibrated upper predictive bounds for the time-to-success of tasks under budget constraints 
(illustrated in Figure~\ref{fig:method_illustration}). 
Specifically, our main contributions are as follows:

\newpage

\begin{itemize}
    \setlength{\itemsep}{1pt}
    \setlength{\parskip}{0pt}
    \item \textbf{Censoring-aware online calibration:}
    Our method constructs UPBs from censored observations with a provably
    valid coverage bound for any fixed time horizon
    (Section~\ref{sec:method:ambiguity};
    Appendix~\ref{app:proof_probe_coverage}).
    We further identify a terminal cancellation condition under which
    this coverage certificate reaches the nominal threshold
    (Theorem ~\ref{thm:probe-coverage}).
    \vspace{0.25cm}

    \item \textbf{Coverage-informed budget allocation:}
    We formulate a constrained optimization objective for allocating budget across tasks to obtain a coverage rate closer to the target level.
    Our allocation aims to reduce future coverage error while satisfying the budget constraint
    (Section~\ref{sec:method:budget};
    Algorithm~\ref{alg:probe-budget}).
    
\end{itemize}

Code and instructions for reproducibility are provided in our \href{https://github.com/Roy-Maor-Lotan/PROBE}{GitHub}
\subsection{Problem Setup}
\label{subsec:problem_setup}

In our setup, tasks arrive sequentially over rounds $t=1,\ldots,T$. We denote by $X_t\in\mathcal X$ all information available before the $t$-th trajectory begins. In our software-engineering application, for example, $X_t$ contains the model name and problem specification of the current task, together with information about the corresponding repository.

For each task $X_t$, the agent $\mathcal A$  interacts sequentially with
a task-specific environment $\Phi_t$. We initialize the interaction
history as $H_t(0)=X_t$.
At each interaction step $k\geq 1$, the agent selects an action based on the information observed so far $A_t(k)=\mathcal A\!\left(H_t(k-1)\right)$. The environment then produces an observation in response to this action $O_t(k)=\Phi_t\!\left(X_t,H_t(k-1),A_t(k)\right)$. A task-specific auditor $\mathcal J_t$ determines whether the event of interest has occurred $k$: $Y_t(k)=
    \mathcal J_t\!\left(X_t,H_t(k)\right)
    \in\{0,1\}$.
If $Y_t(k)=1$, the event occurred and the trajectory terminates; otherwise, the interaction continues to step $k+1$. 
The history is updated as $H_t(k)= \bigl(H_t(k-1),A_t(k),O_t(k), Y_t(k)\bigr)$.
The corresponding uncensored event time is
\begin{equation}
\label{eq:event-time}
    T_t
    =
    \inf\bigl\{
        k\geq1:Y_t(k)=1
    \bigr\},
\end{equation}
with the convention $T_t=\infty$ if the event never occurs.
In our setup, the user runs the task up to time $R_t$.
After this user run, we choose a censoring time $C_t$ beyond this user runtime $R_t$, and the censored time-to-event is thus $\tilde{T}_t = \min (T_t, C_t)$. We denote by $\Delta_t=\mathbf 1\{T_t\le C_t\}$ the censoring indicator: $\Delta_t=1$ means that the event time is observed at or before the timeout, whereas $\Delta_t=0$ certifies only $T_t>C_t$. The indicator resolves equality at the boundary: $\tilde{T}_t=C_t$ with $\Delta_t=1$ means $T_t=C_t$, while the same equality with $\Delta_t=0$ means $T_t>C_t$. 
The actually consumed budget for the $t$-th task is $[\tilde{T}_t-R_t]_{+}$.
Let
\[
    \mathcal D_{t-1}
    =\{H_s(\tilde T_s),R_s,C_s,\Delta_s\}_{s<t}
\]
denote the completed observed history. 
The formal filtration convention is detailed in Appendix~\ref{app:proof_information_timing}.

The budget constraint therefore limits the expected actual interaction
cost to $C_{\mathrm{task}}$ per task at every prefix:
\begin{equation}
\label{eq:problem-budget-objective}
\begin{gathered}
    \sum_{s=1}^{t}
    \mathbb E\left[[\tilde T_s - R_s]_{+} \mid X_s, \mathcal{D}_{s-1}\right]
    \le t C_{\mathrm{task}}, \quad\forall t.
\end{gathered}
\end{equation}
In Appendix~\ref{app:deterministic-budget}, we consider an alternative requirement in which the budget is controlled deterministically rather than in expectation.

In addition to the budget constraint~\eqref{eq:problem-budget-objective}, we adopt the standard assumption that $C$ is conditionally independent of $T$ given $X,R$:
\begin{assumption}[Conditionally independent censoring]
\label{ass:probe-censoring}
For every round $t$,
$C_t\mathrel{\perp\!\!\!\perp}T_t\mid X_t, R_t, \mathcal D_{t-1}$.
\end{assumption}
Our budget allocator satisfies this condition by construction. Since we design the allocator, the distribution of $C_t \mid R_t$ is known, formalized by the conditional survival function:
\begin{equation}
\label{eq:conditional-censoring-survival-function}
    G_t(u)
    =
    \mathbb P\!\left(
        C_t\ge u
        \,\middle|\,
        X_t, R_t, \mathcal D_{t-1}
    \right).
\end{equation}

Our goal is to construct the calibrated UPB $U_t=\hat U_t(X_t)$ for the unknown time-to-event $T_t$ of an arriving task $X_t$ before executing it.
Formally, for a target miscoverage level $\alpha\in(0,1)$, this UPB should miscover $T_t$ with probability at most $\alpha$: 
\begin{equation}
\label{eq:problem-coverage-objective}
    \frac{1}{T}\sum_{t=1}^{T}m_t
    = \frac{1}{T}\sum_{t=1}^{T}\mathbb P\!\left(
        T_t>U_t
        \,\middle|\,
        X_t,\mathcal{D}_{t-1}
    \right)
    \le \alpha.
\end{equation}
Above, $m_t = \mathbb P\!\left(
        T_t>U_t
        \,\middle|\,
        X_t,\mathcal{D}_{t-1}
    \right)$ is the conditional miscoverage probability.
Our objective is therefore to construct the predictive bounds $U_t$ and
allocate censoring times $C_t$ online so that both the coverage and budget
requirements are satisfied despite observing only right-censored event times.

\section{Background}
\label{sec:background}

Adaptive conformal inference (ACI) generates predictive sets in an online manner that achieve a target miscoverage level $\alpha$ by adjusting a calibration parameter in response to past prediction errors~\citep{gibbs2021adaptive,gibbs2024conformal}.
In our setting, the UPB is
$U_t=\widehat q_t(X_t;\tau_t)$, where
$\widehat q_t(x;\tau)$ is the estimated conditional quantile at upper-tail level $1-\tau$.
ACI updates $\tau_t$ over time according to the true UPB error:
\begin{equation}
\label{eq:coverage_indicator}
e_t=\mathbf 1\{T_t>U_t\}
\end{equation}
Assuming $e_t$ is observed, ACI updates $\tau_t$ as
\begin{equation}\label{eq:tau_update}
\tau_{t+1}
=
\tau_t+\eta(\alpha-e_t),
\qquad \eta>0.
\end{equation}
Thus, when the UPB fails to cover ($e_t=1$), the update decreases $\tau_t$ and increases future UPBs, making them more conservative. When the UPB covers ($e_t=0$), it increases $\tau_t$ and turns future UPBs more liberal.~\citet{gibbs2021adaptive} show that this procedure yields UPBs that achieve the target coverage level $1-\alpha$ on average over time.

In our setup, however, observations are right-censored, so the binary error $e_t$ is not always observed. We define the observed-error and censoring-reach indicators:
\begin{equation}
\label{eq:error_indicators}
\begin{aligned}
    I_t^{\mathrm{err}}(u)
    =\mathbf{1}\{(\tilde{T}_t > u) \lor (\tilde{T}_t = u \land \Delta_t = 0)\}, \quad
    I_t^{\mathrm{reach}}(u)
    =\mathbf{1}\{C_t\ge u\}.
\end{aligned}
\end{equation}
Here, $I_t^{\mathrm{err}}(U_t)=1$ certifies observed miscoverage, $T_t>U_t$. The indicator $I_t^{\mathrm{reach}}(u)$ specifies whether the censoring time reaches $u$.

Thus, the standard ACI update in~\eqref{eq:tau_update} cannot directly use the binary error $e_t$ on every round. Treating every censored trajectory as miscover leads to overly conservative bounds, while ignoring unobserved errors leads to undercoverage.

Inverse-probability-of-censoring weighting (IPCW) corrects for the bias of the error signal induced by the missing labels. Recall that
$G_t(U_t)
=
\mathbb P(C_t\ge U_t\mid X_t, R_t,\mathcal D_{t-1})
$
is the probability that the censoring time reaches the bound.

\begin{assumption}[Censoring positivity]
\label{ass:probe-positivity}
For a fixed $g_{\min}>0$, every played round satisfies
\begin{equation}
\label{eq:censoring-design-conditions}
    G_t(U_t)\ge g_{\min}>0.
\end{equation}
\end{assumption}
Positivity ensures that the inverse-probability weights are well defined.
We define the weighted error as
\begin{equation}
\label{eq:background-ipcw-feedback}
    \widehat e_t=\frac{I_t^{\mathrm{err}}(U_t)}{G_t(U_t)}.
\end{equation}
Under Assumptions~\ref{ass:probe-censoring} and~\ref{ass:probe-positivity},
the predictable bound and censoring law yield
\begin{equation}
\label{eq:background-ipcw-identity}
    \mathbb E\!\left[
        \widehat e_t
        \,\middle|\,
        X_t,\mathcal D_{t-1}
    \right]
    =\mathbb P\!\left(
        T_t>U_t
        \,\middle|\,
        X_t,\mathcal D_{t-1}
    \right)
    =m_t.
\end{equation}
Therefore, by applying IPCW, we can obtain a miscoverage signal $\widehat e_t$ that is correct in expectation: observed miscoverages are weighted by $1/G_t(U_t)$ to compensate for failures hidden by censoring.
By replacing $e_t$ in~\eqref{eq:tau_update} with $\widehat e_t$, we obtain the following update for $\tau$
\begin{equation}
\label{eq}
\tau_{t+1}
=
\tau_t+\eta(\alpha-\widehat e_t).
\end{equation}
We refer to this method as ACI$+$IPCW; the full pseudocode is
provided in Algorithm~\ref{alg:aci_ipcw} in Appendix~\ref{app:calibration-budget-algorithms}.

Although ACI$+$IPCW attains a valid
coverage
(Theorem~\ref{thm:aci-ipcw-coverage} in Appendix~\ref{app:proof_aci_ipcw}), it can yield overly
conservative upper predictive bounds, leaving a substantial
gap between achieved coverage and the nominal level
$1-\alpha$, as illustrated in
Figure~\ref{fig:coverage_over_time_nbswe}.

\section{Proposed Method}
\label{sec:method}

In this section, we introduce \ourmethod{}, a framework that generates informative UPBs with a coverage rate closer to the target level. This framework is composed of two components. First, an online calibration mechanism (Section~\ref{sec:method:ambiguity}) that adaptively adjusts the bounds to reduce the coverage gap under censored feedback. Second, a budget allocation strategy (Section~\ref{sec:method:budget}) that aims to reduce the coverage gap as well while satisfying the expected budget constraint.

\subsection{Adaptive Calibration under Censoring}
\label{sec:method:ambiguity}

Censoring before the predictive bound $U_t$ can leave the miscoverage indicator in~\eqref{eq:coverage_indicator} unobserved. We address this missing feedback by augmenting IPCW with an estimated miscoverage probability while preserving conditional unbiasedness.

Before introducing this correction, we reparameterize the calibration parameter to enforce that $\tau_t$ remains in its admissible range. Rather than updating $\tau_t$ directly, we introduce $a_t$ and define
\begin{equation}
\label{eq:probe-parametrization}
    a_t\in[0,\infty),
    \quad
    \tau_t=\tau_{\max}e^{-a_t},
    \quad
    U_t=\widehat q_t(X_t;\tau_t),
\end{equation}
with initialization
$a_1=a_{\mathrm{init}}=\log(\tau_{\max}/\alpha)$ for
$\alpha\le\tau_{\max}$.
This parameterization guarantees that
$\tau_t\in(0,\tau_{\max}]$ for all $t$, and allows for a faster adaptation to distributional shifts~\citep{feldman2023achieving}.

Recall that $\widehat e_t=I_t^{\mathrm{err}}(U_t)/G_t(U_t)$ has conditional mean $m_t$ by~\eqref{eq:background-ipcw-identity}. Let $\widehat F_t(u)$ estimate the conditional CDF $\mathbb P(T_t\le u\mid X_t, \mathcal D_{t-1})$, and define the estimated miscoverage probability by
\begin{equation}
\label{eq:probe-dynamic-rho}
    \rho_t=1-\widehat F_t(U_t)\in[0,1].
\end{equation}
When $I_t^{\mathrm{reach}}(U_t)=1$, the coverage error is observed; under the allocation strategy in Section~\ref{sec:method:budget}, $I_t^{\mathrm{reach}}(U_t)=0$ when $C_t=0$. Under a general censoring policy, coverage may also be observed when the event occurs before a timeout below $U_t$. We use the centered corrected feedback
\begin{equation}
\label{eq:probe-corrected-feedback}
\begin{aligned}
    \widetilde e_t
    &=\widehat e_t+\rho_t\left(1-\frac{I_t^{\mathrm{reach}}(U_t)}{G_t(U_t)}\right),
    \quad\quad s_t
    =\widetilde e_t-\alpha.
\end{aligned}
\end{equation}
Because $\rho_t$ is determined by $(X_t,\mathcal D_{t-1})$ and the reach indicator has conditional mean $G_t(U_t)$, the correction has conditional mean zero. Thus $\mathbb E[\widetilde e_t\mid X_t, \mathcal D_{t-1}]=m_t$ even if $\widehat F_t$ is inaccurate. The feedback cases are
\begin{equation}
\label{eq:probe-feedback-cases}
s_t =
\begin{cases}
    \rho_t-\alpha,
    & I_t^{\mathrm{reach}}(U_t)=0,\\[0.7ex]
    \widehat e_t+\rho_t\bigl(1-G_t(U_t)^{-1}\bigr)-\alpha,
    & I_t^{\mathrm{reach}}(U_t)=1.
\end{cases}
\end{equation}
The calibration parameter is then updated using projected online gradient descent (OGD):
\begin{equation}
\label{eq:probe-corrected-step}
\begin{aligned}
    a_{t+1}
    &=\bigl[a_t+\eta s_t\bigr]_+ .
\end{aligned}
\end{equation}

Observed miscoverage gives $s_t>0$ and increases $a_t$, shifting calibration toward more conservative bounds. Observed coverage on a round that reaches $U_t$ gives $s_t<0$, pushing $a_t$ toward zero and calibration toward less conservative bounds. On rounds that do not reach $U_t$, the sign of $\rho_t-\alpha$ determines the direction of the calibration update. The correction on rounds that reach $U_t$ compensates for this imputation.

Any predictable $\rho_t\in[0,1]$ preserves unbiasedness, but its choice affects the conditional variance of the corrected feedback $\widetilde e_t$. For a fixed censoring law, this variance is minimized when $\rho_t$ equals the conditional miscoverage probability $m_t$, the optimal augmentation for inverse-probability-weighted estimators~\citep{ tsiatis2006semiparametric}. Our choice of $\rho_t$ estimates $m_t$ from the model $\widehat F_t$.

We next introduce the quantities used to bound the residual coverage gap. For each observed trajectory, define the miscoverage threshold as the supremum of the calibration states that yield an observed error:
\begin{equation}
\label{eq:probe-threshold}
\begin{gathered}
    b_t=\sup\!\left\{
    a\ge0:
    I_t^{\mathrm{err}}\!\left(\widehat q_t(X_t;\tau_{\max}e^{-a})\right)=1
    \right\},\\
    \sup\varnothing=0.
\end{gathered}
\end{equation}
We track the cumulative centered correction $A_t^\rho$, the threshold-weighted sum $L_t$, and the sum of squared update signals $Q_t$, initialized at $A_0^\rho=L_0=Q_0=0$:
\begin{align}
\label{eq:probe-candidate-prefix-state}
    A_t^\rho
    &=A_{t-1}^\rho+\rho_t(1-I_t^{\mathrm{reach}}(U_t)/G_t(U_t)),
      \nonumber\\
    L_t
    &=L_{t-1}
      +\widehat e_tb_t+a_t(\rho_t-\alpha)_+(1-I_t^{\mathrm{reach}}(U_t)),
      \\
    Q_t
    &=Q_{t-1}
      +s_t^2. \nonumber
\end{align}

The OGD analysis combines $L_t$ and $Q_t$ into the envelope:
\begin{equation}
\label{eq:probe-gamma-candidate}
\begin{aligned}
    &\Gamma_t
    =\frac{1}{\eta}\Bigl[-a_{\mathrm{init}}+\sqrt{a_{\mathrm{init}}^2+
    \eta\{2L_t+\eta Q_t\}}\Bigr].
\end{aligned}
\end{equation}
The quantity $\Gamma_t-A_t^\rho$ bounds the cumulative weighted excess miscoverage $\sum_{s=1}^t(\widehat e_s-\alpha)$.

To account for stochastic variation in the IPCW feedback, let
\[
    V_t=\sum_{s=1}^t\left\{
    \mathbb E\!\left[
        \widehat e_s^2
        \,\middle|\,
        X_s,\mathcal D_{s-1}
    \right]-m_s^2\right\}
\]
denote its cumulative conditional variance. Since $V_t$
depends on unknown conditional moments, we use the
deterministic upper bound $V_t\le t/g_{\min}$ to define
the computable confidence buffer
\begin{equation}
\label{eq:probe-freedman-buffer}
    \mathfrak F_t(\delta)
    =
    \sqrt{\frac{2t}{g_{\min}}\log\frac1\delta}
    +\frac13\log\frac1\delta.
\end{equation}
This term captures the uncertainty underlying the IPCW feedback.

Appendix~\ref{app:proof_probe_coverage} provides the complete derivation. Theorem~\ref{thm:probe-coverage} bounds the residual coverage gap through $\Gamma_t$, the confidence buffer $\mathfrak F_t(\delta_{\mathrm{cov}})$, and the signed centered correction $A_t^\rho$.

Algorithm~\ref{alg:probe-calibration} summarizes the calibration procedure. The next subsection uses the resulting coverage bound to guide observation probabilities under the budget constraint.

\subsection{Coverage-Informed Budget Allocation}
\label{sec:method:budget}
We adapt the budget allocation of~\citet{davidov2026calibrated}
to our online setting. Specifically, for each round $t$, we choose a probability $\pi_t\in[g_{\min},1]$ of observing the trajectory up to iteration $C_t=u_t:=\lceil U_t\rceil$. With the remaining probability $1-\pi_t$, we set $C_t=0$ and do not run the trajectory.
Formally,
\begin{equation}
\label{eq:probe-two-point-law}
    C_t=
    \begin{cases}
        u_t, & \text{with probability }\pi_t,\\
        0,   & \text{with probability }1-\pi_t .
    \end{cases}
\end{equation}
Observe that for this design choice $G_t(U_t)=\pi_t$, the IPCW weight is $W_t=1/\pi_t$, and the expected consumed budget for this step is at most $\pi_t u_t$.
With fresh randomization independent of $T_t$ conditional on $(X_t, R_t,\mathcal D_{t-1})$, this allocation satisfies Assumptions~\ref{ass:probe-censoring} and~\ref{ass:probe-positivity} by construction, since $G_t(U_t)=\pi_t\ge g_{\min}>0$.
We select $\pi_t$ by minimizing a model-based estimate of the residual coverage deficit subject to the budget constraint. The feasible observation probabilities are
\begin{equation}
\label{eq:probe-budget-feasible-set}
    \Pi_t^{\mathrm{bud}}
    =
    \left\{\pi:
    \sum_{s=1}^{t-1}[\pi_su_s - R_s]_{+}+[\pi u_t - R_t]_{+}
    \le tC_{\mathrm{task}}\right\}.
\end{equation}
Unused allowance under this cumulative constraint can support larger expected allocations on later rounds, allowing $\pi_tu_t$ to exceed $C_{\mathrm{task}}$ while every prefix remains within budget.

Our budget allocation is guided by the same coverage gap as the
calibration scheme. Among feasible allocations, we seek to reduce the
anticipated deficit
$\Gamma_t+\mathfrak F_t(\delta_{\mathrm{cov}})-A_t^\rho$
in Theorem~\ref{thm:probe-coverage}. A candidate probability $\pi$
determines both the frequency of observing coverage feedback and
its inverse-probability weight $W=1/\pi$. These, in turn, affect
the calibration states and the resulting coverage deficit.

Before observing the current outcome, we evaluate each candidate allocation using the same completion model $\widehat F_t$ used to define $\rho_t$. Let
$\widehat{\mathbb E}_{t,W}$ denote 
estimated expectation over the predicted
event time and the candidate censoring draw. The anticipated
deficit is
\begin{equation}
\label{eq:probe-anticipated-deficit}
\begin{aligned}
    \overline{\mathcal E}_t(W)
    &=\frac{1}{t}\widehat{\mathbb E}_{t,W}\!\Bigl[
    \Gamma_t(W)
    +\widehat{\mathfrak F}_t(\delta_{\mathrm{cov}},W)
    -A_t^\rho(W)
    \Bigr].
\end{aligned}
\end{equation}
Here, $\Gamma_t(W)$ and $A_t^\rho(W)$ denote the calibration envelope and cumulative centered correction after the synthetic update under candidate weight $W$. 
The term $\widehat{\mathfrak F}_t(\delta_{\mathrm{cov}},W)$ is a model-based estimate of the confidence buffer $\mathfrak F_t(\delta_{\mathrm{cov}})$.
For each candidate $W$, the synthetic update uses the corresponding censoring law and IPCW weight, while keeping $U_t$, $\rho_t$, and the completed-round states fixed.
Algorithm~\ref{alg:probe-budget} specifies the synthetic updates and the model-based buffer in~\eqref{eq:probe-anticipated-deficit}.

When the feasible set is nonempty, we select the observation probability by solving
\begin{equation}
\label{eq:probe-budget-optimizer}
    \pi_t\in
    \arg\min_{\pi\in\Pi_t^{\mathrm{bud}}}
    \overline{\mathcal E}_t(1/\pi).
\end{equation}
After selecting $\pi_t$, we draw $C_t$ according to~\eqref{eq:probe-two-point-law} and run the trajectory until success or timeout, obtaining $(\tilde T_t, R_t, \Delta_t)$.

The complete \ourmethod{} procedure, combining online calibration
and budget allocation, is summarized in Algorithm~\ref{alg:probe-full}
in Appendix~\ref{app:calibration-budget-algorithms}.
The next subsection establishes the coverage and expected-budget guarantees.

\begin{figure}[t!]
    \centering
    \begin{minipage}{0.49\textwidth}
        \centering
        \includegraphics[width=\linewidth]{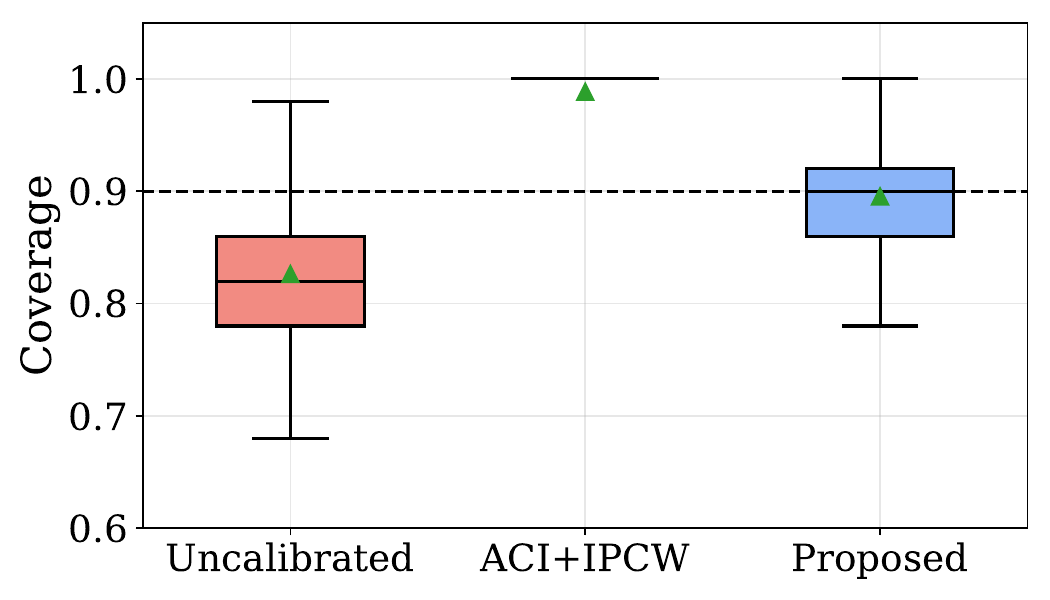}
        
        \vspace{2pt}
        \quad\qquad(a)
        \vspace{3pt}
    \end{minipage}\hfill
    \begin{minipage}{0.49\textwidth}
        \centering
        \includegraphics[width=\linewidth]{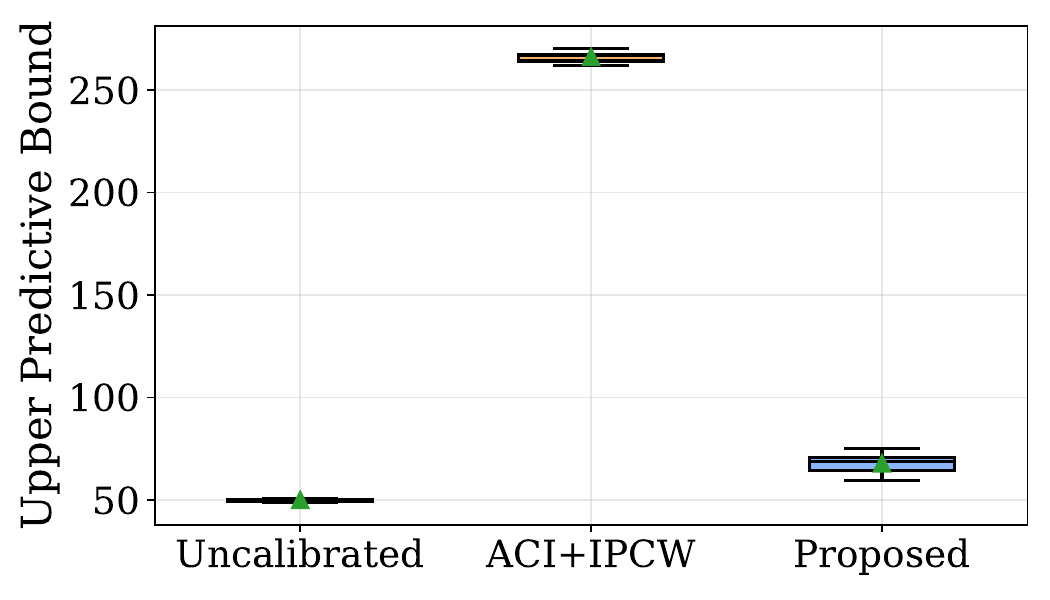}
        
        \vspace{2pt}
        \quad\qquad(b)
        \vspace{2pt}
    \end{minipage}
    
    \caption{\textbf{RAG setting:} 
    Results after 1,000 steps across 10 repetitions for $\alpha=0.10$ and budget constraint $C_{\mathrm{task}}=25$, comparing the \texttt{Uncalibrated}, \texttt{ACI+IPCW}, and Proposed methods. (a) Boxplots of empirical local coverage over window~$=50$; the dashed line marks the target coverage level. (b) Boxplots of the mean upper predictive bound within each repetition.}
    \label{fig:RAG:bxp_coverage_and_upb}
\end{figure}

\subsection{Theoretical guarantees}
\label{sec:method:guarantees}

We establish a bound on average conditional coverage and a bound on average expected resource use, using the quantities defined above.
Let
$A_T^\rho=\sum_{t=1}^T\rho_t(1-I_t^{\mathrm{reach}}(U_t)/G_t(U_t))$,
and let $L_T$ and $Q_T$ denote the terminal realized states
obtained by evaluating the recurrences in
\eqref{eq:probe-candidate-prefix-state} on each round
$t=1,\ldots,T$, with $\rho_t$ defined by~\eqref{eq:probe-dynamic-rho}. Evaluating the previously defined envelope on the observed sequence at horizon $T$ gives
\begin{equation}
\label{eq:probe-realized-gamma}
    \begin{aligned}
    &\Gamma_T
    =\frac{1}{\eta}\Bigl[-a_{\mathrm{init}}+\sqrt{a_{\mathrm{init}}^2+
    \eta\{2L_T+\eta Q_T\}}\Bigr]
\end{aligned}.
\end{equation}

\begin{theorem}[Coverage validity, informal]
\label{thm:probe-coverage}
Under Assumptions~\ref{ass:probe-censoring}, and~\ref{ass:probe-positivity}
\ourmethod{}
\eqref{eq:probe-parametrization}--\eqref{eq:probe-corrected-step}
satisfies, for every fixed $T$ and $\delta_{\mathrm{cov}}\in(0,1)$, with
probability at least $1-\delta_{\mathrm{cov}}$ over $\mathcal D_T$ 

\begin{equation}
\label{eq:probe-residual-coverage}
\begin{aligned}
    &\frac1T\sum_{t=1}^T\mathbb P\!\left(
        T_t\le U_t
        \,\middle|\,
        X_t,\mathcal D_{t-1}
    \right)
    \ge1-\alpha+
    \frac{1}{T}\Bigl[
    A_T^\rho
    - \Gamma_T
    - \mathfrak F_T(\delta_{\mathrm{cov}})
    \Bigr].
\end{aligned}
\end{equation}
\end{theorem}

The proof is provided in Appendix~\ref{app:proof_probe_coverage}.
The bound controls average conditional coverage under adaptive censoring.

\begin{theorem}[Budget validity, informal]
\label{thm:probe-budget-validity}
Under Assumptions~\ref{ass:probe-censoring}, \ref{ass:probe-positivity},
and~\ref{ass:probe-budget},
\ourmethod{} (Algorithm~\ref{alg:probe-full}) satisfies, for every
fixed horizon $T$,
\begin{equation}
\label{eq:probe-expected-budget-main}
    \frac1T\sum_{t=1}^T
    \mathbb E[\tilde{T}_t-R_t]_{+}
    \le C_{\mathrm{task}}.
\end{equation}
\end{theorem}

This theorem is formally proved in Appendix~\ref{app:probe-budget-proof}.
The accumulating constraint limits average expected additional resource consumption
even when the completion model is misspecified.
A corresponding high-probability bound for the realized budget, holding
jointly with the residual coverage certificate, is established in
Theorem~\ref{thm:high-prob-budget} in
Appendix~\ref{app:probe-joint-proof}.

\section{Experiments}
\label{sec:experiments}
We evaluate our proposed framework, \ourmethod{}, across three distinct real-world settings with right-censored outcomes. Adapting the evaluation procedures established by~\citet{davidov2026calibrated,feldman2026} to our online setting, we present tasks sequentially and censor outcomes using finite per-task budgets to assess both empirical coverage and resource use. We compare \ourmethod{} with two baselines: \texttt{Uncalibrated}, which reports raw model predictions, and \texttt{ACI+IPCW}, which applies the censoring-corrected ACI update (Section~\ref{sec:background}) paired with neutral budget allocation (Algorithm~\ref{alg:neutral-budget-allocation}). We first describe the three experimental settings and then examine calibration, informativeness, and budget dynamics.

In the first setting, SWE, we construct UPBs for the interaction lengths of SWE-agent trajectories addressing real-world GitHub issues ~\citep{nebius2024sweagenttrajectories}. The outcome for each task is the number of logged agent interactions in its trajectory. 

Next, in the RAG setting, we consider medical information-retrieval queries from NFCorpus~\citep{thakur2021beir} and construct UPBs for the number of retrieval iterations required to identify all relevant corpus items associated with each query.

In the instruction-following setup, we use tasks from AutoIF~\citep{dong2025selfplay}. 
Here, the agent aims to help a target model satisfy the requirements across sequential iterations until all instructions are satisfied. The response is verified programmatically using Python scripts. Full details regarding all datasets, feature extractions, and quantile predictor architectures are provided in Appendix~\ref{app:exp_details}

To demonstrate the effectiveness of our budget allocation, we consider the challenging setting where we receive no prior user interactions, $R_t=0, \forall t$, and must allocate the entire evaluation budget across tasks.
Appendix~\ref{app:exp_details} details all datasets, features, and model architectures.

As illustrated in Figure~\ref{fig:coverage_over_time_nbswe}, \ourmethod{} achieves the target coverage rate and constructs more informative bounds compared to the \texttt{ACI+IPCW} method, which tends to overcover the response. Furthermore, this figure shows that the \texttt{Uncalibrated} predictor consistently undercovers the true event time.

The same trend is observed in the RAG setting presented in Figure~\ref{fig:RAG:bxp_coverage_and_upb}. Specifically, the \texttt{Uncalibrated} predictor systematically undercovers the time-to-success, whereas \texttt{ACI+IPCW} attains coverage higher than the target level and yields large and less informative UPBs. Our proposed method achieves coverage close to the nominal level with substantially tighter bounds than \texttt{ACI+IPCW}. This demonstrates the effectiveness of our method in accounting for censored feedback to restore reliability while preserving highly informative predictions.

\begin{figure}[t!]
    \centering
    \begin{minipage}[t]{0.48\linewidth}
        \vspace{0pt}
    \centering

\includegraphics[trim={0.45cm 0cm 0.45cm 0cm}, clip, width=\linewidth]{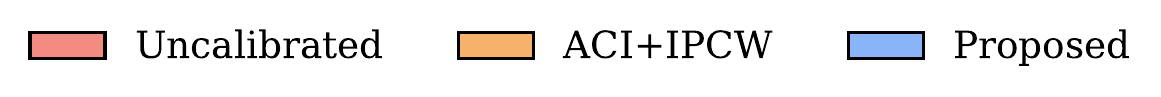}\\
\vspace{-0.15cm}    
\includegraphics[width=\linewidth]{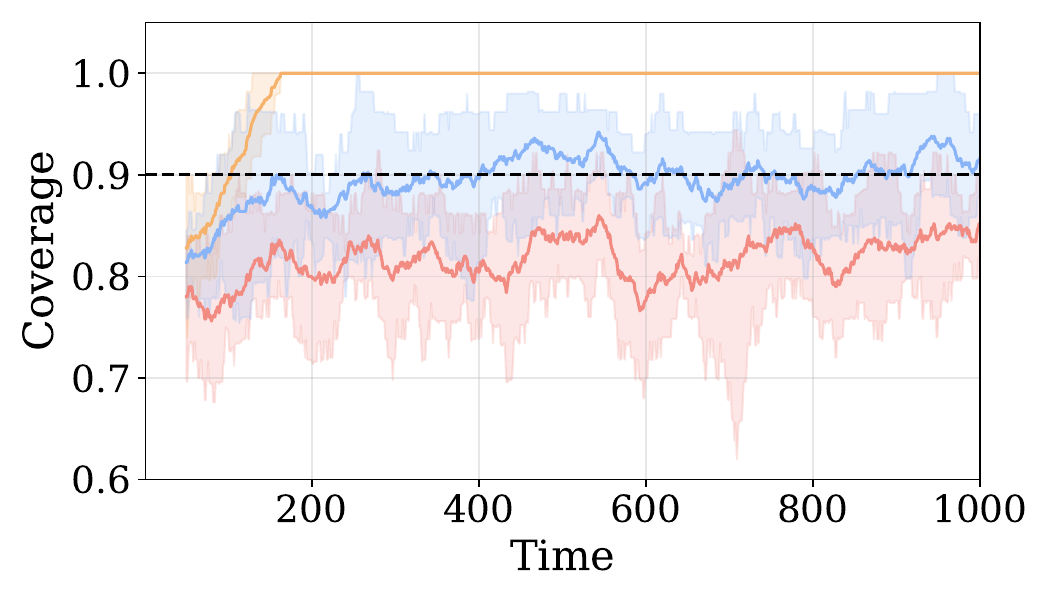}
    \caption{\textbf{SWE setting:} Empirical coverage over rolling windows of 50 tasks, for $\alpha=0.10$ and $C_{\mathrm{task}}=20$, comparing \texttt{Uncalibrated}, \texttt{ACI+IPCW}, and \ourmethod{}. The dashed line marks the target $90\%$ coverage.}
    \label{fig:SWE:local_coverage}
    \end{minipage}\hfill
    \begin{minipage}[t]{0.48\linewidth}
        \vspace{0pt}
        \centering
\includegraphics[width=\linewidth]{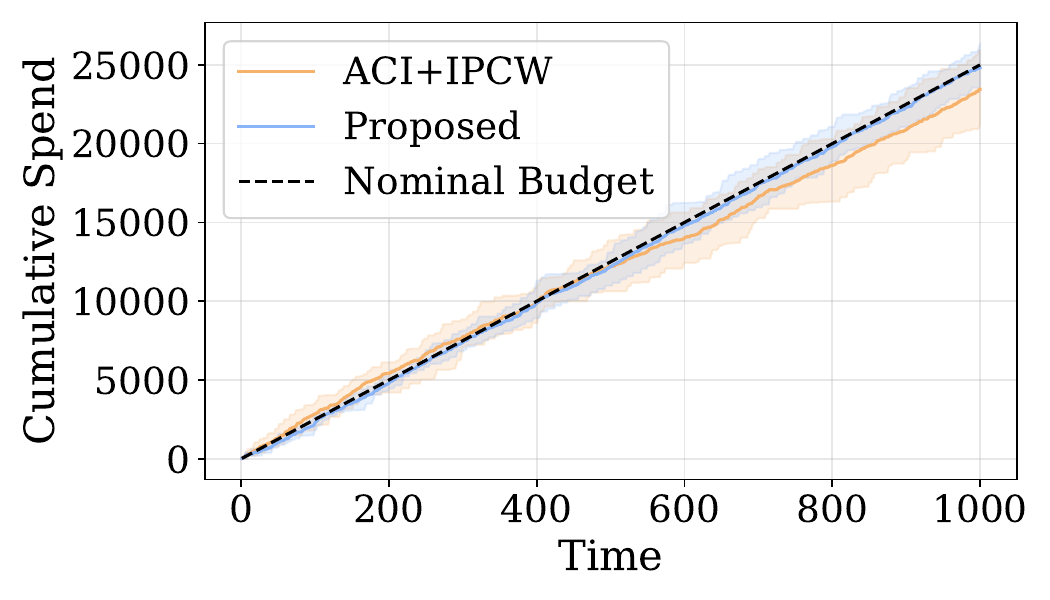}
\caption{\textbf{AutoIF settings:} Mean cumulative realized spend $\sum_{s\leq t}C_s$ at step $t$, for $\alpha=0.1$ and budget constraint $C_{\mathrm{task}}=25$, comparing the \texttt{ACI+IPCW} and Proposed methods over 1,000 steps and 10 repetitions. 
The dashed line marks the nominal budget $tC_{\mathrm{task}}$.}
\label{fig:AutoIF:cumulative_spend}
    \end{minipage}
\end{figure}

We measure the local coverage of our method on the SWE data, and report it in Figure~\ref{fig:SWE:local_coverage}. Specifically, the \texttt{Uncalibrated} predictor severely undercovers, whereas \texttt{ACI+IPCW} overcovers the response. Conversely, the coverage rate attained by \ourmethod{} across moving windows of time is closer to the desired level.

Finally, Figure~\ref{fig:AutoIF:cumulative_spend} demonstrates efficient budget utilization in the AutoIF setting through the cumulative assigned timeout, $\sum_{s\leq t}C_s$. Specifically, both methods sustain continuous allocation throughout the stream, with their mean trajectories closely tracking the nominal allowance $tC_{\mathrm{task}}$. 
For \ourmethod{}, this stable tracking not only supports Theorem~\ref{thm:probe-budget-validity} but also confirms that our accumulating-budget mechanism maintains steady while efficiently utilizing the available resource allowance.

Appendix~\ref{app:additional_experiments} provides additional experiments across the SWE, RAG, and AutoIF datasets.

\section{Discussion}
\label{sec:discussion}

We introduced \ourmethod{}, an online framework for constructing calibrated upper predictive bounds on time-to-event under limited computational resources. The key challenge is that censoring affects both calibration and data collection: when a trajectory is stopped early, the coverage error may remain unobserved, while additional computation spent on one task reduces the budget available for others. \ourmethod{} addresses these two issues jointly by correcting for censored feedback and allocating budget aiming to minimize the coverage gap. Our experiments show that our approach leads to informative and reliable bounds across several agentic evaluation settings.

Our coverage guarantee holds on average over the task stream rather than for each task or subgroup.
An exciting future direction could be extending the framework to individualized or group-conditional coverage, possibly by drawing ideas from~\citep{feldman2021improving, gibbs2025conformal}. Second, we assume that the auditor identifying the event of interest is reliable; in practice, unit tests, automated checkers, and LLM judges may themselves be noisy or incorrect. An important direction is to develop a framework that is robust to such incorrect labels~\citep{einbinder2024label, penso2024conformal}, e.g., by leveraging privileged information~\citep{vapnik2009new,feldman2024robustconformalpredictionusing, feldman2026conformal}. We believe using simulated data to complement censored trajectories could improve the prediction efficiency, making this a promising research direction~\citep{bashari2026general, bashari2026synthetic}.

Finally, our framework can support both utility and safety evaluation. Time-to-event bounds can characterize how efficiently an agent completes tasks, as well as how many iterations are required for undesirable events---such as unsafe behavior or policy violations---to occur.
Reliable evaluation under limited computation can make large-scale auditing more practical and help identify systems or task classes that warrant additional scrutiny. At the same time, the guarantees provided by \ourmethod{} consider the specified event, auditor, and coverage criterion; they should not be interpreted as a complete certificate of system reliability or safety. Thus, our framework should incorporate complementary evaluations of correctness, robustness, and security in deployment pipelines.

\section*{Acknowledgments}
Y.~R., R.~M.~L  and S.~F. were supported by the European Union (ERC, SafetyBounds, 101163414). Views and opinions expressed are however those of the authors only and do not necessarily reflect those of the European Union or the European Research Council Executive Agency. Neither the European Union nor the granting authority can be held responsible for them.
Y.R. thanks the Career Advancement Fellowship, Technion.






\bibliographystyle{plainnat}
\bibliography{ref}

\clearpage
\appendix
\thispagestyle{empty}
\onecolumn
\aistatstitle{How Efficient Is Your Agent?\\ Online Predictive Bounds under Budget Constraints: \\
Supplementary Materials}

\section{Algorithms}
\label{app:algorithms}
\subsection{Calibration and Budget Allocation}
\label{app:calibration-budget-algorithms}

The additional-cost feasible set used by the budget allocator is
\begin{equation}
\label{eq:appendix-budget-feasible-set}
    \Pi_t^{\mathrm{add}}
    =\begin{cases}
      \{\pi\in[g_{\min},1]: S_{t-1}+\pi d_t\le tC_{\mathrm{task}}\},
        & d_t>0,\\
      \{1\}, & d_t=0,
    \end{cases}
\end{equation}
where $d_t=(\lceil U_t\rceil-R_t)_+$ and
$S_{t-1}=\sum_{s<t}\pi_sd_s$. The zero-cost case uses $\pi_t=1$ by convention.

\begin{algorithm}[H]
\caption{\texttt{ACI+IPCW}}
\label{alg:aci_ipcw}
\small
\begin{algorithmic}[1]
\Require Horizon $T$, target $\alpha\in(0,1)$, step size $\eta>0$,
initial calibration state $\tau_1$; prediction rules
$\widehat q_t(x;\tau)$ with declared behavior for every reachable
$\tau$; a censoring-policy rule based on $(X_t,R_t,\mathcal D_{t-1})$
with known survival probability
$G_t(v)=\mathbb P(C_t\ge v\mid X_t,R_t,\mathcal D_{t-1})$ and
$G_t(U_t)\ge g_{\min}>0$.
\State Initialize the completed online history $\mathcal D_0\gets\varnothing$.
\For{$t=1,\ldots,T$}
    \State Receive $X_t$; fix the prediction rule using only
    $(X_t,\mathcal D_{t-1})$ and issue
    $U_t\gets\widehat q_t(X_t;\tau_t)$.
    \State Observe the user's initial run of length $R_t$.
    Commit the censoring law supported on $C_t\ge R_t$ and record $G_t(U_t)$;
    draw $C_t$ using fresh randomness, conditionally independent of $T_t$
    given $(X_t,R_t,\mathcal D_{t-1})$.
    \State Continue any unfinished trajectory until the event or timeout; observe
    $\tilde T_t=\min\{T_t,C_t\}$ and
    $\Delta_t=\mathbf{1}\{T_t\le C_t\}$, with an event at the timeout
    counted as observed.
    \State Compute the observable IPCW feedback and update:
    \[
        \widehat e_t\gets
        \frac{\mathbf{1}\{(\tilde{T}_t>U_t)
        \lor (\tilde{T}_t=U_t\land\Delta_t=0)\}}{G_t(U_t)},
        \qquad
        \tau_{t+1}\gets\tau_t+\eta(\alpha-\widehat e_t).
    \]
    \State Append $(H_t(\tilde T_t),R_t,C_t,\Delta_t)$ to
    $\mathcal D_{t-1}$ to obtain $\mathcal D_t$.
\EndFor
\State \Return issued bounds $(U_t)_{t=1}^{T}$ and state $\tau_{T+1}$.
\end{algorithmic}
\end{algorithm}

\begin{algorithm}[H]
\caption{Neutral Budget Allocation (Online version of ~\cite{davidov2026calibrated})}
\label{alg:neutral-budget-allocation}
\small
\begin{algorithmic}[1]
\Require Round $t$, user runtime $R_t$, finite bound $U_t>0$, budget rate
$C_{\mathrm{task}}>0$, positivity floor $g_{\min}\in(0,1]$,
and expected allocation $S_{t-1}=\sum_{s<t}\pi_sd_s$ (initialize $S_0=0$).
\State Set $u_t\gets\lceil U_t\rceil$ and $d_t\gets(u_t-R_t)_+$.
\If{$d_t=0$}
    \State \Return $(C_t,\pi_t,W_t,d_t,S_t)=(R_t,1,1,0,S_{t-1})$.
\EndIf
\State Set $\overline\pi_t\gets
\min\{1,(tC_{\mathrm{task}}-S_{t-1})/d_t\}$.
\If{$\overline\pi_t<g_{\min}$}
    \State \Return \textsc{infeasible}, leaving the allocation state unchanged.
\EndIf
\State Set $\pi_t\gets\overline\pi_t$ and $W_t\gets1/\pi_t$.
\State Draw $Z_t\sim\operatorname{Bernoulli}(\pi_t)$ using fresh randomness
conditionally independent of $T_t$ given $(X_t,R_t,\mathcal D_{t-1})$.
\State Set $C_t\gets R_t+d_tZ_t$ and $S_t\gets S_{t-1}+\pi_td_t$.
\State \Return $(C_t,\pi_t,W_t,d_t,S_t)$.
\end{algorithmic}
\end{algorithm}

\begin{algorithm}[H]
\caption{Censoring-Aware Online Calibration}
\label{alg:probe-calibration}
\small
\begin{algorithmic}[1]
\Require $T$, $0<\alpha\le\tau_{\max}\le1$ with $\alpha<1$,
$\eta>0$, $\delta_{\mathrm{cov}}\in(0,1)$, $g_{\min}\in(0,1]$;
predictable upper-quantile rules $\widehat q_t(x;\tau)$ nonincreasing
in $\tau$ with finite thresholds $b_t$; predictable completion CDFs
$\widehat F_t$; an external predictable
censoring policy with known $G_t(U_t)\ge g_{\min}$ and fresh
randomness conditionally independent of $T_t$ given
$(X_t,R_t,\mathcal D_{t-1})$.
\State Initialize $\mathcal D_0\gets\varnothing$,
$a_{\mathrm{init}}\gets\log(\tau_{\max}/\alpha)$,
$a_1\gets a_{\mathrm{init}}$, and $A_0^\rho=L_0=Q_0\gets0$.
\For{$t=1,\ldots,T$}
    \State Receive $X_t$ and fix $(\widehat q_t,\widehat F_t)$ using
    $(X_t,\mathcal D_{t-1})$; set
    $\tau_t\gets\tau_{\max}e^{-a_t}$ and issue
    $U_t\gets\widehat q_t(X_t;\tau_t)$; set
    $\rho_t\gets1-\widehat F_t(U_t)$.
    \State Observe the user's initial run of length $R_t$.
    Commit the external censoring law supported on $C_t\ge R_t$, record $G_t(U_t)$,
    and draw $C_t$ from that law.
    \State Continue any unfinished trajectory until the event or timeout; observe
    $\tilde T_t=\min\{T_t,C_t\}$ and
    $\Delta_t=\mathbf{1}\{T_t\le C_t\}$.
    \State Define, for any threshold $v$,
    \[
    \begin{aligned}
        I_t^{\mathrm{err}}(v)&\gets
        \mathbf{1}\{(\tilde{T}_t>v)
        \lor (\tilde{T}_t=v\land\Delta_t=0)\},\\
        I_t^{\mathrm{reach}}(v)&\gets\mathbf 1\{C_t\ge v\}.
    \end{aligned}
    \]
    \State Compute
    \[
        \widehat e_t\gets\frac{I_t^{\mathrm{err}}(U_t)}{G_t(U_t)},
        \qquad
        b_t\gets\sup\{a\ge0:
        I_t^{\mathrm{err}}(\widehat q_t(X_t;\tau_{\max}e^{-a}))=1\},
        \quad \sup\varnothing=0.
    \]
    \State Commit the centered update and realized states:
    \[
    \begin{aligned}
        \widetilde e_t&\gets\widehat e_t+
          \rho_t(1-I_t^{\mathrm{reach}}(U_t)/G_t(U_t)),
        &s_t&\gets\widetilde e_t-\alpha,\\
        A_t^\rho&\gets A_{t-1}^\rho+\rho_t(1-I_t^{\mathrm{reach}}(U_t)/G_t(U_t)),
        &L_t&\gets L_{t-1}+\widehat e_tb_t
              +a_t(\rho_t-\alpha)_+(1-I_t^{\mathrm{reach}}(U_t)),\\
        Q_t&\gets Q_{t-1}+s_t^2,
        &a_{t+1}&\gets\max\{0,a_t+\eta s_t\}.
    \end{aligned}
    \]
    \State Compute $\Gamma_t$ by~\eqref{eq:probe-gamma-candidate}
    and the distribution-free buffer
    $\mathfrak F_t(\delta_{\mathrm{cov}})$ by~\eqref{eq:probe-freedman-buffer}.
    \State Append $(H_t(\tilde T_t),R_t,C_t,\Delta_t)$ to
    $\mathcal D_{t-1}$ to obtain $\mathcal D_t$.
\EndFor
\State \Return issued bounds $(U_t)_{t=1}^{T}$ and the final calibration states.
\end{algorithmic}
\end{algorithm}

\begin{algorithm}[H]
\caption{Censoring-Aware Budget Allocation}
\label{alg:probe-budget}
\small
\begin{algorithmic}[1]
\Require Round $t$, context $X_t$, user runtime $R_t$, finite bound $U_t>0$,
current quantile family $\widehat q_t$, pre-run completion CDF
$\widehat F_t$; $\alpha,\tau_{\max},\eta,a_t,a_{\mathrm{init}},
\delta_{\mathrm{cov}},g_{\min},C_{\mathrm{task}}$;
completed states $(A_{t-1}^\rho,L_{t-1},Q_{t-1},\widehat V_{t-1})$
and expected allocation $S_{t-1}=\sum_{s<t}\pi_sd_s$.
\Require Finite synthetic thresholds; an objective satisfying
Assumption~\ref{ass:probe-budget}, with attained scalar minima.
\State Set $u_t\gets\lceil U_t\rceil$ and $d_t\gets(u_t-R_t)_+$.
\If{$d_t=0$}
    \State \Return $(C_t,\pi_t,W_t,d_t)=(R_t,1,1,0)$.
\EndIf
\State Set $\overline\pi_t\gets
\min\{1,(tC_{\mathrm{task}}-S_{t-1})/d_t\}$.
\If{$\overline\pi_t<g_{\min}$}
    \State \Return \textsc{infeasible}.
\EndIf
\State Set $\Pi_t^{\mathrm{add}}\gets[g_{\min},\overline\pi_t]$
and $\widehat m_t\gets1-\widehat F_t(U_t)$;
hold $\rho_t\gets\widehat m_t$ fixed for all candidates.
\State Define the following objective for each candidate
$\pi\in\Pi_t^{\mathrm{add}}$, with $W=1/\pi$; keep all input states fixed.
\State Under the synthetic law $\widehat{\mathbb P}_{t,W}$, let
$T_t^\ast\sim\widehat F_t$ and, independently,
$C_t^\ast=R_t+d_t$ with probability $1/W$, and $C_t^\ast=R_t$ otherwise.
Set $\tilde T_t^\ast=\min\{T_t^\ast,C_t^\ast\}$,
$\Delta_t^\ast=\mathbf{1}\{T_t^\ast\le C_t^\ast\}$, and
$Z_t^\ast=(C_t^\ast,\tilde T_t^\ast,\Delta_t^\ast)$.
\State For a synthetic observation $z=(c,y,d)$, define
\[
\begin{aligned}
    i_z^{\mathrm{err}}(v)&=\mathbf{1}\{(y>v)\lor(y=v\land d=0)\},
    &i_z^{\mathrm{reach}}(v)&=\mathbf 1\{c\ge v\},\\
    \widehat e_t(W;z)&=W i_z^{\mathrm{err}}(U_t),
    &b_t(z)&=\sup\{a\ge0:
      i_z^{\mathrm{err}}(\widehat q_t(X_t;\tau_{\max}e^{-a}))=1\},
\end{aligned}
\]
with $\sup\varnothing=0$.
\State Form temporary candidate states with the fixed $\rho_t$:
\[
\begin{aligned}
    s_t^*(W;z)&=\widehat e_t(W;z)+\rho_t(1-Wi_z^{\mathrm{reach}}(U_t))-\alpha,\\
    A_t^\rho(W;z)&=A_{t-1}^\rho+\rho_t(1-Wi_z^{\mathrm{reach}}(U_t)),\\
    L_t(W;z)&=L_{t-1}+\widehat e_t(W;z)b_t(z)
      +a_t(\rho_t-\alpha)_+(1-i_z^{\mathrm{reach}}(U_t)),\\
    Q_t(W;z)&=Q_{t-1}+\{s_t^*(W;z)\}^2,\\
    \Gamma_t(W;z)&=
      \frac{-a_{\mathrm{init}}+
      \sqrt{a_{\mathrm{init}}^2+
      \eta\{2L_t(W;z)+\eta Q_t(W;z)\}}}{\eta}.
\end{aligned}
\]
\State Compute the candidate model-based buffer:
\[
    \widehat V_t(W)=\widehat V_{t-1}+\widehat m_tW-\widehat m_t^2,
    \qquad
    \widehat{\mathfrak F}_t(\delta_{\mathrm{cov}},W)=
    \sqrt{2\widehat V_t(W)\log\frac1{\delta_{\mathrm{cov}}}}
    +\frac13\log\frac1{\delta_{\mathrm{cov}}}.
\]
\State Average over both synthetic draws:
\[
    \overline{\mathcal E}_t(W)=\frac1t
    \widehat{\mathbb E}_{t,W}\!\left[
    \Gamma_t(W;Z_t^\ast)
    +\widehat{\mathfrak F}_t(\delta_{\mathrm{cov}},W)
    -A_t^\rho(W;Z_t^\ast)\right].
\]
\State Select $\pi_t\gets
\min\operatorname*{arg\,min}_{\pi\in\Pi_t^{\mathrm{add}}}
\overline{\mathcal E}_t(1/\pi)$ and set $W_t\gets1/\pi_t$.
\State Draw $C_t=R_t+d_t$ with probability $\pi_t$, and $C_t=R_t$ otherwise,
using fresh randomness conditionally independent of $T_t$ given
$(X_t,R_t,\mathcal D_{t-1})$;
record $G_t(U_t)=\pi_t$.
\State \Return $(C_t,\pi_t,W_t,d_t)$; discard all synthetic states.
\end{algorithmic}
\end{algorithm}

The synthetic expectation and $\widehat V_t$ in
Algorithm~\ref{alg:probe-budget} are working-model surrogates evaluated
at the observed $R_t$, using the fixed pre-run CDF $\widehat F_t$.
They need not equal conditional moments given $R_t$. The coverage proofs
use the distribution-free buffer~\eqref{eq:probe-freedman-buffer}, and
the budget proofs use only the feasible additional-cost ledger.

\begin{algorithm}[H]
\caption{Full PROBE Algorithm}
\label{alg:probe-full}
\small
\begin{algorithmic}[1]
\Require $T$, $0<\alpha\le\tau_{\max}\le1$ with $\alpha<1$,
$\eta>0$, $\delta_{\mathrm{cov}}\in(0,1)$, $g_{\min}\in(0,1]$,
$C_{\mathrm{task}}>0$; predictable upper-quantile families
$\widehat q_t$ nonincreasing in $\tau$ and completion CDFs
$\widehat F_t$, satisfying the manuscript's quantile and allocator
regularity assumptions, with $0<U_t$ and
$0\le d_t=(\lceil U_t\rceil-R_t)_+\le d_{\max}<\infty$.
\State Initialize $\mathcal D_0\gets\varnothing$,
$a_{\mathrm{init}}\gets\log(\tau_{\max}/\alpha)$,
$a_1\gets a_{\mathrm{init}}$, and
$A_0^\rho=L_0=Q_0=\widehat V_0=S_0\gets0$.
\For{$t=1,\ldots,T$}
    \State Receive $X_t$ and fix $(\widehat q_t,\widehat F_t)$ using
    $(X_t,\mathcal D_{t-1})$; set
    $\tau_t\gets\tau_{\max}e^{-a_t}$ and issue
    $U_t\gets\widehat q_t(X_t;\tau_t)$;
    set $\rho_t\gets1-\widehat F_t(U_t)$.
    \State Observe the user's initial run of length $R_t$.
    \State Call Algorithm~\ref{alg:probe-budget} with
    $(t,X_t,R_t,U_t,\widehat q_t,\widehat F_t)$,
    $(\alpha,\tau_{\max},\eta,a_t,a_{\mathrm{init}},
    \delta_{\mathrm{cov}},g_{\min},C_{\mathrm{task}})$,
    $(A_{t-1}^\rho,L_{t-1},Q_{t-1},\widehat V_{t-1})$, and $S_{t-1}$.
    \If{the allocator returns \textsc{infeasible}}
        \State \Return the completed prefix and unchanged states;
        report incompatibility of budget and positivity at round $t$.
    \EndIf
    \State Receive $(C_t,\pi_t,W_t,d_t)$; record
    $G_t(U_t)=\pi_t$.
    Continue any unfinished trajectory until the event or timeout and observe
    $\tilde T_t=\min\{T_t,C_t\}$,
    $\Delta_t=\mathbf{1}\{T_t\le C_t\}$.
    \State Define, for any $v$,
    \[
    \begin{aligned}
        I_t^{\mathrm{err}}(v)&\gets
        \mathbf{1}\{(\tilde{T}_t>v)
        \lor (\tilde{T}_t=v\land\Delta_t=0)\},\\
        I_t^{\mathrm{reach}}(v)&\gets\mathbf 1\{C_t\ge v\}.
    \end{aligned}
    \]
    \State Compute
    \[
        \widehat e_t\gets W_t I_t^{\mathrm{err}}(U_t),
        \qquad
        b_t\gets\sup\{a\ge0:
        I_t^{\mathrm{err}}(\widehat q_t(X_t;\tau_{\max}e^{-a}))=1\},
        \quad \sup\varnothing=0.
    \]
    \State Commit the realized calibration and allocation states:
    \[
    \begin{aligned}
        \widetilde e_t&\gets\widehat e_t+
          \rho_t(1-I_t^{\mathrm{reach}}(U_t)/G_t(U_t)),
        &s_t&\gets\widetilde e_t-\alpha,\\
        A_t^\rho&\gets A_{t-1}^\rho+\rho_t(1-I_t^{\mathrm{reach}}(U_t)/G_t(U_t)),
        &L_t&\gets L_{t-1}+\widehat e_tb_t
              +a_t(\rho_t-\alpha)_+(1-I_t^{\mathrm{reach}}(U_t)),\\
        Q_t&\gets Q_{t-1}+s_t^2,
        &a_{t+1}&\gets\max\{0,a_t+\eta s_t\},\\
        \widehat m_t&\gets\rho_t,
        &S_t&\gets S_{t-1}+\pi_td_t,\\
        \widehat V_t&\gets\widehat V_{t-1}
          +\widehat m_tW_t-\widehat m_t^2.
    \end{aligned}
    \]
    \State Compute $\Gamma_t$ by~\eqref{eq:probe-gamma-candidate}
    and $\mathfrak F_t(\delta_{\mathrm{cov}})$ by~\eqref{eq:probe-freedman-buffer}.
    \State Append $(H_t(\tilde T_t),R_t,C_t,\Delta_t)$ to
    $\mathcal D_{t-1}$ to obtain $\mathcal D_t$; only then update models
    for future rounds.
\EndFor
\State \Return issued bounds $(U_t)_{t=1}^{T}$ and the final states.
\end{algorithmic}
\end{algorithm}

\section{Theory}
\label{app:theory}

\newtheorem{appendixtheorem}{Theorem}
\newtheorem{appendixcorollary}{Corollary}
\newenvironment{restatedtheorem}[2]
  {\renewcommand{\theappendixtheorem}{\ref*{#1}}\begin{appendixtheorem}[#2]}
  {\end{appendixtheorem}}
\newenvironment{restatedcorollary}[2]
  {\renewcommand{\theappendixcorollary}{\ref*{#1}}\begin{appendixcorollary}[#2]}
  {\end{appendixcorollary}}

\subsection{Proof of Theorem~\ref{thm:probe-coverage}: Coverage Validity}
\label{app:proof_probe_coverage}
\label{app:probe-ambiguity-analysis}

\begin{restatedtheorem}{thm:probe-coverage}{Coverage validity, informal}
\label{thm:probe-coverage-formal}
Under Assumptions~\ref{ass:probe-censoring}, \ref{ass:probe-positivity},
and~\ref{ass:probe-quantile},
\ourmethod{}
\eqref{eq:probe-parametrization}--\eqref{eq:probe-corrected-step}
satisfies, for every fixed $T$ and $\delta_{\mathrm{cov}}\in(0,1)$, with
probability at least $1-\delta_{\mathrm{cov}}$ over $\mathcal D_T$
\begin{equation*}
\label{eq:probe-residual-coverage-formal}
\tag{\ref*{eq:probe-residual-coverage}}
\begin{aligned}
    &\frac1T\sum_{t=1}^T\mathbb P\!\left(
        T_t\le U_t
        \,\middle|\,
        X_t,\mathcal D_{t-1}
    \right)\\
    &\quad\quad\quad\ge1-\alpha+
    \frac{1}{T}\Bigl[
    A_T^\rho
    - \Gamma_T
    - \mathfrak F_T(\delta_{\mathrm{cov}})
    \Bigr].
\end{aligned}
\end{equation*}
\end{restatedtheorem}

\subsubsection{Proof-Specific Assumptions and Information Timing}
\label{app:proof_information_timing}
\label{app:probe-assumptions}

\begin{assumption}[Quantile and threshold regularity]
\label{ass:probe-quantile}
For every $t$ and $x$, the map
$\tau\mapsto\widehat q_t(x;\tau)$ is nonincreasing on
$(0,\tau_{\max}]$. The threshold $b_t$ in
\eqref{eq:probe-threshold}, with $\sup\varnothing=0$, is finite on
every played round.
\end{assumption}

\paragraph{Observable history and information timing.}
The observed history is
\[
    \mathcal D_{t-1}
    =\bigl\{H_s(\tilde T_s),R_s,C_s,\Delta_s\bigr\}_{s<t}.
\]
We take initialization, fitted prediction rules, predictable completion
models, and tie-breaking rules as fixed. Every completed calibration and
allocation state is therefore a measurable function of this history.
Let $\mathbb P_{\mathrm{seq}}$ and $\mathbb E_{\mathrm{seq}}$ denote
probability and expectation under the full sequential law, including task,
environment, and algorithmic randomness. Define the completed-round and
prediction-time information by
\[
    \mathcal F_{t-1}=\sigma(\mathcal D_{t-1}),
    \qquad
    \mathcal P_t=\sigma(\mathcal F_{t-1},X_t),
\]
with $\mathcal F_0$ trivial. By the main-text predictability convention,
$U_t$, $\widehat F_t$, and $\rho_t$ are $\mathcal P_t$-measurable and fixed
before the user run. The committed censoring law may also depend on the
subsequently observed runtime $R_t$.
Let $\mathcal G_t$ denote the information after committing this law but
before revealing its fresh randomization or continuing the trajectory. Thus,
up to completion under $\mathbb P_{\mathrm{seq}}$,
\[
    \mathcal G_t=\sigma(\mathcal P_t,R_t)
    =\sigma(X_t,R_t,\mathcal D_{t-1}).
\]
For the two-point allocator, $\pi_t$, $u_t$, and $d_t$ are already
measurable at this time. After observing the censored trajectory, set
\[
    \mathcal D_t
    =\bigl\{H_s(\tilde T_s),R_s,C_s,\Delta_s\bigr\}_{s\le t},
    \qquad
    \mathcal F_t
    =\sigma\!\left(\mathcal G_t,H_t(\tilde T_t),R_t,C_t,\Delta_t\right)
    =\sigma(\mathcal D_t).
\]
The information flow is
\begin{equation}
\label{eq:information-flow}
    \mathcal F_{t-1}
    \subseteq\mathcal P_t
    \subseteq\mathcal G_t
    \subseteq\mathcal F_t.
\end{equation}
We use the conditional operators
\begin{equation}
\label{eq:appendix-conditional-law}
    \mathbb P_t(\,\cdot\,)
    =\mathbb P_{\mathrm{seq}}(\,\cdot\mid\mathcal G_t),
    \qquad
    \mathbb E_t[\,\cdot\,]
    =\mathbb E_{\mathrm{seq}}[\,\cdot\,\mid\mathcal G_t],
    \qquad
    G_t(u)=\mathbb P_t(C_t\ge u).
\end{equation}
In particular,
$\mathbb E_t[Z]=\mathbb E[Z\mid X_t,R_t,\mathcal D_{t-1}]$ for every integrable
$Z$. Distinguish the allocation-time miscoverage probability
$m_t^R=\mathbb P_t(T_t>U_t)$ from the prediction-time target
$m_t=\mathbb P(T_t>U_t\mid\mathcal P_t)$ in the main text. The tower property
gives $\mathbb E[m_t^R\mid\mathcal P_t]=m_t$; no independence between $R_t$
and $T_t$ is assumed.

The potential event time $T_t$ is defined under the fixed agent--environment
policy without the evaluator timeout. The timeout only truncates observation
and does not alter that policy. Conditional on $\mathcal G_t$, fresh
randomization samples $C_t$ independently of $T_t$ from the known survival
function $G_t$, as required by Assumption~\ref{ass:probe-censoring}.
Assumption~\ref{ass:probe-positivity} ensures that $G_t(U_t)\ge g_{\min}$.
The completion-based correction $\rho_t=1-\widehat F_t(U_t)\in[0,1]$ is
$\mathcal P_t$-measurable before the censoring draw. Since
$\mathbb E_t[I_t^{\mathrm{reach}}(U_t)]=G_t(U_t)$, the centered correction
$\rho_t\{1-I_t^{\mathrm{reach}}(U_t)/G_t(U_t)\}$ has conditional mean zero.
The tower property gives mean zero conditional on $\mathcal P_t$ as well.
No independence across tasks, stationarity, or exchangeability is required.

\paragraph{Variance of the correction.}
For a fixed allocation rule, write $w_t=G_t(U_t)^{-1}-1$. The part of
$\operatorname{Var}(\widetilde e_t\mid\mathcal P_t)$ depending on a
$\mathcal P_t$-measurable $\rho_t$ is
$\mathbb E[w_t(\rho_t^2-2\rho_tm_t^R)\mid\mathcal P_t]$. Thus its minimizer is
\[
    \rho_t^\star
    =\frac{\mathbb E[w_tm_t^R\mid\mathcal P_t]}
           {\mathbb E[w_t\mid\mathcal P_t]}
\]
when the denominator is positive; otherwise every $\rho_t\in[0,1]$ gives
the same variance. In particular, $\rho_t=m_t$ is optimal when $G_t(U_t)$ is
$\mathcal P_t$-measurable. If the propensity depends on informative $R_t$,
this choice need not be optimal, but unbiasedness is unchanged.

\begin{proof}[Proof of Theorem~\ref{thm:probe-coverage}]
We first establish the pathwise calibration bound and then transfer it to
conditional miscoverage.

\begin{lemma}[Threshold implication]
\label{lem:probe-threshold}
Under Assumption~\ref{ass:probe-quantile},
\[
    I_t^{\mathrm{err}}(U_t)=1\quad\Longrightarrow\quad a_t\le b_t.
\]
\end{lemma}

\par\noindent\textit{Derivation.}
When $I_t^{\mathrm{err}}(U_t)=1$, substituting
$U_t=\widehat q_t(X_t;\tau_{\max}e^{-a_t})$ shows that $a_t$ belongs to the
set whose supremum defines $b_t$.  Hence $a_t\le b_t$.  The empty-set
convention is relevant only when the premise is false.

\begin{lemma}[Compatibility of the corrected signal]
\label{lem:probe-compatibility}
Under Assumptions~\ref{ass:probe-quantile} and
\ref{ass:probe-positivity}, for every predictable $\rho_t\in[0,1]$,
\begin{equation}
\label{eq:app-probe-compatibility}
    a_t s_t
    \le
    \widehat e_tb_t+a_t(\rho_t-\alpha)_+(1-I_t^{\mathrm{reach}}(U_t)) .
\end{equation}
\end{lemma}

\par\noindent\textit{Derivation.}
On $I_t^{\mathrm{reach}}(U_t)=0$, the observed error indicator vanishes and
$s_t=\rho_t-\alpha$, hence
$a_ts_t\le a_t(\rho_t-\alpha)_+$.
On $I_t^{\mathrm{reach}}(U_t)=1$ with $I_t^{\mathrm{err}}(U_t)=0$, we have
$s_t=-\alpha-\rho_t(1-G_t(U_t))/G_t(U_t)\le0$.
On $I_t^{\mathrm{reach}}(U_t)=1$ with $I_t^{\mathrm{err}}(U_t)=1$,
Lemma~\ref{lem:probe-threshold} gives $a_t\le b_t$; because
$\rho_t(1-I_t^{\mathrm{reach}}(U_t)/G_t(U_t))\le0$, we obtain
\[
    a_ts_t
    \le a_t\widehat e_t
    \le \widehat e_tb_t.
\]
These cases exhaust all outcomes, and the right-hand side in
\eqref{eq:app-probe-compatibility} is nonnegative.

\begin{lemma}[Projected-OGD regret]
\label{lem:probe-ogd-regret}
For every comparator $a\ge0$, the update in
\eqref{eq:probe-corrected-step} satisfies
\begin{equation}
\label{eq:app-probe-ogd-regret}
    \sum_{t=1}^T
    s_t(a-a_t)
    \le
    \frac{(a-a_{\mathrm{init}})^2}{2\eta}
    +\frac{\eta}{2}Q_T,
    \qquad
    Q_T
    =\sum_{t=1}^T(s_t)^2,
    \quad Q_0=0.
\end{equation}
\end{lemma}

\par\noindent\textit{Derivation.}
Projection onto $[0,\infty)$ is nonexpansive.  Consequently,
\[
    (a_{t+1}-a)^2
    \le
    (a_t+\eta s_t-a)^2.
\]
Expanding and rearranging gives
\[
    s_t(a-a_t)
    \le
    \frac{(a_t-a)^2-(a_{t+1}-a)^2}{2\eta}
    +\frac{\eta}{2}(s_t)^2.
\]
Summation telescopes the distance terms; dropping the nonpositive terminal
distance and using $a_1=a_{\mathrm{init}}$ proves
\eqref{eq:app-probe-ogd-regret}.

\begin{proposition}[Observable surrogate envelope]
\label{prop:probe-surrogate-envelope}
Let
\[
\begin{aligned}
    L_T
    &=\sum_{t=1}^T
    \{\widehat e_tb_t+a_t(\rho_t-\alpha)_+(1-I_t^{\mathrm{reach}}(U_t))\},
    \quad L_0=0,\\
    A_T^\rho&=\sum_{t=1}^T\rho_t(1-I_t^{\mathrm{reach}}(U_t)/G_t(U_t)),
    \quad A_0^\rho=0.
\end{aligned}
\]
Then, pathwise,
\begin{equation}
\label{eq:app-probe-surrogate-envelope}
    \sum_{t=1}^T(\widehat e_t-\alpha)
    \le
    \Gamma_T-A_T^\rho,
\end{equation}
where $\Gamma_T$ is defined in
\eqref{eq:probe-realized-gamma}.
\end{proposition}

\par\noindent\textit{Derivation.}
Write
$S_T^{\mathrm{cal}}=\sum_{t=1}^Ts_t$ and define the regret
envelope
\[
    \mathcal R_T(a)
    =
    \frac{(a-a_{\mathrm{init}})^2}{2\eta}
    +\frac{\eta}{2}Q_T,
    \qquad a\ge0.
\]
Expanding~\eqref{eq:app-probe-ogd-regret}, taking the supremum over $a\ge0$,
and applying Lemma~\ref{lem:probe-compatibility} yields the Fenchel reduction
\[
    \mathcal R_T^\star(S_T^{\mathrm{cal}})
    \le
    \sum_{t=1}^T a_ts_t
    \le L_T.
\]
The conjugate over the half-line is
\[
    \mathcal R_T^\star(z)
    =
    \begin{cases}
    a_{\mathrm{init}}z+\dfrac{\eta z^2}{2}
      -\dfrac{\eta}{2}Q_T,
      & z\ge-a_{\mathrm{init}}/\eta,\\[2ex]
    -\dfrac{a_{\mathrm{init}}^2}{2\eta}
      -\dfrac{\eta}{2}Q_T,
      & z<-a_{\mathrm{init}}/\eta.
    \end{cases}
\]
If $S_T^{\mathrm{cal}}\le0$, its asserted upper bound is immediate.  Otherwise
the first branch applies, so
\[
    \eta(S_T^{\mathrm{cal}})^2
    +2a_{\mathrm{init}}S_T^{\mathrm{cal}}
    -(2L_T+\eta Q_T)
    \le0.
\]
Solving this quadratic inequality gives
$S_T^{\mathrm{cal}}\le\Gamma_T$.  Finally,
\[
    S_T^{\mathrm{cal}}
    =
    \sum_{t=1}^T(\widehat e_t-\alpha)+A_T^\rho,
\]
which proves~\eqref{eq:app-probe-surrogate-envelope}.

\paragraph{IPCW moments and concentration.}
The boundary convention $\tilde T_t=\min\{T_t,C_t\}$ and
$\Delta_t=\mathbf 1\{T_t\le C_t\}$ gives
\[
    I_t^{\mathrm{err}}(U_t)
    =\mathbf 1\{T_t>U_t,\ C_t\ge U_t\}.
\]
Indeed, $\tilde T_t>U_t$ certifies both inequalities, while
$\tilde T_t=U_t$ and $\Delta_t=0$ certifies $C_t=U_t<T_t$.
Conversely, $T_t>U_t$ and $C_t\ge U_t$ give one of these two observed cases.
Conditional independence therefore yields
$\mathbb E_t[I_t^{\mathrm{err}}(U_t)]=m_t^RG_t(U_t)$.
Since the indicator is binary and
$\widehat e_t=I_t^{\mathrm{err}}(U_t)/G_t(U_t)$, positivity gives
\begin{equation}
\label{eq:proof-ipcw-moments}
    \mathbb E_t[\widehat e_t]=m_t^R,
    \qquad
    \mathbb E_t[\widehat e_t^2]=\frac{m_t^R}{G_t(U_t)}
    \le\frac1{g_{\min}},
    \qquad
    0\le\widehat e_t\le\frac1{g_{\min}}.
\end{equation}
Conditioning these identities back on prediction-time information gives
\[
    \mathbb E[\widehat e_t\mid\mathcal P_t]=m_t,
    \qquad
    \mathbb E[\widehat e_t^2\mid\mathcal P_t]
    =\mathbb E\!\left[\frac{m_t^R}{G_t(U_t)}
        \,\middle|\,\mathcal P_t\right]
    \le\frac1{g_{\min}}.
\]
Together with the centered-correction identity, this also proves
$\mathbb E[\widetilde e_t\mid\mathcal P_t]=m_t$.
Let $\xi_t=\widehat e_t-m_t$. Then
$\mathbb E[\xi_t\mid\mathcal P_t]=0$ and
\begin{equation}
\label{eq:app-probe-variance}
    \mathbb E[\xi_t^2\mid\mathcal P_t]
    =\mathbb E\!\left[\frac{m_t^R}{G_t(U_t)}
        \,\middle|\,\mathcal P_t\right]-m_t^2
    \le\frac1{g_{\min}}.
\end{equation}
These variances sum to $V_T$ as defined in the main text.
For the filtration $\mathcal H_{2t-1}=\mathcal P_t$,
$\mathcal H_{2t}=\mathcal F_t$, with $\mathcal H_0=\mathcal F_0$, inserting
zero increments at odd indices makes $\xi_t$ the even martingale difference.

Set $\varepsilon_t=-\xi_t=m_t-\widehat e_t$. For every $\lambda\ge0$, the inequalities
$e^{-x}\le1-x+x^2/2$ for $x\ge0$ and $1+y\le e^y$ imply
\begin{align*}
    \mathbb E[e^{\lambda \varepsilon_t}\mid\mathcal P_t]
    &\le e^{\lambda m_t}
    \left(1-\lambda m_t
       +\frac{\lambda^2}{2}
           \mathbb E[\widehat e_t^2\mid\mathcal P_t]\right)\\
    &\le\exp\!\left(\frac{\lambda^2}{2g_{\min}}\right).
\end{align*}
Iterated conditioning along~\eqref{eq:information-flow} gives
\[
    \mathbb E_{\mathrm{seq}}\!\left[
        \exp\!\left(\lambda\sum_{t=1}^T \varepsilon_t\right)\right]
    \le\exp\!\left(\frac{T\lambda^2}{2g_{\min}}\right).
\]
Exponential Markov inequality, optimized over $\lambda\ge0$, therefore gives
\begin{equation}
\label{eq:proof-ipcw-concentration}
    \mathbb P_{\mathrm{seq}}\!\left(
        \sum_{t=1}^T \varepsilon_t>
        \sqrt{\frac{2T}{g_{\min}}\log\frac1\delta}
    \right)\le\delta.
\end{equation}
In particular, with probability at least $1-\delta$,
\begin{equation}
\label{eq:app-probe-freedman}
    -\sum_{t=1}^T\xi_t
    \le\sqrt{\frac{2T}{g_{\min}}\log\frac1\delta}
    +\frac13\log\frac1\delta
    =\mathfrak F_T(\delta).
\end{equation}
The elementary exponential-moment calculation thus implies the confidence
buffer used in the main statement.

\paragraph{Coverage conclusion.}
The basic decomposition is
\[
    \sum_{t=1}^T(m_t-\alpha)
    =
    \sum_{t=1}^T(\widehat e_t-\alpha)
    -\sum_{t=1}^T\xi_t.
\]
Combining Proposition~\ref{prop:probe-surrogate-envelope} with
\eqref{eq:app-probe-freedman} gives, with probability at least
$1-\delta_{\mathrm{cov}}$,
\[
    \sum_{t=1}^T(m_t-\alpha)
    \le
    \Gamma_T-A_T^\rho
    +\mathfrak F_T(\delta_{\mathrm{cov}}).
\]
Because $\mathbb P(T_t\le U_t\mid\mathcal P_t)=1-m_t$, rearranging proves
\eqref{eq:probe-residual-coverage}.
\end{proof}

The certificate bounds average conditional coverage using the observed
calibration envelope and signed centered correction, with a distribution-free
confidence buffer. Terminal cancellation is not required for this residual
bound.

\subsection{Proof of Theorem~\ref{thm:probe-budget-validity}: Expected-Budget Validity}
\label{app:probe-budget-proof}

\begin{restatedtheorem}{thm:probe-budget-validity}{Budget validity, informal}
\label{thm:probe-budget-validity-formal}
Under Assumptions~\ref{ass:probe-censoring}, \ref{ass:probe-positivity},
and~\ref{ass:probe-budget},
\ourmethod{} (Algorithm~\ref{alg:probe-full}) satisfies, for every
fixed horizon $T$,
\begin{equation*}
\label{eq:probe-expected-budget-formal}
\tag{\ref*{eq:probe-expected-budget-main}}
    \frac1T\sum_{t=1}^T
    \mathbb E[(\tilde{T}_t-R_t)_+]
    \le C_{\mathrm{task}}.
\end{equation*}
\end{restatedtheorem}

\begin{assumption}[Budget and allocator regularity]
\label{ass:probe-budget}
The requested additional runtimes satisfy
$0\le d_t=(u_t-R_t)_+\le d_{\max}<\infty$, where $u_t=\lceil U_t\rceil$.
For each fixed horizon $T$ under consideration, the feasible interval
$\Pi_t^{\mathrm{add}}$ in
\eqref{eq:appendix-budget-feasible-set} is nonempty almost surely
at every round $t=1,\ldots,T$.
The condition $C_{\mathrm{task}}\ge g_{\min}d_{\max}$
suffices inductively.
The completion model $\widehat F_t$ is a right-continuous distribution
function measurable with respect to $(X_t,\mathcal D_{t-1})$, before the user run.
The objective in~\eqref{eq:probe-anticipated-deficit} is jointly measurable,
lower semicontinuous in $\pi$ on the compact feasible interval, and finite
at a feasible point, with a fixed measurable tie-breaking rule.
\end{assumption}

\begin{proof}[Proof of Theorem~\ref{thm:probe-budget-validity}]
\begin{proposition}[Predictable validity of the selected law]
\label{prop:probe-allocation-validity}
Under Assumption~\ref{ass:probe-budget}, minimizing
\eqref{eq:probe-anticipated-deficit} over $\Pi_t^{\mathrm{add}}$, with
the zero-demand convention in Algorithm~\ref{alg:probe-budget}, yields a
$\mathcal G_t$-measurable allocation probability.  If its Bernoulli draw is fresh and
conditionally independent of $T_t$, then
\[
    \mathbb E_t[\tilde T_t-R_t]\le\mathbb E_t[C_t-R_t]=\pi_td_t,
    \qquad
    \sum_{s=1}^t\pi_sd_s\le tC_{\mathrm{task}},
    \qquad
    G_t(U_t)=\pi_t\ge g_{\min},
\]
and $C_t\perp T_t\mid\mathcal G_t$.
\end{proposition}

\par\noindent\textit{Derivation.}
The past sum $\sum_{s<t}\pi_sd_s$, $R_t$, and $d_t$ are
$\mathcal G_t$-measurable. On $\{d_t>0\}$,
\[
    \overline\pi_t
    =
    \min\!\left\{1,
    \frac{tC_{\mathrm{task}}-\sum_{s<t}\pi_sd_s}{d_t}\right\}
\]
is measurable and
$\Pi_t^{\mathrm{add}}=[g_{\min},\overline\pi_t]$ is a nonempty compact
measurable correspondence. On $\{d_t=0\}$, the algorithm sets
$\pi_t=1$ by convention, without division by $d_t$.
Lower semicontinuity and the deterministic tie-breaking rule give a
measurable minimizer when $d_t>0$, and the zero-demand branch is measurable. Writing the two-point law as
$C_t=R_t+d_tZ_t$, with $Z_t\sim\operatorname{Bernoulli}(\pi_t)$
conditionally on $\mathcal G_t$, gives
\[
    0\le\tilde T_t-R_t\le C_t-R_t=d_tZ_t,
    \qquad
    \mathbb E_t[C_t-R_t]=\pi_td_t.
\]
Here the first inequality uses $R_t\le T_t$ and $R_t\le C_t$.
Conditional monotonicity and membership in the feasible set give the
claimed cost and prefix inequalities. If $d_t>0$, integer-valued $R_t$
implies $R_t<U_t\le R_t+d_t$, so
$\{C_t\ge U_t\}=\{Z_t=1\}$ and $G_t(U_t)=\pi_t$.
If $d_t=0$, then $R_t\ge U_t$ and $G_t(U_t)=1=\pi_t$.
Thus $G_t(U_t)\ge g_{\min}$ in both cases. Finally, $R_t$, $d_t$,
and $\pi_t$ are fixed under $\mathcal G_t$, and the fresh Bernoulli
draw is conditionally independent of $T_t$, establishing the last claim.

By Assumption~\ref{ass:probe-budget}, the feasible-set constraint
holds almost surely at every round through $T$. Hence, for every
prefix $t\le T$,
\begingroup
\let\budgetprefixlabel\label
\begin{equation}
\label{eq:app-probe-prefix-budget}
\budgetprefixlabel{eq:probe-prefix-expected-budget-main}
    \sum_{s=1}^t\mathbb E_s[\tilde T_s-R_s]
    \le
    \sum_{s=1}^t\mathbb E_s[C_s-R_s]
    =
    \sum_{s=1}^t\pi_sd_s
    \le tC_{\mathrm{task}}
    \qquad\text{almost surely}.
\end{equation}
\endgroup
Here $\mathbb E_s[C_s-R_s]=\pi_sd_s$ follows from the predictable two-point
law, and the first inequality is conditional monotonicity applied to
$\tilde T_s-R_s\le C_s-R_s$. Here
$\mathbb E_s[\cdot]=\mathbb E[\cdot\mid X_s,R_s,\mathcal D_{s-1}]$,
so this is the prefix bound conditional also on the user runtime.
It implies~\eqref{eq:problem-budget-objective} when each $R_s$ is
measurable with respect to $(X_s,\mathcal D_{s-1})$.
For general random user runtimes,~\eqref{eq:problem-budget-objective}
does not follow from this ledger alone; the unconditional guarantee below
still holds. Taking outer expectations
and applying the tower property at any prefix gives
\[
    \mathbb E_{\mathrm{seq}}\!\left[\sum_{s=1}^t(\tilde{T}_s-R_s)\right]
    \le
    \mathbb E_{\mathrm{seq}}\!\left[\sum_{s=1}^t(C_s-R_s)\right]
    \le tC_{\mathrm{task}},
    \qquad t=1,\ldots,T.
\]
Setting $t=T$ and dividing by $T$ proves
\eqref{eq:probe-expected-budget-main}.
\end{proof}

The accumulating constraint controls expected additional consumption at every prefix.
The conclusion follows from predictable feasibility and conditional
monotonicity, without requiring an accurate completion model.

\subsection{Joint Coverage and High-Probability Budget Validity}
\label{app:probe-joint-proof}

\begin{theorem}[Joint coverage and high-probability budget validity]
\label{thm:high-prob-budget}
Under Assumptions~\ref{ass:probe-censoring}, \ref{ass:probe-positivity},
\ref{ass:probe-quantile}, and~\ref{ass:probe-budget}, let \ourmethod{} use
Algorithm~\ref{alg:probe-full}. Fix
$\delta_{\mathrm{cov}},\delta_{\mathrm{bud}}\in(0,1)$ with
$\delta_{\mathrm{cov}}+\delta_{\mathrm{bud}}<1$. Then, for every fixed
horizon $T$, with probability at least
$1-\delta_{\mathrm{cov}}-\delta_{\mathrm{bud}}$, the residual coverage
certificate~\eqref{eq:probe-residual-coverage} and
\begin{equation}
\label{eq:probe-realized-budget-main}
\begin{aligned}
    &\sum_{t=1}^T(\tilde{T}_t-R_t)
    \le\sum_{t=1}^T(C_t-R_t)\\
    &\quad\le TC_{\mathrm{task}}
    +\frac{d_{\max}}{3}\log\frac1{\delta_{\mathrm{bud}}}\\
    &\qquad+\sqrt{
    2d_{\max}TC_{\mathrm{task}}\log\frac1{\delta_{\mathrm{bud}}}
    +\frac{d_{\max}^2}{9}
    \log^2\frac1{\delta_{\mathrm{bud}}}}
\end{aligned}
\end{equation}
hold simultaneously.

\end{theorem}

\begin{proof}[Proof of Theorem~\ref{thm:high-prob-budget}]
For the realized additional allocation, define
$\zeta_t=(C_t-R_t)-\pi_td_t=d_t(Z_t-\pi_t)$, where
$Z_t\mid\mathcal G_t\sim\operatorname{Bernoulli}(\pi_t)$.
Conditional on $\mathcal G_t$,
$\mathbb E_t[\zeta_t]=0$,
$\zeta_t\le d_t\le d_{\max}$, and
\[
    \mathbb E_t[\zeta_t^2]
    =\pi_t(1-\pi_t)d_t^2.
\]
Inserting a zero increment between $\mathcal F_{t-1}$ and
$\mathcal G_t$ produces an ordinary martingale difference sequence.  Its predictable quadratic variation obeys
\[
\begin{split}
    V_T^{\mathrm{bud}}
    &=
    \sum_{t=1}^T\pi_t(1-\pi_t)d_t^2\\
    &\le
    d_{\max}\sum_{t=1}^T\pi_td_t
    \le d_{\max}TC_{\mathrm{task}}.
\end{split}
\]
One-sided Freedman concentration therefore implies that, with probability at
least $1-\delta_{\mathrm{bud}}$,
\[
    \sum_{t=1}^T\zeta_t
    \le
    \frac{d_{\max}}{3}\log\frac1{\delta_{\mathrm{bud}}}
    +\sqrt{
    2d_{\max}TC_{\mathrm{task}}\log\frac1{\delta_{\mathrm{bud}}}
    +\frac{d_{\max}^2}{9}
    \log^2\frac1{\delta_{\mathrm{bud}}}}.
\]
Combining this inequality with
$\sum_t\pi_td_t\le TC_{\mathrm{task}}$ and
$\tilde{T}_t-R_t\le C_t-R_t$ proves
\eqref{eq:probe-realized-budget-main} on an event of probability at least
$1-\delta_{\mathrm{bud}}$.

Assumptions~\ref{ass:probe-censoring}, \ref{ass:probe-positivity},
and~\ref{ass:probe-quantile}
allow us to apply Theorem~\ref{thm:probe-coverage}. Hence the residual
coverage certificate
\eqref{eq:probe-residual-coverage} holds on an event of probability at least
$1-\delta_{\mathrm{cov}}$. A union bound shows that this event and the budget
event above hold simultaneously with probability at least
$1-\delta_{\mathrm{cov}}-\delta_{\mathrm{bud}}$; no independence between
the events is required.

\end{proof}

The realized-budget guarantee is a fixed-horizon high-probability statement;
it is not a strict pathwise hard-budget cap.

\subsection{Coverage of ACI+IPCW}
\label{app:proof_aci_ipcw}
\label{app:background-ipcw}

\begin{theorem}[Coverage of ACI$+$IPCW]
\label{thm:aci-ipcw-coverage}
Consider Algorithm~\ref{alg:aci_ipcw} with fixed step size
$\eta>0$ and deterministic initialization $\tau_1$.
Assume Assumptions~\ref{ass:probe-censoring}
and~\ref{ass:probe-positivity}, predictable bounds and
censoring laws, and known censoring probabilities.
The prediction rule must be defined at every reachable
calibration state, and the update is unprojected:
\[
    \tau_{t+1}
    =\tau_t+\eta(\alpha-\widehat e_t).
\]
Define the average conditional coverage
\[
    \overline c_T
    :=\frac1T\sum_{t=1}^T
    \mathbb P\!\left(
        T_t\le U_t
        \,\middle|\,
        X_t,\mathcal D_{t-1}
    \right).
\]
For every fixed horizon $T\ge1$ and $\delta\in(0,1)$,
with probability at least $1-\delta$,
\begin{equation}
\label{eq:aci-ipcw-coverage-bound}
    \overline c_T
    \ge
    1-\alpha
    -\frac{\tau_1-\tau_{T+1}}{\eta T}
    -\sqrt{\frac{2\log(1/\delta)}{Tg_{\min}}}.
\end{equation}
If, additionally,
\begin{equation}
\label{eq:aci-ipcw-state-condition}
    \frac{(\tau_1-\tau_{T+1})_+}{T}
    \longrightarrow0
    \qquad\text{almost surely},
\end{equation}
where $(x)_+=\max\{x,0\}$, then
\[
    \liminf_{T\to\infty}\overline c_T
    \ge1-\alpha
    \qquad\text{almost surely}.
\]
Condition~\eqref{eq:aci-ipcw-state-condition} holds, in
particular, whenever $\inf_{t\ge1}\tau_t>-\infty$
almost surely.
\end{theorem}

\begin{proof}
Let $\mathcal P_t=\sigma(X_t,\mathcal D_{t-1})$,
$\mathcal G_t=\sigma(\mathcal P_t,R_t)$,
and $\mathcal F_t=\sigma(\mathcal D_t)$, so that
$\mathcal F_{t-1}\subseteq\mathcal P_t\subseteq\mathcal G_t
\subseteq\mathcal F_t$. The bound $U_t$ is fixed given $\mathcal P_t$;
the censoring law is fixed given $\mathcal G_t$.
The boundary-aware indicator satisfies
\[
    I_t^{\mathrm{err}}(U_t)
    =\mathbf 1\{T_t>U_t,\ C_t\ge U_t\}.
\]
Conditional independence and positivity give
$\mathbb E[\widehat e_t\mid\mathcal G_t]=m_t^R$ and
$\mathbb E[\widehat e_t^2\mid\mathcal G_t]=m_t^R/G_t(U_t)$,
where $m_t^R=\mathbb P(T_t>U_t\mid\mathcal G_t)$. The tower property
therefore gives the prediction-time moments
\begin{equation}
\label{eq:aci-ipcw-proof-moments}
\begin{aligned}
    \mathbb E[\widehat e_t\mid\mathcal P_t]
    &=m_t,\\
    \mathbb E[\widehat e_t^2\mid\mathcal P_t]
    &=\mathbb E\!\left[\frac{m_t^R}{G_t(U_t)}
        \,\middle|\,\mathcal P_t\right]
    \le\frac1{g_{\min}}.
\end{aligned}
\end{equation}

Set $\varepsilon_t=m_t-\widehat e_t$.
Since $\widehat e_t\ge0$, the elementary inequalities
$e^{-x}\le1-x+x^2/2$ for $x\ge0$ and
$1+y\le e^y$ imply, for every $\lambda\ge0$,
\begin{align*}
    \mathbb E[e^{\lambda \varepsilon_t}\mid\mathcal P_t]
    &\le
    e^{\lambda m_t}
    \left(
        1-\lambda m_t
        +\frac{\lambda^2}{2}
        \mathbb E[\widehat e_t^2\mid\mathcal P_t]
    \right)
    \le
    \exp\!\left(\frac{\lambda^2}{2g_{\min}}\right).
\end{align*}
Iterated conditioning along the above filtration yields
\[
    \mathbb E\!\left[
        \exp\!\left(\lambda\sum_{t=1}^T \varepsilon_t\right)
    \right]
    \le
    \exp\!\left(\frac{T\lambda^2}{2g_{\min}}\right).
\]
Consequently, exponential Markov inequality, optimized
over $\lambda\ge0$, gives
\[
    \mathbb P\!\left(
        \sum_{t=1}^T \varepsilon_t
        >
        \sqrt{\frac{2T}{g_{\min}}\log\frac1\delta}
    \right)
    \le\delta.
\]

Summing the unprojected ACI update gives the exact identity
\[
    \sum_{t=1}^T(\widehat e_t-\alpha)
    =\frac{\tau_1-\tau_{T+1}}{\eta}.
\]
Hence
\[
    \overline c_T
    =
    1-\alpha
    -\frac{\tau_1-\tau_{T+1}}{\eta T}
    -\frac1T\sum_{t=1}^T \varepsilon_t.
\]
Combining the last two displays proves
\eqref{eq:aci-ipcw-coverage-bound}.

Finally, apply the finite-horizon bound with
$\delta_T=T^{-2}$ for $T\ge2$.
Since $\sum_{T=2}^{\infty}\delta_T<\infty$,
the Borel--Cantelli lemma implies that almost surely,
for all sufficiently large $T$,
\[
    \overline c_T
    \ge
    1-\alpha
    -\frac{(\tau_1-\tau_{T+1})_+}{\eta T}
    -\sqrt{\frac{4\log T}{Tg_{\min}}}.
\]
Condition~\eqref{eq:aci-ipcw-state-condition}
then proves the asymptotic claim.
\end{proof}

The calibration-state condition is an additional
requirement; conditional unbiasedness of the IPCW
feedback alone does not establish it.

\section{Selection Rule for the Positivity Floor
\texorpdfstring{$g_{\min}$}{gmin}}
\label{app:probe-gmin-selection}

For the additional-budget allocator in Algorithm~\ref{alg:probe-budget}, a
larger $g_{\min}$ limits IPCW weights but increases the minimum expected
additional allocation $g_{\min}d_t$. From a disjoint pilot sample, estimate the mean
requested additional runtime $\widehat\mu_d=n^{-1}\sum_{i=1}^n d_i^{\mathrm{ref}}$,
where $d_i^{\mathrm{ref}}=(\lceil U_i^{\mathrm{ref}}\rceil-R_i^{\mathrm{ref}})_+$.
With a planning-slack factor $\gamma\in(0,1)$, set
\[
    g_{\min}=\begin{cases}
    \min\left\{1,\gamma C_{\mathrm{task}}/\widehat\mu_d\right\},
       &\widehat\mu_d>0,\\
    1, &\widehat\mu_d=0,
    \end{cases}
\]
and freeze the floor before the evaluated online run. At the pilot mean,
the floor commits at most a $\gamma$ fraction of the per-task budget.

If a valid positive upper bound $d_t\le d_{\max}$ is available, additionally cap
$g_{\min}$ at $C_{\mathrm{task}}/d_{\max}$ to guarantee a nonempty feasible
interval on every round (Assumption~\ref{ass:probe-budget}). Without this
cap, the pilot rule is a heuristic: check
$\Pi_t^{\mathrm{add}}$ before each allocation and invoke the declared
infeasibility protocol if it is empty, as in Algorithm~\ref{alg:probe-full}.

\section{Deterministic Budget Allocation}
\label{app:deterministic-budget}

For a hard cap on additional evaluation cost, fix a finite budget $B_{\mathrm{global}}\ge0$
for an accounting window and initialize $B_1=B_{\mathrm{global}}$, with
no replenishment. After the user run and before continuing the evaluation, compute
$d_t=(\lceil U_t\rceil-R_t)_+$. If $B_t\ge d_t$, set $C_t=R_t+d_t$,
reserve the additional runtime $C_t-R_t=d_t$, continue the round, and update
$B_{t+1}=B_t-(C_t-R_t)$. Debit the full additional reservation even if
the task completes earlier. Since $0\le C_t-R_t\le B_t$, induction gives
$B_{t+1}\ge0$. Telescoping the ledger update yields, for every prefix,
\[
    0\le\sum_{t=1}^n(\tilde T_t-R_t)
    \le\sum_{t=1}^n(C_t-R_t)
    =B_{\mathrm{global}}-B_{n+1}
    \le B_{\mathrm{global}}.
\]
If $B_t<d_t$, set $C_t=R_t$ and $B_{t+1}=B_t$, report budget exhaustion, and terminate the
active sequence. Freeze the calibration state and IPCW statistics and set
all subsequent additional allocations $C_s-R_s$ in the window to zero.
Because $B_t<d_t$ implies $R_t<U_t$, this branch has $G_t(U_t)=0$.
Do not form the undefined
IPCW ratio or replace it with an artificial zero.

On funded rounds, $G_t(U_t)=1$ and the coverage error is fully observed.
The hard cap holds pathwise; coverage is not claimed for skipped rounds or
an average stopped at exhaustion. A fixed-horizon coverage guarantee still
requires the original coverage assumptions and funding every round;
$B_{\mathrm{global}}\ge T d_{\max}$ suffices when $d_t\le d_{\max}$.

\newpage
\section{Additional Experiments}
\label{app:additional_experiments}
In this section, we provide additional empirical results that further support the findings presented in the main text. Specifically, we expand upon the evaluations of our proposed \ourmethod{} framework across various target coverage levels and budget constraints.  As in Section~\ref{sec:experiments}, all experiments use $R_t=0$; the reported assigned budget $\sum_t C_t$ therefore equals the additional assigned budget $\sum_t(C_t-R_t)$. 

\subsection{SWE Settings}
Building upon the software engineering evaluation in Section~\ref{sec:experiments}, we present further experiments on the SWE dataset(~\ref{app:exp_details:swe}) under more stringent resource limitations. All experiments in this subsection are evaluated over 1,000 sequential trials across 10 independent repetitions, with an expected budget constraint of $C_{\mathrm{task}} = 15.0$. The evaluation protocol and metrics remain identical to the experiments appearing in the main paper.   

\begin{figure}[t!]
    \centering
    \begin{minipage}[t]{0.58\linewidth}
        \vspace{0pt}
        \includegraphics[width=\linewidth]{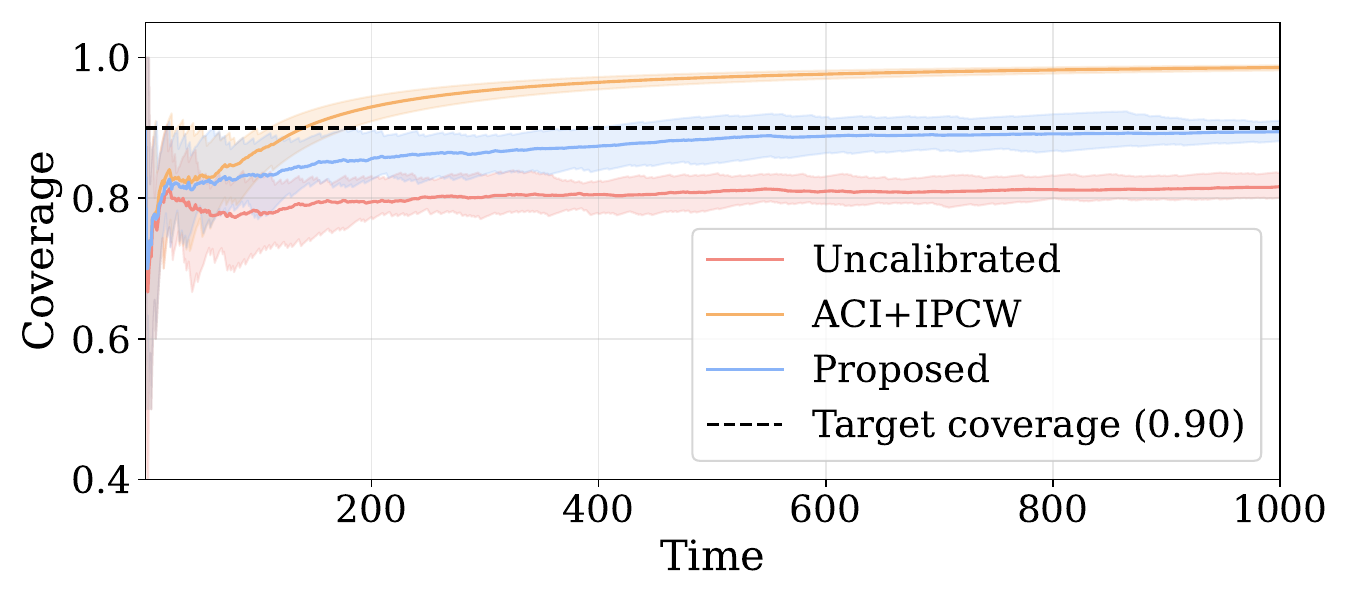}
        \caption{Cumulative coverage rate over time, $\alpha = 0.1$}
    \end{minipage}\hfill
    \begin{minipage}[t]{0.41\linewidth}
        \vspace{0pt}
        \includegraphics[width=\linewidth]{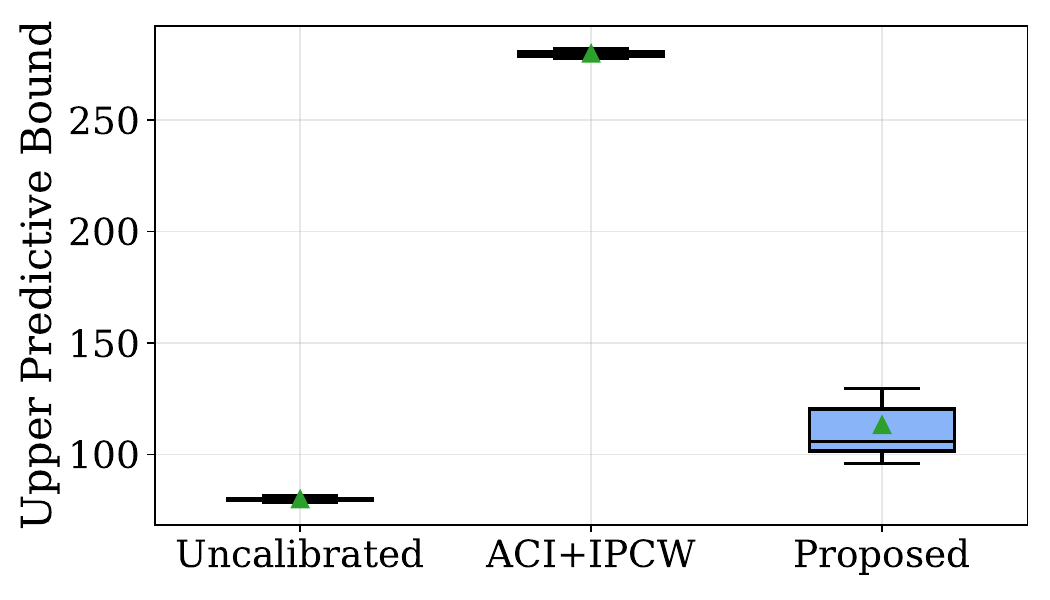}
        \vspace{0.05cm}
        \caption{Upper predictive bound, $\alpha = 0.1$}
    \end{minipage}
\end{figure}

\begin{figure}[t!]
    \centering
    \begin{minipage}[t]{0.58\linewidth}
        \vspace{0pt}
        \includegraphics[width=\linewidth]{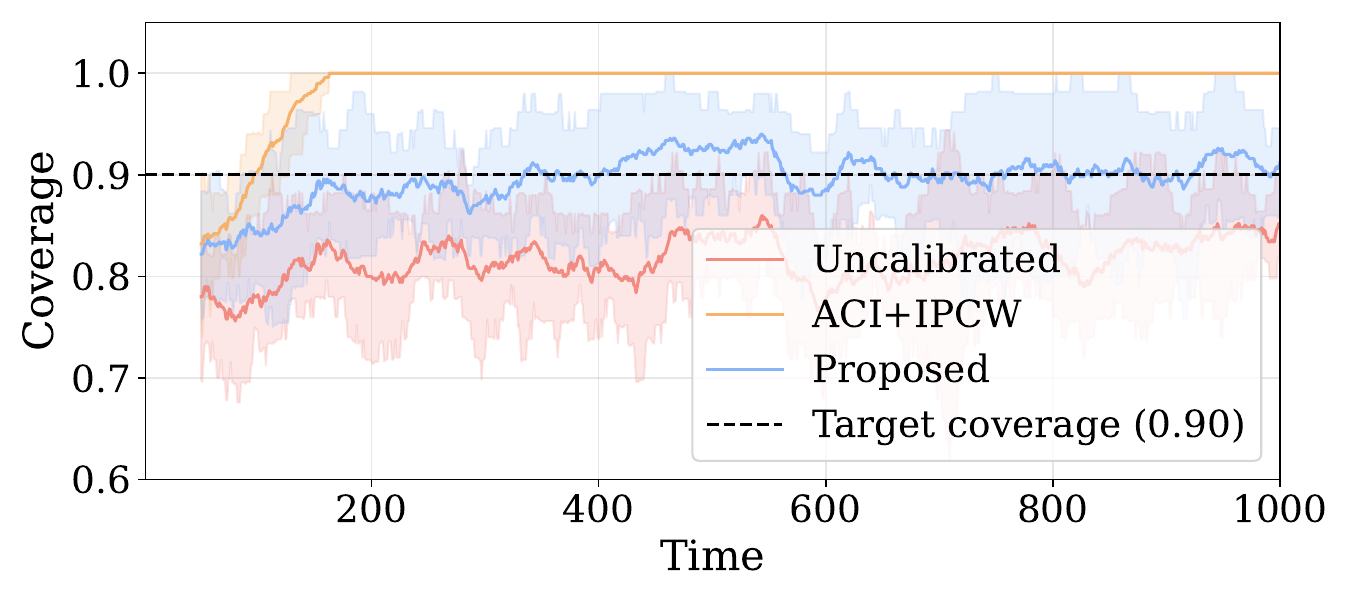}
        \caption{Local coverage over time, $\alpha = 0.1$, window$=50$}
    \end{minipage}\hfill
    \begin{minipage}[t]{0.41\linewidth}
        \vspace{0pt}
        \includegraphics[width=\linewidth]{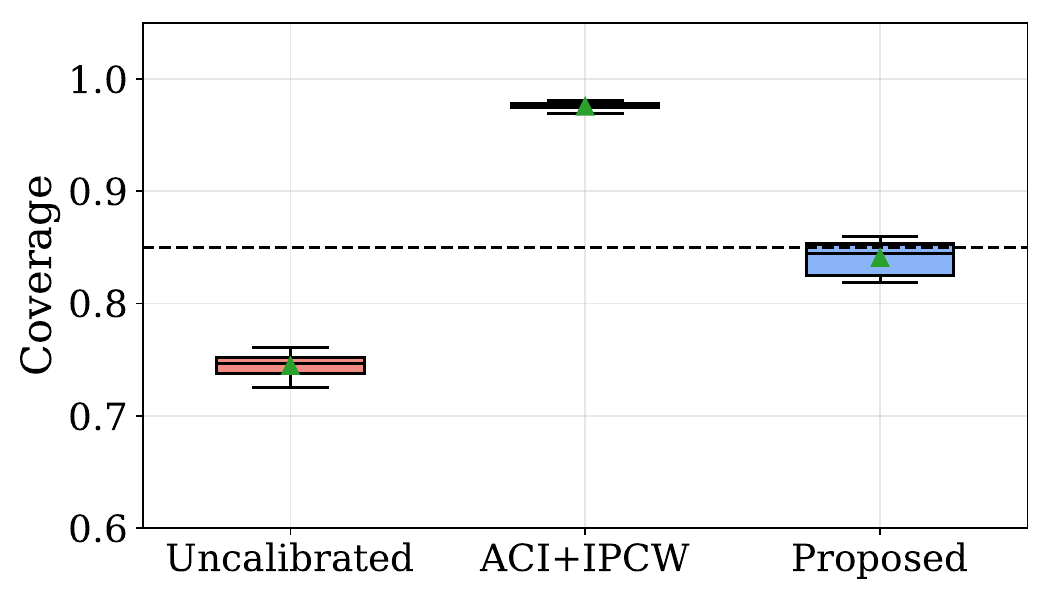}
        \vspace{0.05cm}
        \caption{Coverage boxplot, $\alpha = 0.15$}
    \end{minipage}
\end{figure}

\begin{figure}[t!]
    \centering
    \begin{minipage}[t]{0.58\linewidth}
        \vspace{0pt}
        \includegraphics[width=\linewidth]{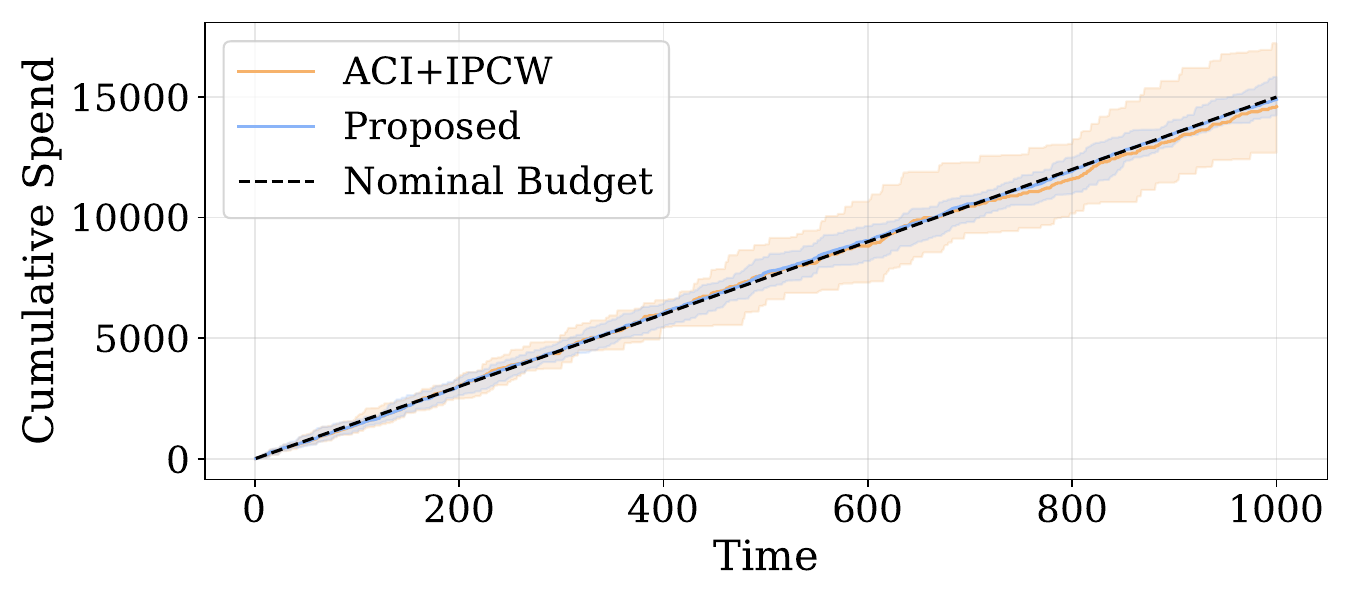}
        \caption{Cumulative budget spend over time, $\alpha = 0.1$}
    \end{minipage}\hfill
    \begin{minipage}[t]{0.41\linewidth}
        \vspace{0pt}
        \includegraphics[width=\linewidth]{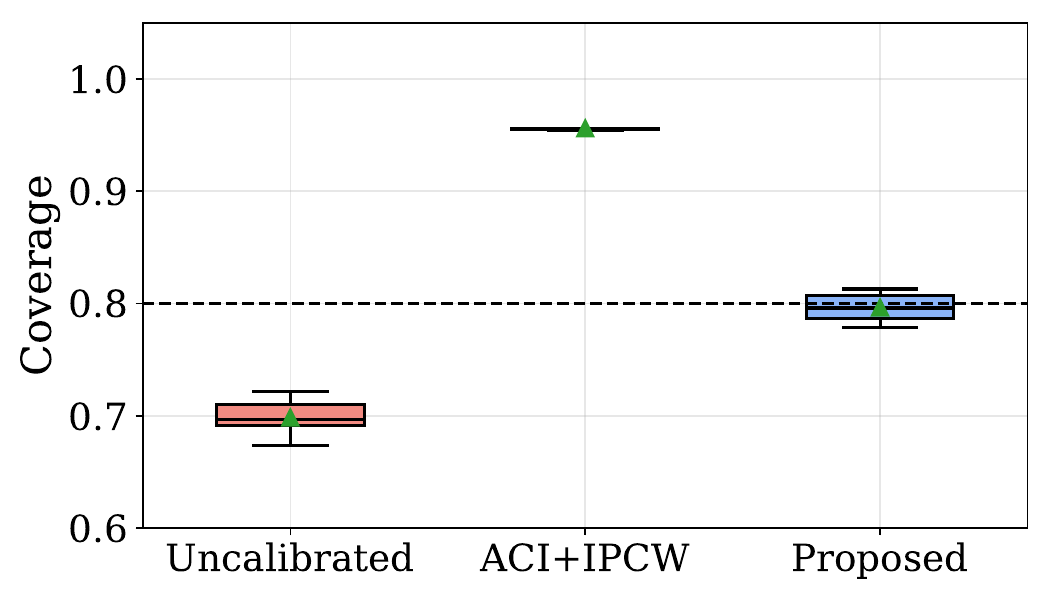}
        \vspace{0.05cm}
        \caption{Coverage boxplot, $\alpha = 0.2$}
    \end{minipage}
\end{figure}

\newpage
\subsection{RAG Settings}
Building upon the retrieval-augmented generation evaluation in Section~\ref{sec:experiments}, we present further experiments on the NFCorpus benchmark~\ref{app:exp_details:rag}. All experiments in this subsection evaluate the predicted number of relevant corpus items over 1,000 sequential trials across 10 independent repetitions. These evaluations apply various expected budget constraints $C_{\mathrm{task}}$ and target miscoverage levels $\alpha$, mirroring the experimental structure introduced in the main text.  

\begin{figure}[t!]
    \centering
    \begin{minipage}[t]{0.58\linewidth}
        \vspace{0pt}
        \includegraphics[width=\linewidth]{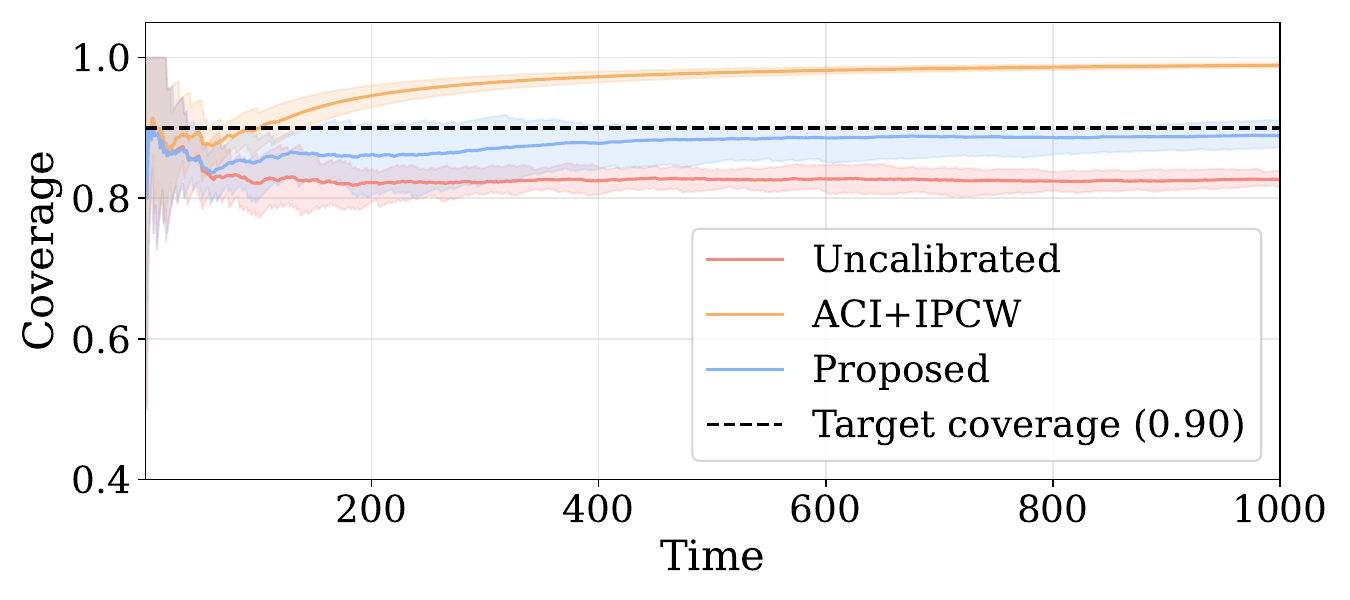}
        \caption{Cumulative coverage rate over time, $\alpha = 0.1$}
    \end{minipage}\hfill
    \begin{minipage}[t]{0.41\linewidth}
        \vspace{0pt}
        \includegraphics[width=\linewidth]{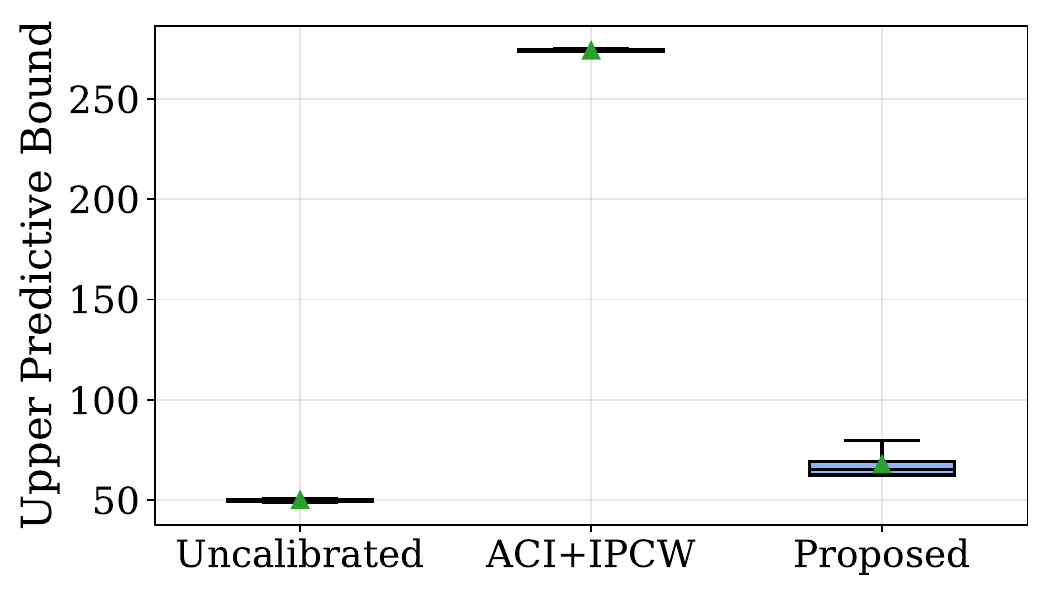}
        \vspace{0.05cm}
        \caption{Upper predictive bound, $\alpha = 0.1$}
    \end{minipage}
\end{figure}

\begin{figure}[t!]
    \centering
    \begin{minipage}[t]{0.58\linewidth}
        \vspace{0pt}
        \includegraphics[width=\linewidth]{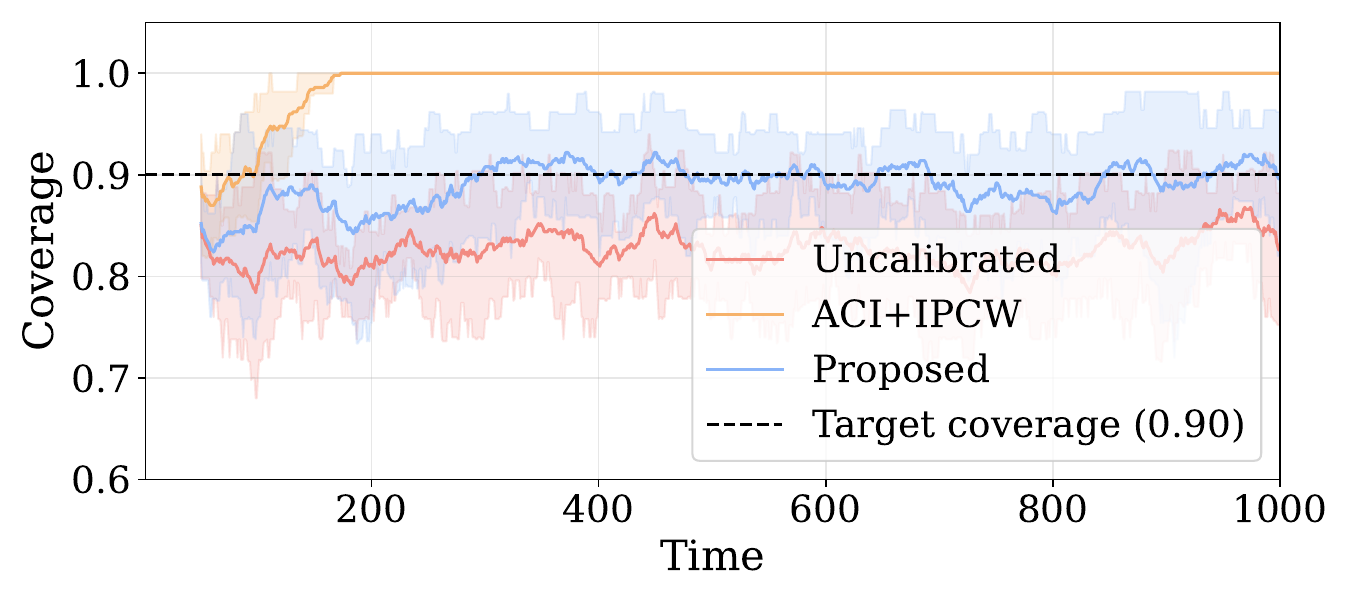}
        \caption{Local coverage over time, $\alpha = 0.1$, window$=50$}
    \end{minipage}\hfill
    \begin{minipage}[t]{0.41\linewidth}
        \vspace{0pt}
        \includegraphics[width=\linewidth]{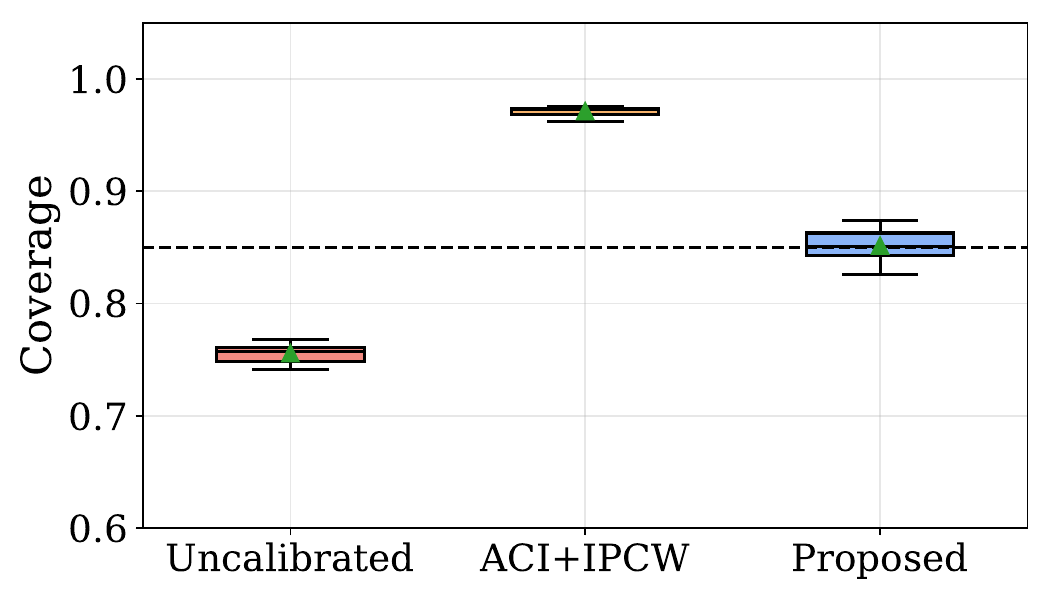}
        \vspace{0.05cm}
        \caption{Cumulative coverage, $\alpha = 0.15$}
    \end{minipage}
\end{figure}

\begin{figure}[t!]
    \centering
    \begin{minipage}[t]{0.58\linewidth}
        \vspace{0pt}
        \includegraphics[width=\linewidth]{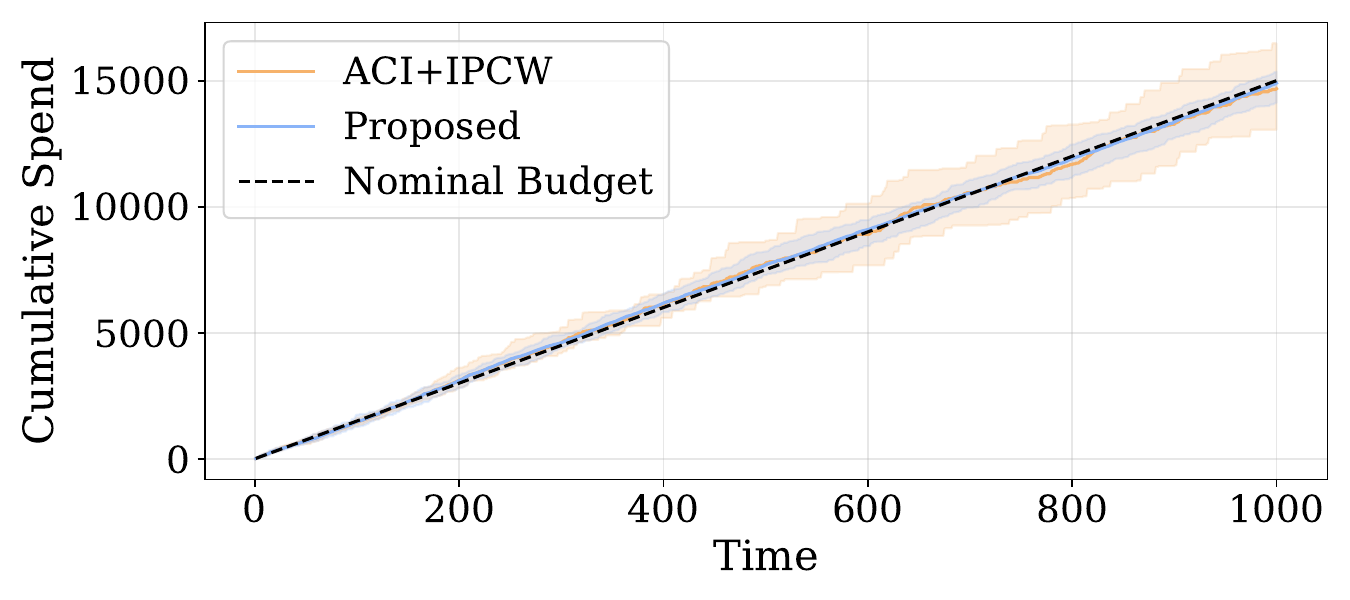}
        \caption{Cumulative budget spend over time, $\alpha = 0.1$}
    \end{minipage}\hfill
    \begin{minipage}[t]{0.41\linewidth}
        \vspace{0pt}
        \includegraphics[width=\linewidth]{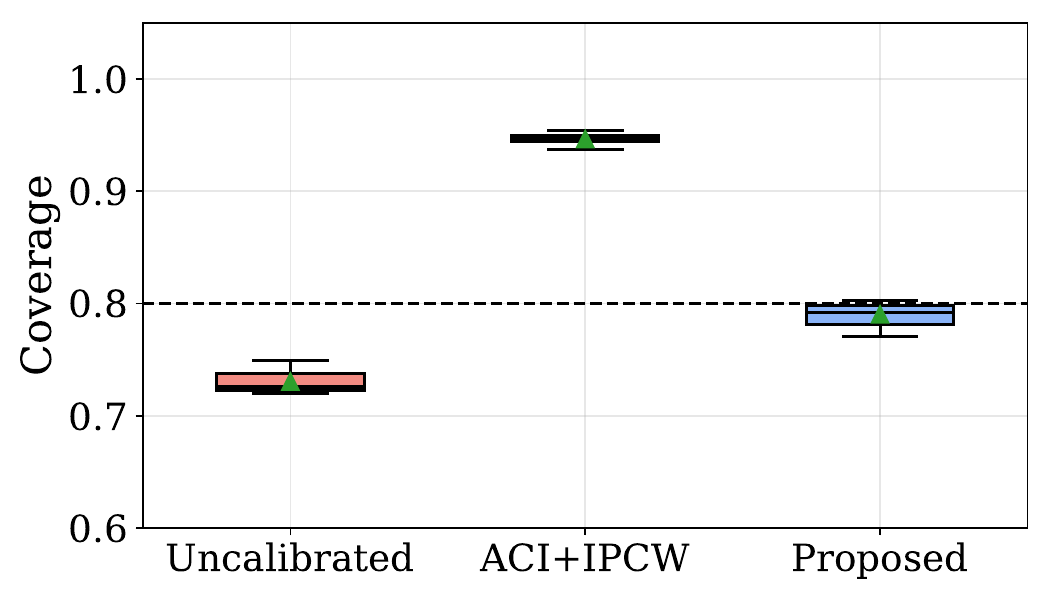}
        \vspace{0.05cm}
        \caption{Cumulative coverage, $\alpha = 0.2$}
    \end{minipage}
\end{figure}

\newpage
\subsection{AutoIF Settings}
Building upon the instruction following evaluation in Section~\ref{sec:experiments}, we present further experiments on the NFCorpus benchmark~\ref{app:exp_details:autoif}. All experiments in this subsection evaluate the predicted number of relevant corpus items over 1,000 sequential trials across 10 independent repetitions. These evaluations apply various expected budget constraints $C_{\mathrm{task}}$ and target miscoverage levels $\alpha$, mirroring the experimental structure introduced in the main text.

\begin{figure}[t!]
    \centering
    \begin{minipage}[t]{0.58\linewidth}
        \vspace{0pt}
        \includegraphics[width=\linewidth]{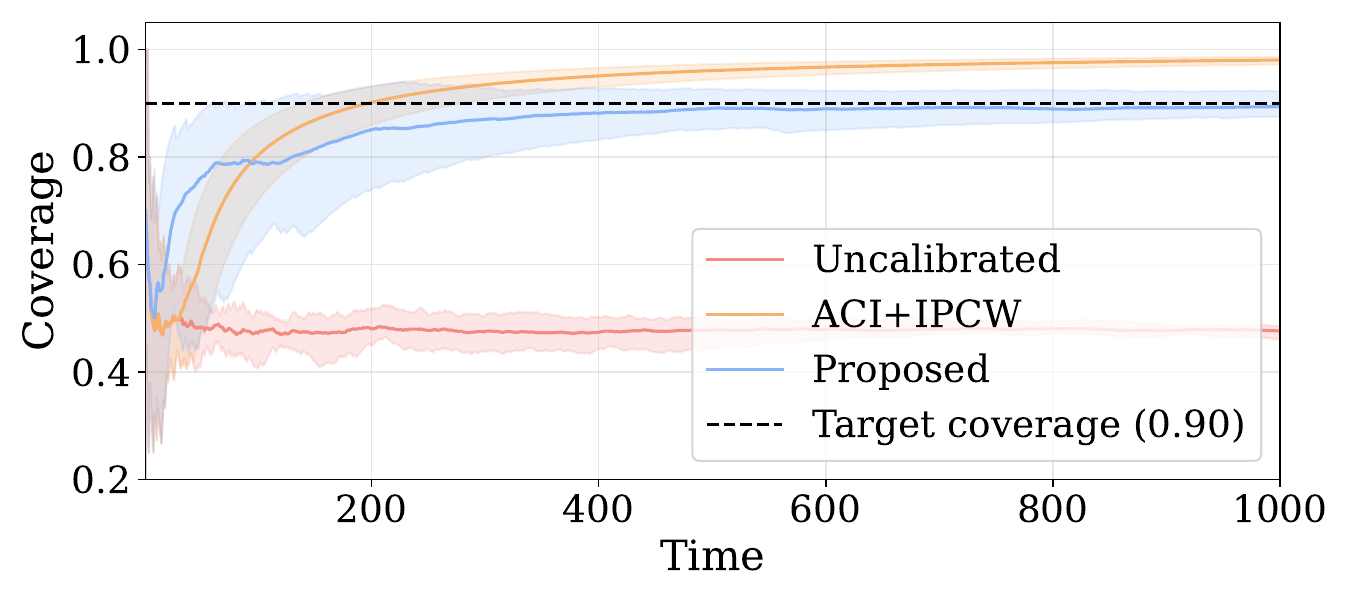}
        \caption{Cumulative coverage rate over time, $\alpha = 0.1$}
    \end{minipage}\hfill
    \begin{minipage}[t]{0.41\linewidth}
        \vspace{0pt}
        \includegraphics[width=\linewidth]{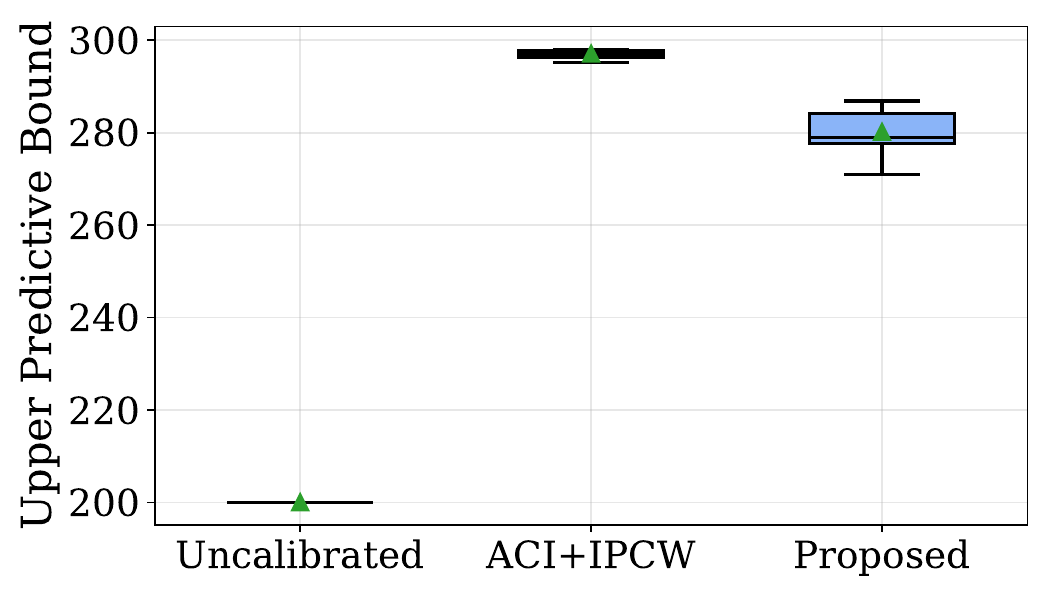}
        \vspace{0.05cm}
        \caption{Upper predictive bound, $\alpha = 0.1$}
    \end{minipage}
\end{figure}

\begin{figure}[t!]
    \centering
    \begin{minipage}[t]{0.58\linewidth}
        \vspace{0pt}
        \includegraphics[width=\linewidth]{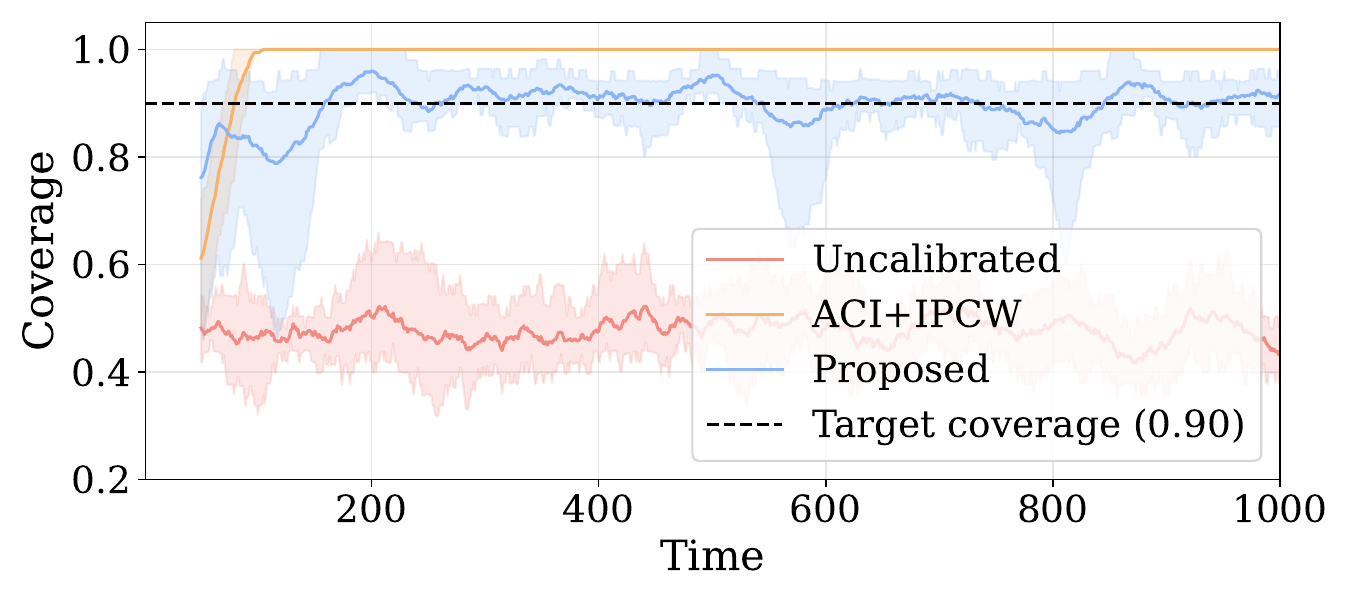}
        \caption{Local coverage over time, $\alpha = 0.1$, window$=50$}
    \end{minipage}\hfill
    \begin{minipage}[t]{0.41\linewidth}
        \vspace{0pt}
        \includegraphics[width=\linewidth]{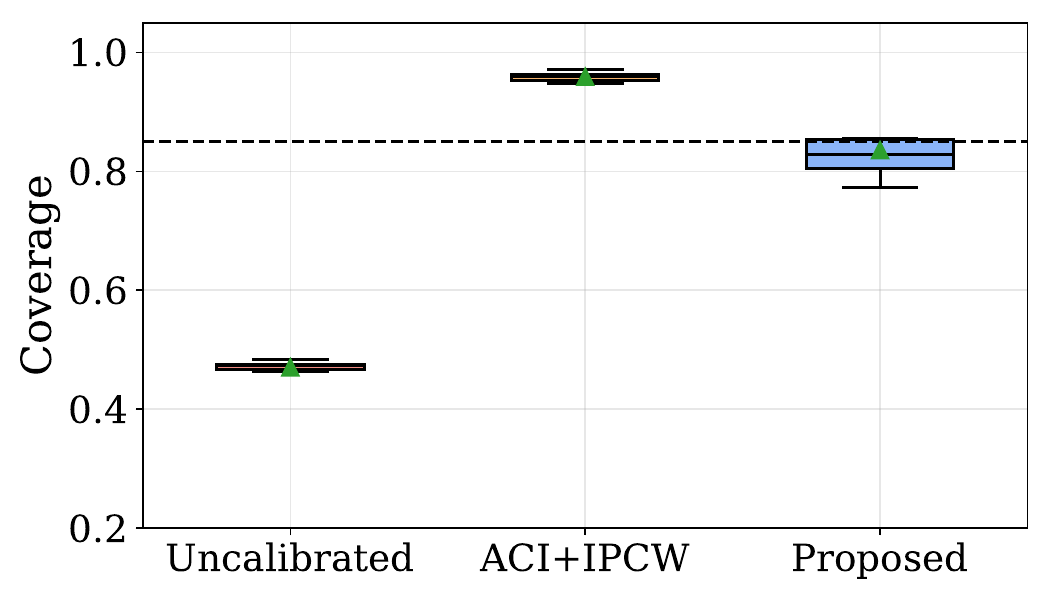}
        \vspace{0.05cm}
        \caption{Cumulative coverage, $\alpha = 0.15$}
    \end{minipage}
\end{figure}

\begin{figure}[t!]
    \centering
    \begin{minipage}[t]{0.58\linewidth}
        \vspace{0pt}
        \includegraphics[width=\linewidth]{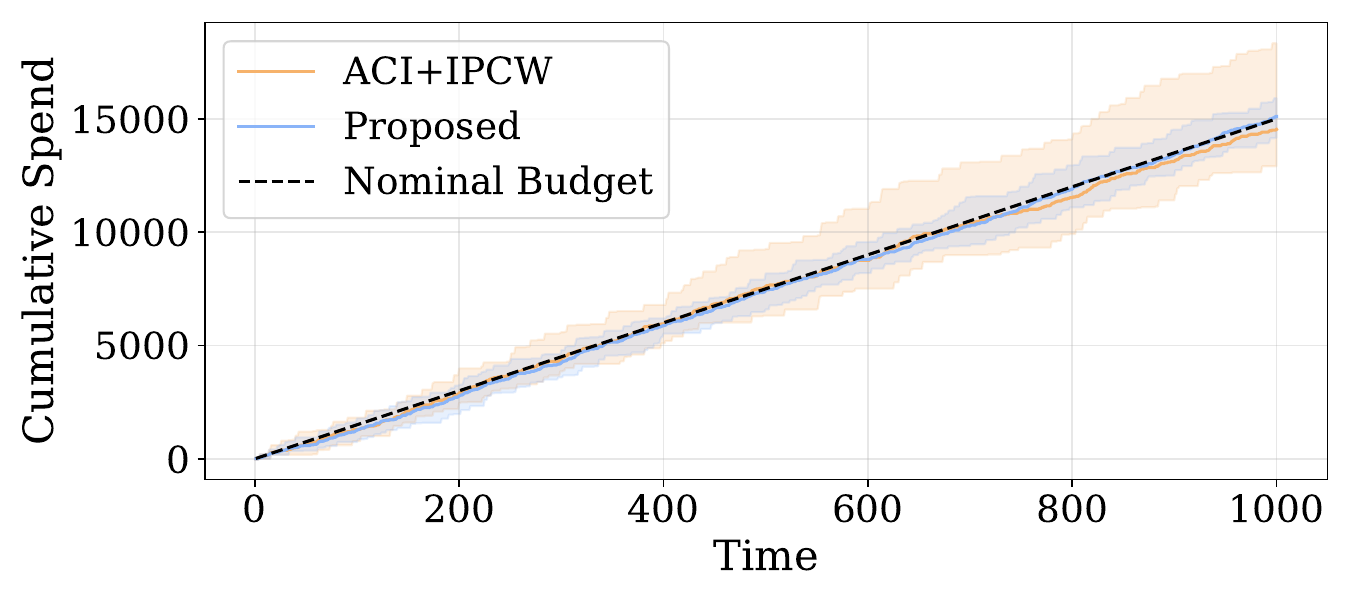}
        \caption{Cumulative budget spend over time, $\alpha = 0.1$}
    \end{minipage}\hfill
    \begin{minipage}[t]{0.41\linewidth}
        \vspace{0pt}
        \includegraphics[width=\linewidth]{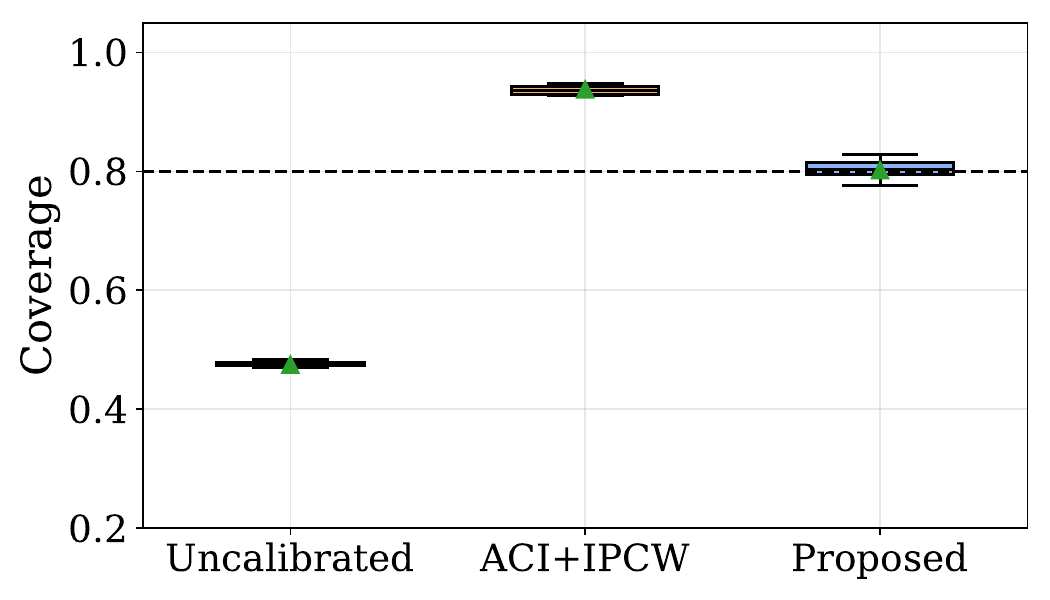}
        \vspace{0.05cm}
        \caption{Cumulative coverage, $\alpha = 0.2$}
    \end{minipage}
\end{figure}

\subsection{Ablation Study: ACI+IPCW Calibration Variants}
\label{app:additional_experiments:aci_variants}
We assess the effect of a small implementation adjustment to
\texttt{ACI+IPCW}: assigning a unit penalty on zero-budget rounds
($C_t=0$). The variant \texttt{ACI+IPCW (w/o penalty)} uses the
standard IPCW feedback, whereas \texttt{ACI+IPCW (w/ penalty)}
replaces its zero feedback on these rounds with one. Under tight
budgets, the unpenalized implementation exhibits undercoverage:
skipped rounds induce optimistic calibration updates, while clipping
limits the compensating corrections from rare, large IPCW penalties.
The unit penalty counteracts this effect and restores empirical
coverage in the reported experiments, at the cost of wider upper
predictive bounds. We therefore use the penalized variant in the
main experiments to provide a fair comparison with a baseline that
maintains empirical coverage. Appendix~\ref{app:exp_details:implementation}
details this conservative adjustment and its relation to the
theoretical coverage guarantee.

\subsubsection{Real Data (SWE settings)}
We first examine the effect of the zero-budget penalty on the real-world SWE data described in Appendix~\ref{app:exp_details:swe}, using an expected budget of $C_{\mathrm{task}}=15$. The results below compare the two \texttt{ACI+IPCW} variants with \ourmethod{} and the uncalibrated baseline in terms of cumulative and local coverage, upper predictive bounds, and budget expenditure.

\begin{figure}[t!]
    \centering
    \begin{minipage}[t]{0.58\linewidth}
        \vspace{0pt}
        \includegraphics[width=\linewidth]{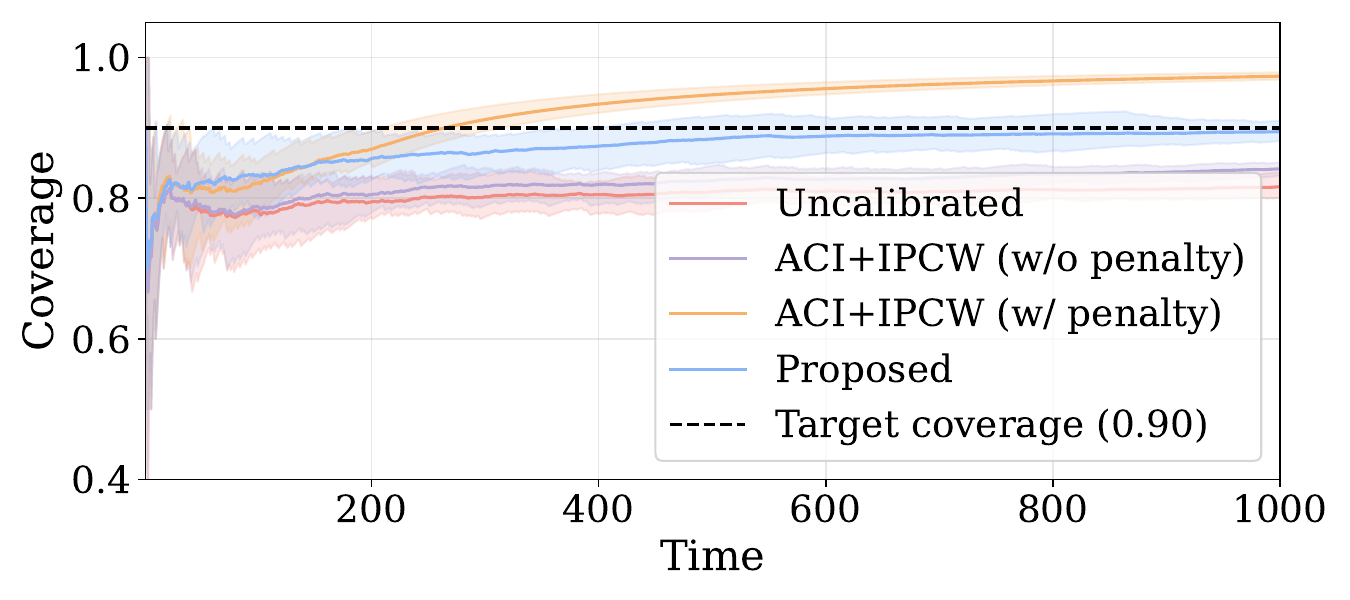}
        \caption{Cumulative coverage rate over time, $\alpha = 0.1$}
    \end{minipage}\hfill
    \begin{minipage}[t]{0.41\linewidth}
        \vspace{0pt}
        \includegraphics[trim={1cm 0cm 1cm 0cm}, clip, width=\linewidth]{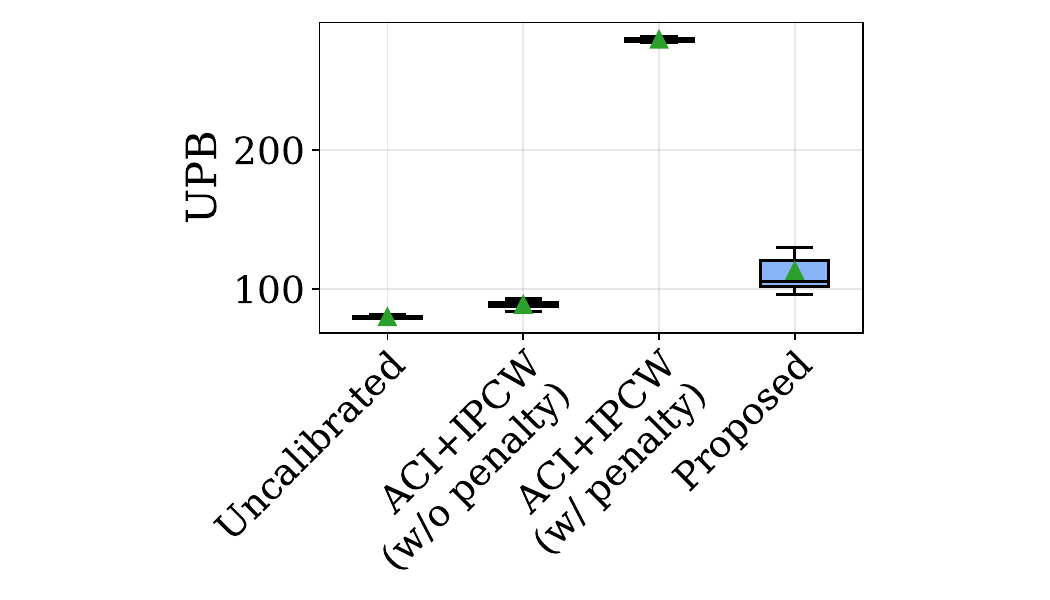}
        \caption{Upper predictive bound, $\alpha = 0.1$}
    \end{minipage}
\end{figure}

\begin{figure}[t!]
    \centering
    \begin{minipage}[t]{0.58\linewidth}
        \vspace{0pt}
        \includegraphics[width=\linewidth]{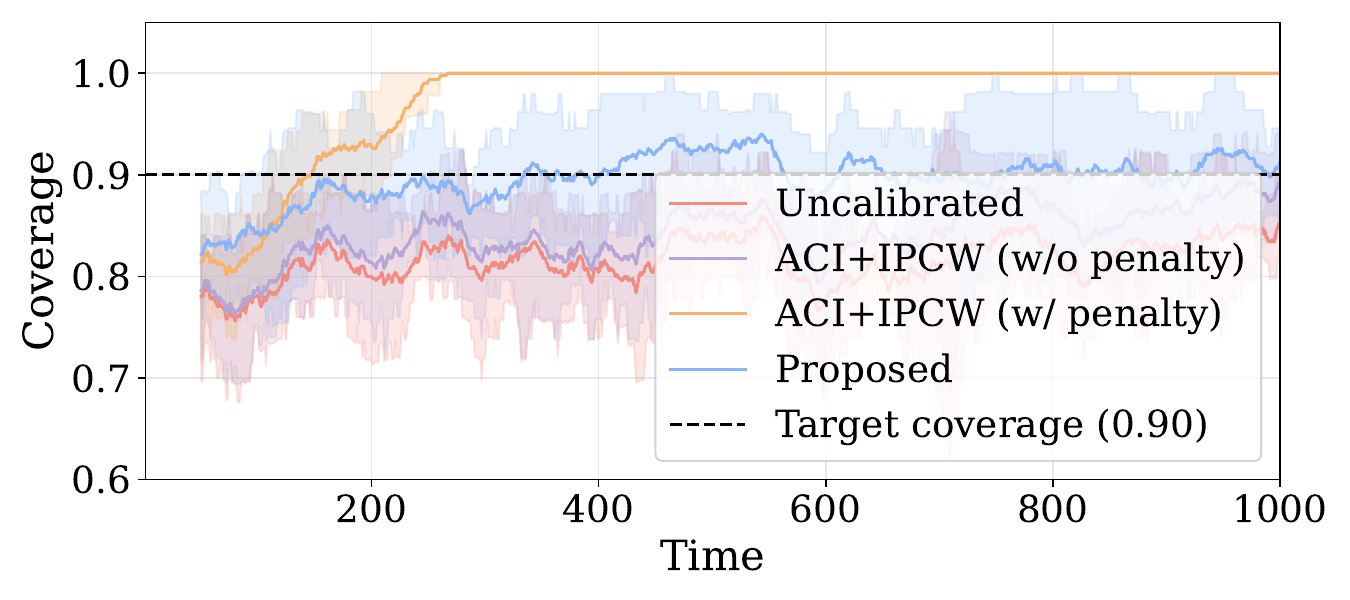}
        \caption{Local coverage over time, $\alpha = 0.1$, window$=50$}
    \end{minipage}\hfill
    \begin{minipage}[t]{0.41\linewidth}
        \vspace{0pt}
        
        \includegraphics[trim={1cm 0cm 1cm 0cm}, clip, width=\linewidth]{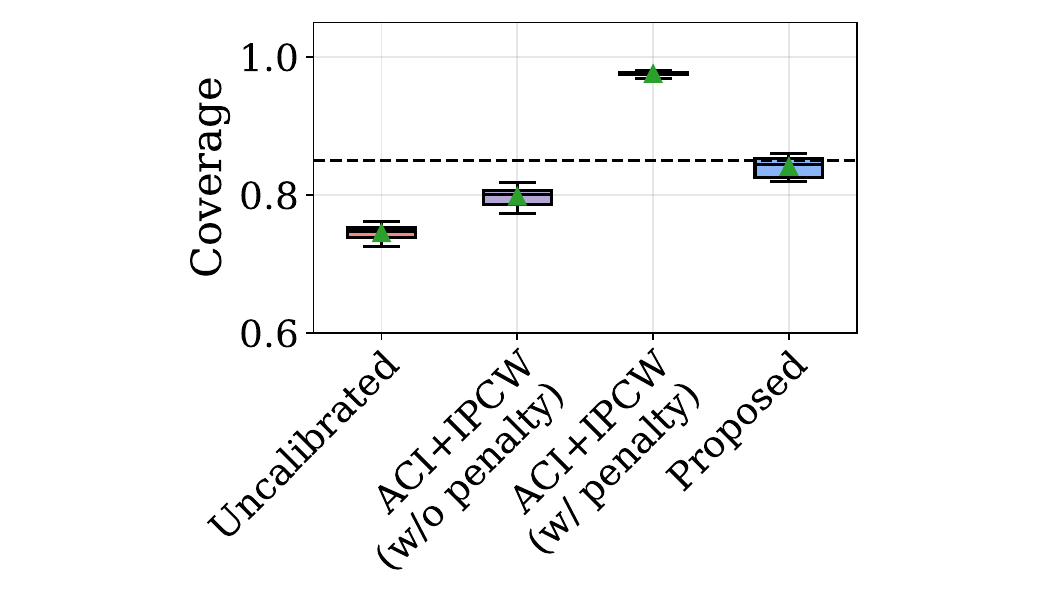}
        \caption{Cumulative coverage, $\alpha = 0.15$}
    \end{minipage}
\end{figure}

\begin{figure}[t!]
    \centering
    \begin{minipage}[t]{0.58\linewidth}
        \vspace{0pt}
        \includegraphics[width=\linewidth]{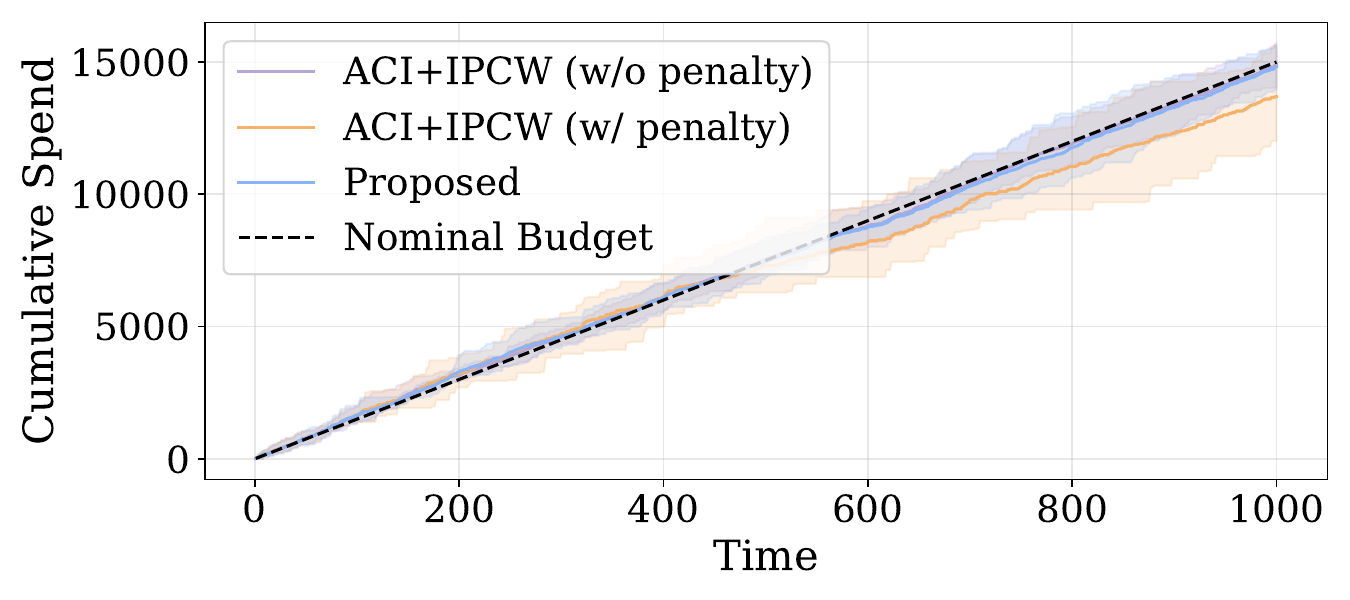}
        \caption{Cumulative budget spend over time, $\alpha = 0.1$}
    \end{minipage}\hfill
    \begin{minipage}[t]{0.41\linewidth}
        \vspace{0pt}
        \includegraphics[trim={1cm 0cm 1cm 0cm}, clip, width=\linewidth]{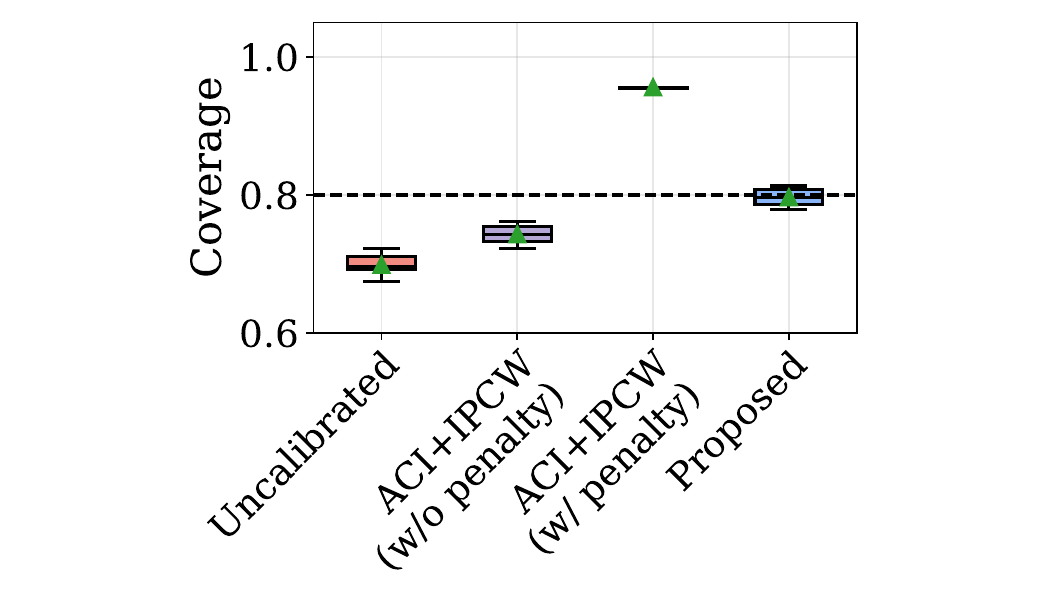}
        \caption{Cumulative coverage, $\alpha = 0.2$}
    \end{minipage}
\end{figure}

\subsubsection{Synthetic Data}
We next evaluate the calibration variants on the synthetic distribution shift dataset described in Appendix~\ref{app:exp_details:synthetic}. This controlled setting tests adaptation when the target distribution drifts upward while the uncalibrated predictive distribution remains stationary, progressively reducing the coverage of the uncalibrated UPBs. We compare \ourmethod{}, the two \texttt{ACI+IPCW} variants, and the uncalibrated baseline at nominal coverage $1-\alpha=0.9$, using $C_{\mathrm{task}}=5$. Each configuration is evaluated over ten repetitions, preserving the same temporal data order while varying the observation randomness.

\begin{figure}[t!]
    \centering
    \begin{minipage}[t]{0.58\linewidth}
        \vspace{0pt}
        \includegraphics[width=\linewidth]{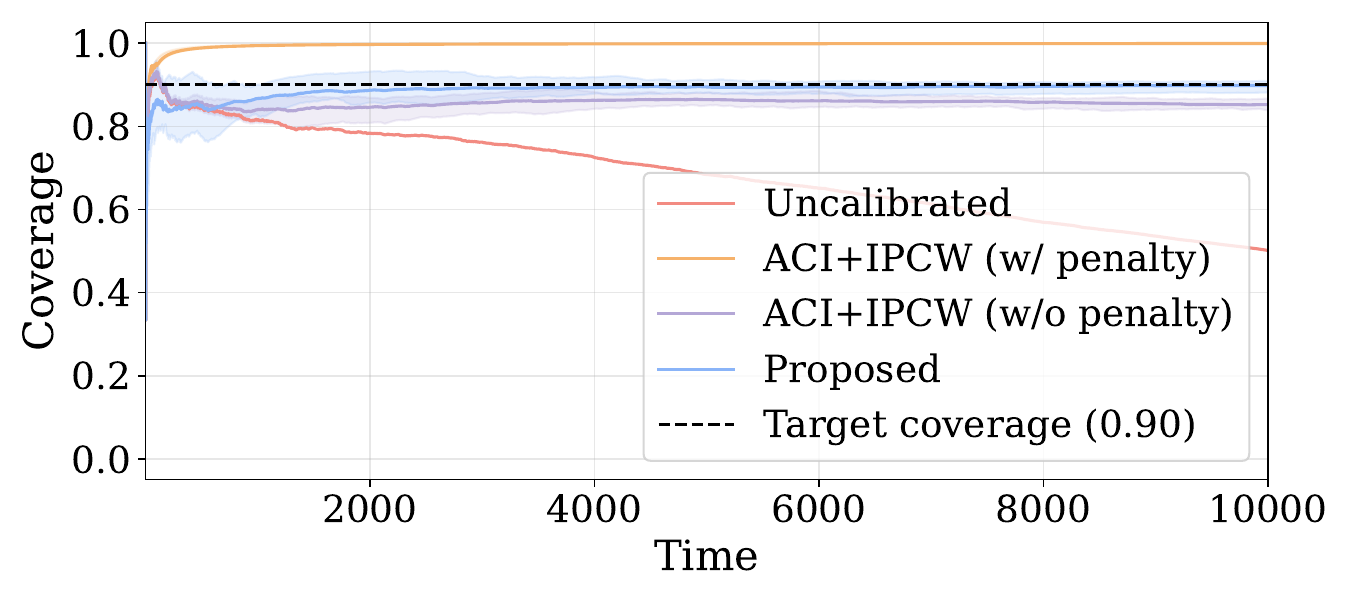}
        \caption{Cumulative coverage rate over time, $\alpha = 0.1$}
    \end{minipage}\hfill
    \begin{minipage}[t]{0.41\linewidth}
        \vspace{0pt}
        \includegraphics[trim={1cm 0cm 1cm 0cm}, clip, width=\linewidth]{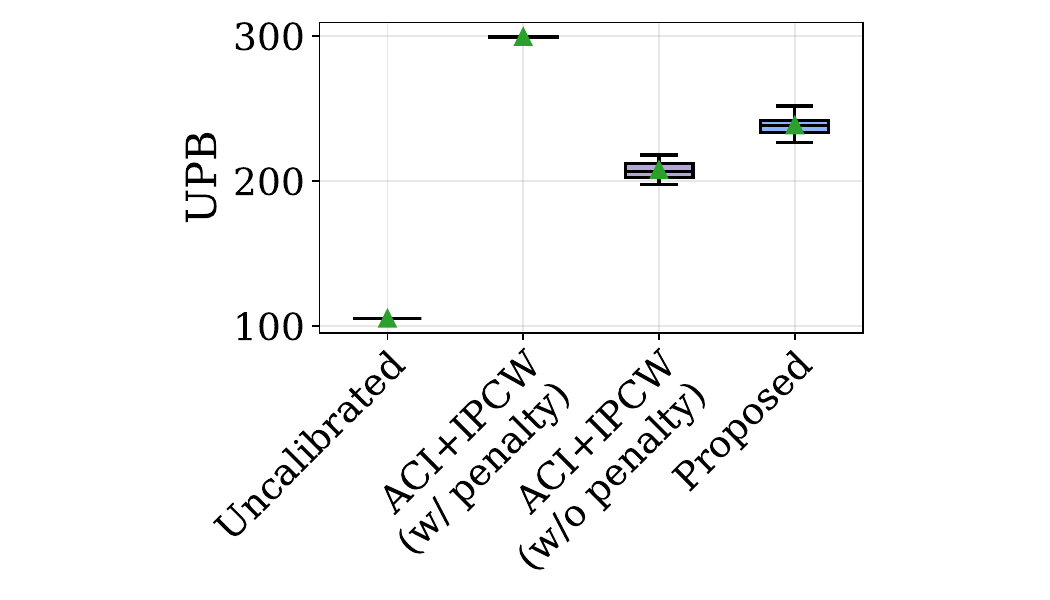}
        \caption{Upper predictive bound, $\alpha = 0.1$}
    \end{minipage}
\end{figure}

\begin{figure}[t!]
    \centering
    \begin{minipage}[t]{0.58\linewidth}
        \vspace{0pt}
        \includegraphics[width=\linewidth]{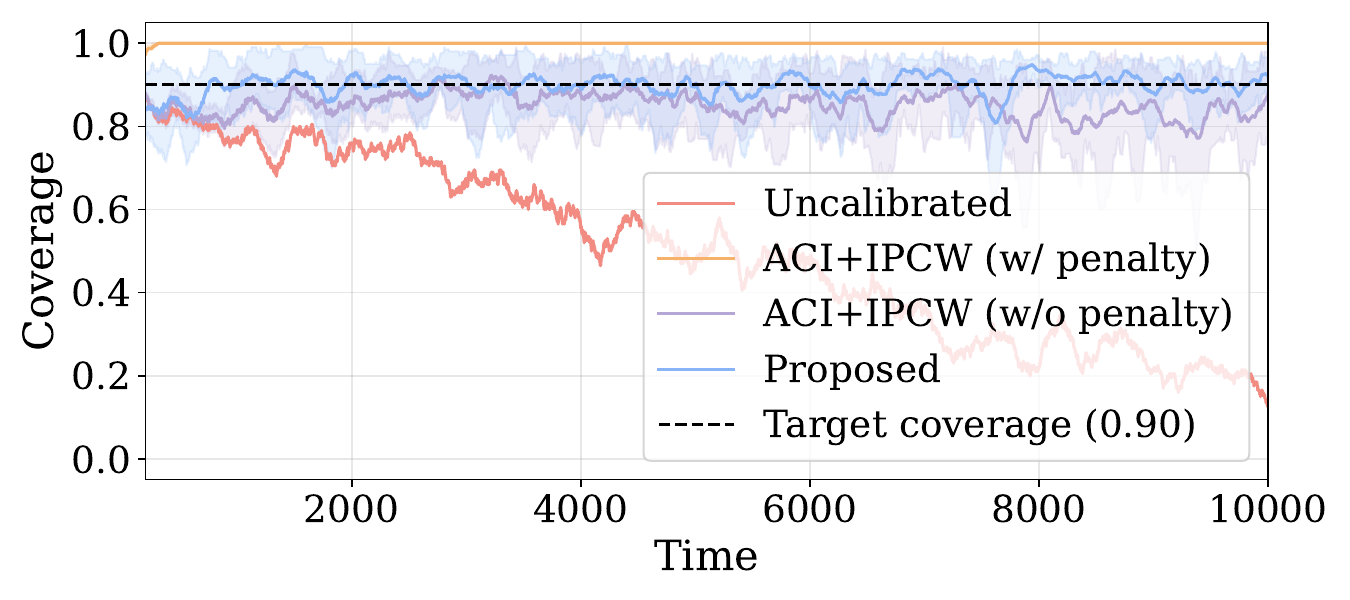}
        \caption{Local coverage over time, $\alpha = 0.1$, window$=200$}
    \end{minipage}\hfill
    \begin{minipage}[t]{0.41\linewidth}
        \vspace{0pt}
        
        \includegraphics[trim={1cm 0cm 1cm 0cm}, clip, width=\linewidth]{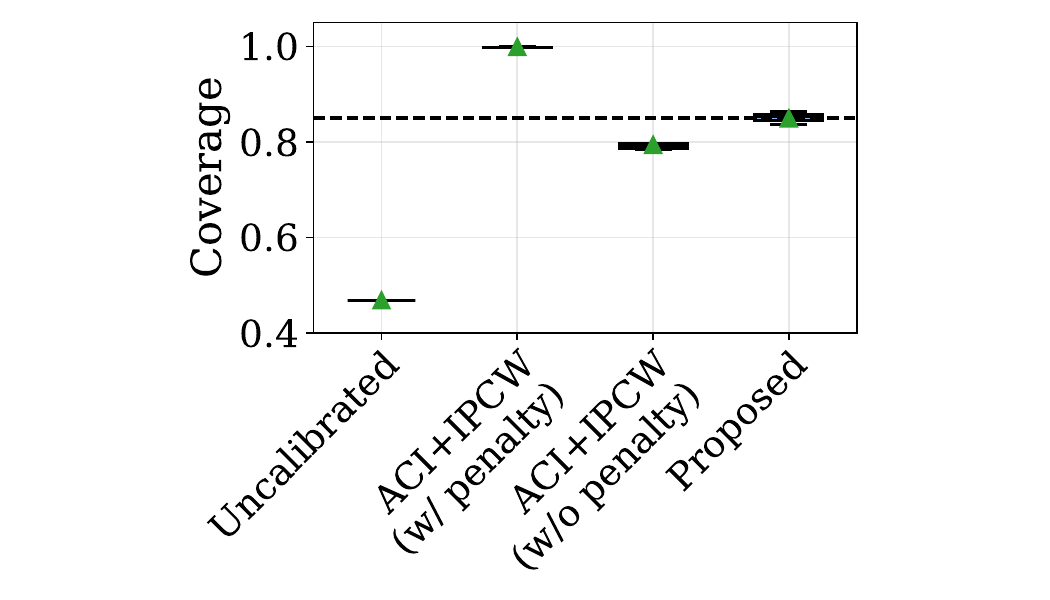}
        \caption{Cumulative coverage, $\alpha = 0.15$}
    \end{minipage}
\end{figure}

\begin{figure}[t!]
    \centering
    \begin{minipage}[t]{0.58\linewidth}
        \vspace{0pt}
        \includegraphics[width=\linewidth]{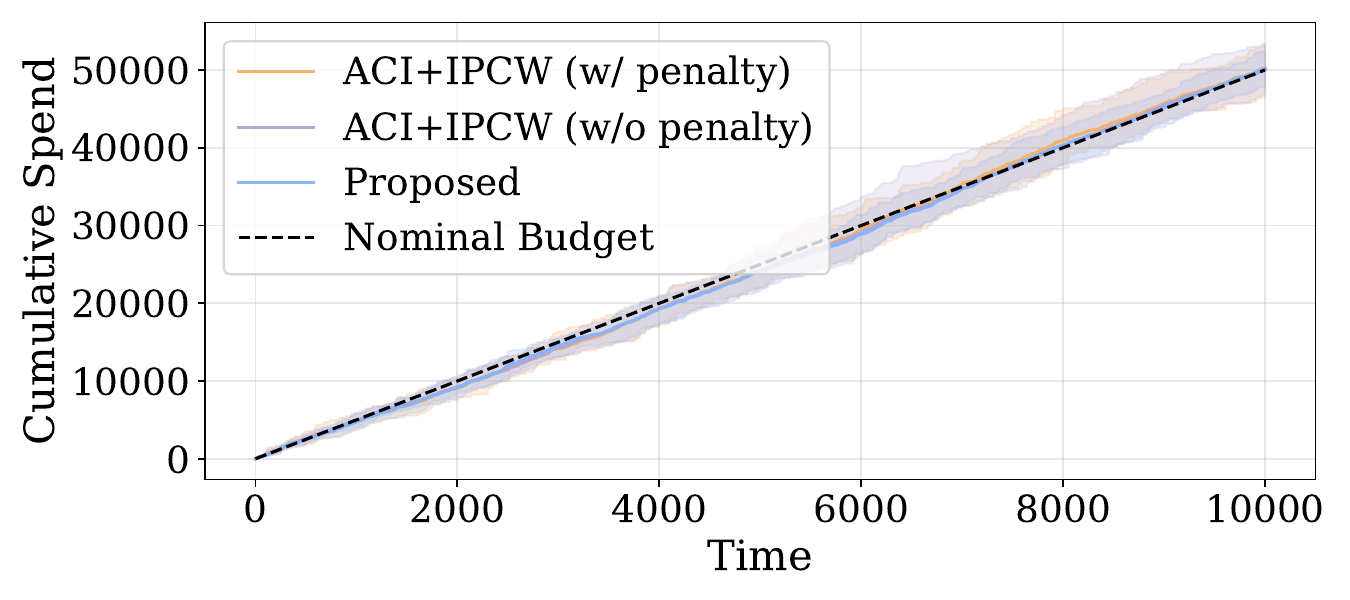}
        \caption{Cumulative budget spend over time, $\alpha = 0.1$}
    \end{minipage}\hfill
    \begin{minipage}[t]{0.41\linewidth}
        \vspace{0pt}
        \includegraphics[trim={1cm 0cm 1cm 0cm}, clip, width=\linewidth]{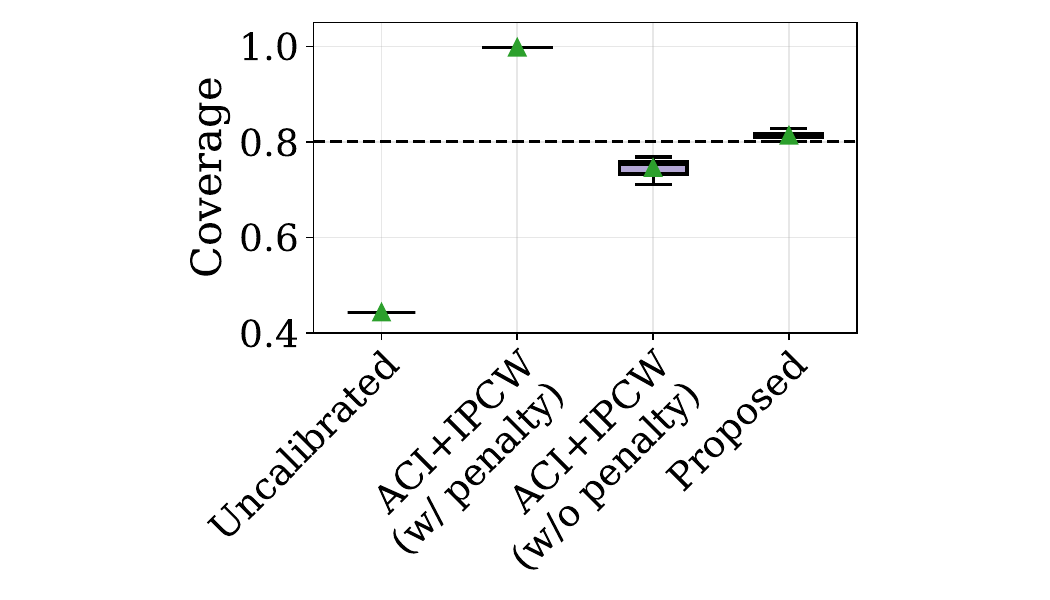}
        \caption{Cumulative coverage, $\alpha = 0.2$}
    \end{minipage}
\end{figure}

\subsubsection{Ablation of the Penalty and Exponential Parametrization}
\label{app:ablation-penalty-exp}

We investigate whether the advantage of our method can be explained solely by exponential parametrization~\citep{feldman2023achieving} or by adding a conservative penalty under ambiguous feedback. Table~\ref{tab:swe-ablation-budget-15} compares the proposed method, the uncalibrated baseline, and four ACI+IPCW variants obtained by independently enabling these two components. Results use the real SWE dataset with $C_{\mathrm{task}}=15.0$, $\alpha\in\{0.1,0.15,0.2\}$, and ten repetitions of 1,000 rounds. All ACI variants share the remaining hyperparameters, including $\gamma=5\times10^{-4}$. Entries report mean $\pm$ sample standard deviation across repetitions; local coverage averages complete 50-round windows.

In Table~\ref{tab:swe-ablation-budget-15}, the proposed method achieves mean coverage closest to the nominal target at every evaluated $\alpha$. Exponential ACI+IPCW undercovers even with the penalty, while the non-exponential variant with the penalty becomes overly conservative and produces much larger bounds. For example, at $\alpha=0.2$, the proposed method achieves coverage $0.7963$ with mean UPB $79.5382$, compared with $0.9126$ and $191.5697$ for the non-exponential penalized variant, at comparable spending. These results support an advantage beyond either modification alone under the tested settings, although they do not separately isolate the contribution of the proposed budget allocator.

\begin{table*}[ht]
\centering
\small
\caption{SWE ablation of the ACI+IPCW penalty and exponential parametrization at $C_{\mathrm{task}}=15.0$. Values are mean $\pm$ sample standard deviation across 10 repetitions of 1,000 rounds each. Local coverage is the mean coverage over all complete trailing 50-round windows within each repetition. Coverage and UPB are the final cumulative coverage and mean requested upper predictive bound, respectively. Budget spend is total assigned budget $\sum_t C_t$. The uncalibrated baseline uses full observation without the budget constraint. Bold entries in Local coverage and Coverage minimize the absolute distance of the mean from the target $1-\alpha$ among methods at the same $\alpha$.}
\label{tab:swe-ablation-budget-15}
\setlength{\tabcolsep}{4pt}
\resizebox{\textwidth}{!}{\begin{tabular}{lrrrrr}
\toprule
Method & $\alpha$ & Local coverage & Coverage & UPB & Budget spend \\
\midrule
Uncalibrated & 0.1 & $0.8162 \pm 0.0145$ & $0.8165 \pm 0.0156$ & $79.7751 \pm 0.8762$ & $80261.6000 \pm 876.0211$ \\
Uncalibrated & 0.15 & $0.7440 \pm 0.0120$ & $0.7443 \pm 0.0116$ & $68.3316 \pm 0.8254$ & $68834.4000 \pm 825.7039$ \\
Uncalibrated & 0.2 & $0.6978 \pm 0.0149$ & $0.6983 \pm 0.0148$ & $62.5447 \pm 0.9930$ & $62938.1000 \pm 996.2390$ \\
\addlinespace
ACI + IPCW w/ penalty w/ exp & 0.1 & $0.8450 \pm 0.0101$ & $0.8456 \pm 0.0112$ & $85.4615 \pm 0.9260$ & $14990.4000 \pm 928.3650$ \\
ACI + IPCW w/ penalty w/ exp & 0.15 & $0.7842 \pm 0.0097$ & $0.7840 \pm 0.0092$ & $74.9188 \pm 1.0486$ & $15476.7000 \pm 788.1963$ \\
ACI + IPCW w/ penalty w/ exp & 0.2 & $0.7398 \pm 0.0134$ & $0.7398 \pm 0.0126$ & $67.1935 \pm 0.9746$ & $14892.6000 \pm 809.8021$ \\
\addlinespace
ACI + IPCW w/o penalty w/ exp & 0.1 & $0.8253 \pm 0.0125$ & $0.8254 \pm 0.0140$ & $81.5104 \pm 0.8863$ & $15079.1000 \pm 797.0818$ \\
ACI + IPCW w/o penalty w/ exp & 0.15 & $0.7559 \pm 0.0099$ & $0.7560 \pm 0.0099$ & $69.6379 \pm 0.8447$ & $15409.0000 \pm 729.7825$ \\
ACI + IPCW w/o penalty w/ exp & 0.2 & $0.7055 \pm 0.0124$ & $0.7060 \pm 0.0121$ & $63.2919 \pm 0.8932$ & $14821.8000 \pm 778.4994$ \\
\addlinespace
ACI + IPCW w/ penalty w/o exp & 0.1 & $0.9771 \pm 0.0049$ & $0.9736 \pm 0.0048$ & $258.2220 \pm 1.7977$ & $13774.2000 \pm 2042.9827$ \\
ACI + IPCW w/ penalty w/o exp & 0.15 & $0.9552 \pm 0.0054$ & $0.9510 \pm 0.0059$ & $229.1209 \pm 3.5313$ & $15586.4000 \pm 2169.3867$ \\
ACI + IPCW w/ penalty w/o exp & 0.2 & $0.9148 \pm 0.0102$ & $0.9126 \pm 0.0099$ & $191.5697 \pm 4.5605$ & $14749.7000 \pm 1575.0946$ \\
\addlinespace
ACI + IPCW w/o penalty w/o exp & 0.1 & $0.8419 \pm 0.0106$ & $0.8420 \pm 0.0114$ & $85.0455 \pm 1.5883$ & $15008.2000 \pm 847.6282$ \\
ACI + IPCW w/o penalty w/o exp & 0.15 & $0.7762 \pm 0.0098$ & $0.7761 \pm 0.0095$ & $73.3383 \pm 2.1522$ & $15415.0000 \pm 714.1243$ \\
ACI + IPCW w/o penalty w/o exp & 0.2 & $0.7233 \pm 0.0118$ & $0.7237 \pm 0.0108$ & $65.2465 \pm 0.8760$ & $14899.0000 \pm 789.2252$ \\
\midrule
Proposed & 0.1 & \textbf{\boldmath $0.8962 \pm 0.0124$} & \textbf{\boldmath $0.8948 \pm 0.0124$} & $113.0419 \pm 19.6496$ & $14820.4000 \pm 963.8793$ \\
Proposed & 0.15 & \textbf{\boldmath $0.8419 \pm 0.0182$} & \textbf{\boldmath $0.8405 \pm 0.0165$} & $89.8948 \pm 6.0664$ & $15507.6000 \pm 835.3538$ \\
Proposed & 0.2 & \textbf{\boldmath $0.7983 \pm 0.0137$} & \textbf{\boldmath $0.7963 \pm 0.0129$} & $79.5382 \pm 3.6981$ & $14905.5000 \pm 820.1531$ \\
\bottomrule
\end{tabular}
}
\end{table*}

\section{Additional Related Work}
\label{app:related_work}

Our work builds on conformal prediction~\citep{vovk2005algorithmic,shafer2008tutorial}, including split conformal prediction~\citep{papadopoulos2002inductive} and weighted conformal methods for covariate shift~\citep{tibshirani2019conformal}. Conformal prediction has recently been extended to censored survival outcomes~\citep{candes2023conformalized,gui2024conformalized,sesia2024doubly,sesia2025conformal,davidovconformalized,davidov2026calibrated,feldman2026}, including history-aware methods with inverse-probability-of-censoring weighting~\citep{wang2026history}. Adaptive conformal inference further extends calibration to sequential and potentially nonstationary data by updating the nominal level according to past errors~\citep{gibbs2021adaptive,gibbs2024conformal, feldman2023achieving}. Our setting includes all these challenges: censoring hides coverage errors, while the censoring mechanism itself is chosen adaptively under a computational budget. Thus, we jointly adapt online the predictive bound and the mechanism that determines which outcomes are observed.

Our budget allocation is related to active learning~\citep{settles2009active}, optimal experimental design~\citep{fedorov2013theory}, and adaptive data collection for valid statistical inference~\citep{zrnic2024active,hahn2011adaptive}. Related works also study adaptive experimentation with censored survival outcomes~\citep{wang2026adaptive}. These methods typically decide which observations or participants to sample. In contrast, we decide how many interactions to sample from a sequential trajectory, which induces right censoring and changes the feedback available for the calibration. Furthermore, our allocator optimizes the stated model-based coverage criterion under the accumulating budget constraint.

Modern agent benchmarks evaluate systems over iterative trajectories, including software-engineering agents~\citep{jimenez2024swe,yang2024swe} and browser agents~\citep{zhou2024webarena}. Our framework complements standard evaluation of agentic utility by modeling the number of interactions until an event of interest. This perspective also applies to LLM safety, where prior work studies harmful texts~\citep{gehman2020realtoxicityprompts,Vidgen2023SimpleSafetyTestsAT}, inference-time guardrails~\citep{inan2023llama,rebedea2023nemo}, adversarial red teaming and jailbreaks~\citep{perez2022red,zou2023universal,wei2023jailbroken,mehrotra2024tree,chao2025jailbreaking}, and certified safety guarantees~\citep{kumar2023certifying,robey2023smoothllm}. These approaches focus on whether such events occur; in contrast, our objective is to provide calibrated bounds on \emph{when} an event occurs along a multi-turn trajectory.

Our framework also relates to methods that reduce the cost of model evaluation~\citep{kossen2021active}, including active testing~\citep{li2024active}, bandit-based allocation~\citep{zhou2025on}, adaptive dataset partitioning~\citep{wang2025cer}, and finite-population inference with adaptive benchmark-item selection~\citep{wu2026efficient}. These methods primarily choose \emph{which} benchmark items to evaluate. We instead choose \emph{how long} to evaluate each trajectory. Early stopping therefore produces censored observations rather than simply excluding examples, requiring the evaluation and budget-allocation procedures to account explicitly for censoring.

\section{Experimental Details}
\label{app:exp_details}

\subsection{Software Engineering Settings (SWE)}

\label{app:exp_details:swe}

\paragraph{Original Dataset.}
We use the \texttt{train} split of the
\texttt{nebius/SWE-agent-trajectories} 
dataset hosted on Hugging Face
Datasets~\citep{nebius2024sweagenttrajectories}. The collection contains
SWE-agent-style trajectories for resolving real-world GitHub software issues,
following the task setting introduced by SWE-bench and
SWE-agent~\citep{jimenez2024swebench,yang2024swe}. A raw record contains an
instance identifier, model name, trajectory, generated patch, exit status,
evaluation logs, and a Boolean resolution target. The Boolean target is not the
response modeled by our UPB predictor. Instead, we define
\(T^{\mathrm{true}}_i\) as the number of logged entries in the agent
trajectory. The parser first accepts trajectories represented as Python lists,
JSON strings, or Python-literal strings. If these parsing paths fail, it counts
the logged role markers and returns at least one step. The retrieved raw
artifact contains \(80{,}036\) records.

\paragraph{Feature Extraction.}
For each trajectory, we extract six lexical features from its first logged
entry: character length, line count, whitespace-delimited word count, number of
literal triple-backtick delimiters, the summed case-insensitive occurrence
count of \texttt{error}, \texttt{exception}, \texttt{traceback}, and
\texttt{fail}, and the summed case-sensitive occurrence count of
\texttt{.py}, \texttt{.json}, \texttt{.md}, \texttt{.sh}, and \texttt{.txt}.
We also retain the model name and derive the repository identifier from the
portion of \texttt{instance\_id} preceding \texttt{\_\_}. Using only the
training split, the categorical variables are target encoded as the
model-specific mean and standard deviation of \(T^{\mathrm{true}}\), together
with the repository-specific mean. An unseen model or repository receives the
global training mean; an unseen model receives a standard-deviation feature of
zero. Thus,
\[
 X_i =
 \bigl[
   \mu_{\mathrm{model}},
   \sigma_{\mathrm{model}},
   \mu_{\mathrm{repo}},
   n_{\mathrm{char}},
   n_{\mathrm{line}},
   n_{\mathrm{word}},
   n_{\mathrm{backtick}},
   n_{\mathrm{error}},
   n_{\mathrm{file}}
 \bigr]_i
 \in\mathbb{R}^{9}.
\]
The resulting matrix is stored as \texttt{float32} and standardized
coordinate-wise using means and standard deviations computed on the training
split. A standard deviation smaller than \(10^{-12}\) is replaced by one.

\paragraph{Quantile Predictor.}
Records with \(T^{\mathrm{true}}>300\) are removed, leaving \(79{,}090\)
examples; duplicate removal is disabled. A permutation generated with seed
\(42\) assigns \(500\) examples to training, \(55\) to validation, and the
remaining \(78{,}535\) to the reported holdout set. For every
\(\alpha\in\{0.01,0.02,\ldots,0.99\}\), we train a separate predictor for
quantile level \(\beta=1-\alpha\). Its architecture is
\[
  9
  \xrightarrow{\mathrm{Linear}}128
  \xrightarrow{\mathrm{ReLU}}128
  \xrightarrow{\mathrm{Linear}}128
  \xrightarrow{\mathrm{ReLU}}128
  \xrightarrow{\mathrm{Linear}}1
  \xrightarrow{\mathrm{Softplus}}1.
\]
The objective is the mean pinball loss
$
  \ell_\beta(y,\widehat y)
  =
  \max\!\left\{
    \beta(y-\widehat y),
    (\beta-1)(y-\widehat y)
  \right\}.
$
Each network is optimized for \(50\) epochs using AdamW, a learning rate of
\(2\times10^{-3}\), weight decay \(10^{-4}\), and shuffled mini-batches of size
\(512\). Validation loss and empirical coverage are logged, while the
parameters from the final epoch are retained. Predictions are upper-clipped at
\(300\). Device selection is automatic; the generated experiment metadata
records CUDA execution.

\subsection{Retrieval-Augmented Generation Settings (RAG)}

\label{app:exp_details:rag}

\paragraph{Original Dataset.}
We use the BEIR distribution of NFCorpus, loading queries from
\texttt{BeIR/nfcorpus} with configuration and split \texttt{queries}, and
relevance judgments from the \texttt{train} and \texttt{test} splits of
\texttt{BeIR/nfcorpus-qrels}. NFCorpus is a medical information-retrieval
dataset, and BEIR provides its standardized retrieval
representation~\citep{boteva2016nfcorpus,thakur2021beir}. The raw task is to
retrieve corpus items relevant to a natural-language medical query. After
verifying that query--corpus identifier pairs are unique, we retain judgments
whose relevance score is at least one. We define
\[
  T^{\mathrm{true}}_i
  =
  \#\{\text{retained relevance judgments associated with query }i\}.
\]
The query table is inner-joined with these counts, producing \(2{,}913\) raw
examples.

\paragraph{Feature Extraction.}
Only the query text, denoted by \(\mathrm{query}_i\) for query \(i\), is used to construct \(X_i\). Each query is encoded with the
SentenceTransformers checkpoint
\texttt{sentence-transformers/all-MiniLM-L6-v2}, using an encoding batch size
of \(64\)~\citep{reimers2019sentencebert,wang2020minilm,allminilml6v2}. The
checkpoint produces a \(384\)-dimensional embedding. We append the query's
character length divided by \(100\), giving
\[
  X_i
  =
  \left[
    \operatorname{Embed}_{\texttt{all-MiniLM-L6-v2}}(\mathrm{query}_i);
    \frac{\operatorname{chars}(\mathrm{query}_i)}{100}
  \right]
  \in\mathbb{R}^{385}.
\]
The matrix is converted to \texttt{float32} and standardized coordinate-wise
using the training-set mean and standard deviation, with standard deviations
below \(10^{-12}\) replaced by one. The repository pipeline does not add
document embeddings or document-level metadata.

\paragraph{Quantile Predictor.}
Examples with \(T^{\mathrm{true}}>200\) are excluded, leaving \(2{,}804\)
examples; duplicate removal is disabled. With random seed \(42\), the pipeline
uses \(500\) training examples, \(55\) validation examples, and \(2{,}249\)
holdout examples. A separate network is trained for every
\(\alpha\in\{0.01,0.02,\ldots,0.99\}\), with \(\beta=1-\alpha\) and architecture
\[
  385
  \xrightarrow{\mathrm{Linear}}128
  \xrightarrow{\mathrm{ReLU}}128
  \xrightarrow{\mathrm{Linear}}128
  \xrightarrow{\mathrm{ReLU}}128
  \xrightarrow{\mathrm{Linear}}1
  \xrightarrow{\mathrm{Softplus}}1.
\]
Training minimizes the mean pinball loss
$
  \ell_\beta(y,\widehat y)
  =
  \max\!\left\{
    \beta(y-\widehat y),
    (\beta-1)(y-\widehat y)
  \right\}
$
using AdamW for \(50\) epochs, with learning rate \(2\times10^{-3}\), weight
decay \(10^{-4}\), and shuffled mini-batches of size \(512\). Validation loss
and empirical coverage are logged, but the final-epoch parameters are used.
Predictions are upper-clipped at \(200\). Device selection is automatic; the
generated experiment metadata records CUDA execution.

\subsection{Instruction Following Settings (AutoIF)}
\label{app:exp_details:autoif}

\paragraph{Original Dataset.}
To generate the raw dataset, we followed the procedure of ~\citet{davidov2026calibrated, feldman2026}. We sample \(10{,}000\) instruction-following tasks from the AutoIF dataset, as well as \(10{,}000\) harmful prompts from the Anthropic Red Team dataset for adversarial jailbreaks. We generate the raw multi-turn attacker-target conversations using the PAIR algorithm. We deploy Qwen 2.5 14B Instruct as the attacking model, which acts as a helper agent for the AutoIF dataset to maximize the probability that the target model passes programmatic constraints. For the target models, we evaluate safety and utility across Qwen 2.5 14B Instruct, Llama 3.1 8B Instruct, Phi 4 Mini Instruct, and Gemma 3 4B Instruct. In all generated conversations, we cap the maximum sequence length at \(t_{\max}=200\) turns. The resulting dataset of interactions is divided into three disjoint subsets: a training set of \(4{,}000\) prompts, a calibration set of \(3{,}000\) prompts, and a test set of \(3{,}000\) prompts.

\paragraph{Feature Extraction.}
We embed all texts into vectors using the \texttt{Multilingual-E5-large} model. These high-dimensional text embeddings of the conversation history serve as the inputs for our predictive model.

\paragraph{Quantile Predictor.}
The predictive model architecture begins by projecting the high-dimensional embeddings into a lower-dimensional latent space of \(256\) using a linear layer, followed by a ReLU activation and dropout with a rate of \(0.2\). To capture the temporal dependencies of the conversational history, we apply a Transformer Encoder utilizing \(3\) pre-normalized Transformer layers with \(6\) attention heads, a feed-forward dimension of \(1024\), and GELU activations. An upper-triangular attention mask is applied during encoding to prevent the model from using future time steps. The sequence encoding is then concatenated with the initial projected embeddings and passed through a multi-layer perceptron (three linear layers separated by ReLU activations and \(0.2\) dropout) to produce the raw temporal logits. For task \(t\), the model outputs a \(t_{\max} \times (t_{\max} + 1)\) matrix containing the probabilities of an event occurring at any future step, \(\mathbb{P}(T_t=k_2\mid T_t>k_1,H_t(k_1))\), and the probability of reaching the horizon without an event. Here, \(k_1\) and \(k_2\) are the current and future within-conversation steps, respectively. The network is trained using the Adam optimizer with a learning rate of \(10^{-4}\), weight decay of \(0.001\), and a batch size of \(64\). We optimize the model for a maximum of \(500\) epochs to minimize the survival negative log-likelihood loss, stopping early if the validation loss does not improve for \(50\) consecutive epochs. Finally, the outputs are calibrated on a held-out validation set (a \(10\%\) portion of the training set) using a temperature scaling L-BFGS optimizer.

\subsection{Synthetic Data Settings}
\label{app:exp_details:synthetic}

\paragraph{Data Generation.}
We generate $N=10{,}000$ sequential examples with targets
\[
T_i=\mu_i+\varepsilon_i, \qquad
\mu_i=95+(115-95)\frac{i-1}{N-1}, \qquad
\varepsilon_i\overset{\mathrm{iid}}{\sim}\mathcal{N}(0,10^2).
\]
The target mean increases linearly from $95$ to $115$. All examples are
processed without shuffling, preserving the temporal shift.

\paragraph{Uncalibrated Predictions.}
For each example $i$, the 99 uncalibrated UPBs are defined as
\[
U^{\mathrm{uncal}}_{i,\alpha}
=100+\xi_i+4\Phi^{-1}(1-\alpha), \qquad
\alpha\in\{0.01,0.02,\ldots,0.99\},
\]
where $\xi_i\overset{\mathrm{iid}}{\sim}\operatorname{Uniform}(-1,1)$ and
$\Phi$ denotes the standard normal cumulative distribution function. The
predictive-location noise is independent of the target noise. Thus, the
uncalibrated predictive distribution remains stationary while the target
distribution drifts upward.

\subsection{Additional Implementation Details}
\label{app:exp_details:implementation}

Raw predictions at different quantile levels can cross. For each task, let
$\tau_1^{\mathrm{grid}}>\cdots>\tau_K^{\mathrm{grid}}>0$ denote the available tail levels no greater than
$\tau_{\max}$, and let $r_{t,j}$ be the corresponding raw predictions.
We construct the conservative envelope
$
    v_{t,j}=\max_{1\le i\le j}r_{t,i},
    \; d_j=\log(\tau_{\max}/\tau_j^{\mathrm{grid}}).
$
Thus the repaired bound increases as the requested tail probability decreases,
and no table entry is reduced. This is the smallest monotone majorant in the given ordering. Writing $q_t(a)=\widehat q_t(X_t;\tau_{\max}e^{-a})$, we set $q_t(a)=v_{t,j}$ for the first index $j$ satisfying $a\le d_j$.
The resulting function is piecewise constant, with each endpoint assigned to the interval on its left. 

Furthermore, our ACI$+$IPCW implementation clips the calibration parameter $\tau_t$ to $[0,1]$. Under tight
budgets, zero-budget rounds yield raw IPCW feedback $\widehat e_t=0$, increasing $\tau_t$ and shrinking subsequent bounds, while clipping can truncate the compensating updates from observed errors weighted by $1/\pi_t$. We therefore apply a small conservative adjustment, using
$\widehat e_t^{\mathrm{pen}}
=\widehat e_t+\mathbf{1}\{C_t=0\}$,
which assigns unit feedback on skipped rounds. Under the two-point allocation in~\eqref{eq:probe-two-point-law},
$
    \mathbb E\!\left[
        \widehat e_t^{\mathrm{pen}}
        \,\middle|\,X_t,\mathcal D_{t-1}
    \right]
    =m_t+1-G_t(U_t)\ge m_t.
$
This conservative modification preserves the coverage lower-bound argument for the unprojected update in Theorem~\ref{thm:aci-ipcw-coverage}, with asymptotic validity retaining its calibration-state condition. Its behavior with practical clipping is assessed empirically in Appendix~\ref{app:additional_experiments:aci_variants}, motivating its use in the main evaluation.

Additionally, the completion distribution \(\widehat F_t\) can be learned using a dedicated predictive model; we use a lightweight estimator to reduce computational overhead and enable faster experimentation. Full implementation details, all hyperparameters, and the configurations used to generate the experimental results reported in this paper are available 
in our \href{https://github.com/Roy-Maor-Lotan/PROBE}{GitHub}

\subsection{Compute Infrastructure and Hardware Specifications}
\label{app:exp_details:compute_and_hardware}
All experiments were conducted on a Linux server running Ubuntu 22.04.5 LTS, equipped with an Intel Xeon 6746E processor (224 threads) and 2 TB of RAM. Hardware acceleration was provided by a cluster of 8 NVIDIA H200 NVL GPUs, each with 141 GB of VRAM. The experimental framework and models were implemented in Python using PyTorch 2.13.0 (CUDA 13.2, cuDNN 9.2.0).

\end{document}